\documentclass[twoside,11pt]{article}

\usepackage{jmlr2e}

\usepackage{changepage}
\usepackage{booktabs}
\usepackage{tikz}
\usepackage{acronym}

\usepackage[latin1]{inputenc}
\usepackage{graphicx}

\graphicspath{{./figures/}}
\usepackage[strings]{underscore}
\usepackage{adjustbox}
\usepackage[strings]{underscore}
\usepackage{multirow}
\usepackage{colortbl}
\usepackage{blindtext}
\usepackage{scrextend}
\usepackage{array}
\usepackage{acronym}
\usepackage{amsmath}
\usepackage{dsfont}
\usepackage{array}
\usepackage{cite}
\usepackage{url}
\usepackage{xcolor}
\usepackage{rotating} 
\newacro{alba}[ALBA]{Active Learning-Based Approach}
\newacro{ann}[ANN]{Artificial Neural Network}
\newacro{ava}[AVA]{Aggregate Valuation of Antecedents}
\newacro{anova}[ANOVA]{Analysis of Variance}
\newacro{astrid}[ASTRID]{Automatic Structure Identification}
\newacro{beta}[BETA]{Black box Explanations through Transparent Approximations}
\newacro{biore}[BIO-RE]{Binarized Input-Output Rule Extraction}
\newacro{birch}[BIRCH]{Balanced Iterative Reducing and Clustering using Hierarchies}
\newacro{blm}[BLM]{Bayesian List Machine}
\newacro{boa}[BOA]{Bayesian Or's of And's}
\newacro{brianne}[BRIANNE]{Building Representations for Artificial Intelligence using Neural Networks}
\newacro{brl}[BRL]{Bayesian Rule List}
\newacro{brs}[BRS]{Bayesian Rule Set}
\newacro{bur}[BUR]{Bust Unordered Rule}
\newacro{cart}[CART]{Classification and Regression Trees}
\newacro{cf}[CF]{Clustering Feature}
\newacro{ci}[CI]{Confidence Interval}
\newacro{clear}[CLEAR]{Class-Enhanced Attentive Response}
\newacro{clurefi}[CluReFI]{Cluster-Representatives' Feature Importance}
\newacro{cnn}[CNN]{Convolutional Neural Network}
\newacro{cred}[CRED]{Continuous Rule Extractor via Decision tree induction}
\newacro{deepred}[DeepRED]{Deep Neural Network Rule Extraction via Decision tree induction}
\newacro{df}[df]{degrees of freedom}
\newacro{dhc}[DHC]{hill climbing with downhill moves}
\newacro{dnn}[DNN]{Deep Neural Network}
\newacro{dream}[DReaM]{Discriminative Rectangle Mixture}
\newacro{ererx}[E-Re-RX]{Ensemble-Recursive-Rule Extraction}
\newacro{eu}[EU]{European Union}
\newacro{fab}[FAB]{factorized asymptotic Bayesian}
\newacro{fernn}[FERNN]{Fast Extraction of Rules from Neural Networks}
\newacro{gam}[GAM]{Generalized Additive Model}
\newacro{ga2ms}[GA\textsuperscript{2}Ms]{Generalized Additive Models plus Interactions}
\newacro{gdpr}[GDPR]{General Data Protection Regulation}
\newacro{gex}[GEX]{Genetic Rule Extraction}
\newacro{gfa}[GFA]{Gradient Feature Auditing}
\newacro{glm}[GLM]{Generalized Linear Model}
\newacro{gradcam}[Grad-CAM]{Gradient-weighted Class Activation Mapping}
\newacro{grex}[G-Rex]{Genetic-Rule Extraction}
\newacro{grg}[GRG]{Greedy Rule Generation}
\newacro{gsa}[GSA]{Global Sensitivity Analysis}
\newacro{ice}[ICE]{Individual Conditional Expectation}
\newacro{ids}[IDS]{Interpretable Decision Set}
\newacro{ime}[IME]{Interaction-based Method for Explanations}
\newacro{kdd}[KDD]{Knowledge Discovery in Databases}
\newacro{kt}[KT]{Knowledgetron}
\newacro{lda}[LDA]{latent Dirichlet allocation}
\newacro{lime}[LIME]{Local Interpretable model-agnostic Explanations}
\newacro{loco}[LOCO]{Leave-One-Covariate-Out}
\newacro{lrcn}[LRCN]{Long-term Recurrent Convolutional Networks}
\newacro{lrp}[LRP]{Layerwise Relevance Propagation}
\newacro{lstm}[LSTM]{Long Short-Term Memory}
\newacro{lstm}[LSTM]{Long Short-Term Memories}
\newacro{maple}[MAPLE]{Model Agnostic Supervised Local Explanations}
\newacro{mds}[MDS]{Multi-Dimensional Scaling}
\newacro{mes}[MES]{Model Explanation System}
\newacro{mgm}[MGM]{Mind the Gap Mode}
\newacro{mio}[MIO]{mixed integer optimization}
\newacro{mlp}[MLP]{Multi-layer Perceptron}
\newacro{mmd}[MMD]{maximum mean discrepancy}
\newacro{mse}[MSE]{mean squared error}
\newacro{msg}[MSG]{mean square between groups}
\newacro{msgl}[MSGL]{Multi-class Sparse Group Lasso}
\newacro{nb}[NB]{Naive Bayes}
\newacro{nn}[NN]{Neural Network}
\newacro{orc}[ORC]{Ordered Rules for Classification}
\newacro{osre}[OSRE]{Orthogonal Search based Rule Extraction}
\newacro{otspam}[OT-SpAM]{Oblique Treed Sparse Additive Model}
\newacro{pca}[PCA]{Principal Component Analysis}
\newacro{pdp}[PDP]{Partial Dependence Plot}
\newacro{pilm}[PILM]{Personalized Integer Linear Model}
\newacro{pjx}[PJ-X]{Pointing and Justification}
\newacro{pvm}[PVM]{Prototype Vector Machine}
\newacro{qii}[QII]{Quantitative Input Influence}
\newacro{rbf}[RBF]{radial basis function}
\newacro{refne}[REFNE]{Rule Extraction From Neural Networks Ensemble}
\newacro{relu}[ReLU]{Rectified Linear Unit}
\newacro{rerx}[Re-RX]{Recursive Rule Extraction}
\newacro{rf}[RF]{Random Forest}
\newacro{ripper}[RIPPER]{Repeated Incremental Pruning to Produce Error}
\newacro{rxncm}[RxNCM]{Rule Extraction from Neural Networks using Classified and Miss-classified data}
\newacro{rxren}[RxREN]{Rule Extraction by Reverse Engeneering}
\newacro{sd}[SD]{Standard Deviation}
\newacro{sem}[SEM]{standard error of the mean}
\newacro{sgl}[SGL]{Sparse Group Lasso}
\newacro{p}[Sig.]{significance (p-value)}
\newacro{slim}[SLIM]{Supersparse Linear Integer Model}
\newacro{sne}[SNE]{Stochastic Neighbor Embedding}
\newacro{socrat}[SOCRAT]{Structured-Output Causal Rationalizer}
\newacro{som}[SOM]{self-organizing map}
\newacro{srae}[SRAE]{Sparse Reconstruction Autoencoder}
\newacro{stare}[STARE]{Statistics based Rule Extraction}
\newacro{splime}[SP-LIME]{Sub-modular pick \ac{lime}}
\newacro{svm}[SVM]{Support Vector Machine}
\newacro{tilm}[TILM]{Threshold-Rule Integer Linear Model}
\newacro{vec}[VEC]{Variable Effect Characteristic}
\newacro{via}[VIA]{Validity Interval Analysis}
\newacro {muse}[MUSE] {Model Understanding through Subspace Explanations}
\newacro{lore}[LORE]{LOcal Rule-based Explanations}
\newacro{mcr}[MCR]{Model Class Reliance}
\newacro{shap}[SHAP]{SHapley Additive exPlanations}
\newacro{xai}[XAI]{Explainable Artificial Intelligence}
\newacro{sml}[SML]{Supervised Machine Learning}
\newacro{cart}[CART]{Classification and Regression Trees}
\newacro{abs}[ABS]{Attention Based Summarization}
\newacro{iabm}[IABM]{Image Attention Based Model}
\newacro{sa}[SA]{Sensitivity Analysis}
\newacro{lrp}[LRP]{Layer-wise relevance propagation}
\newacro{senn}[SENN]{Self Explaining Neural Networks}
\newacro{vem}[VEM]{Visual Explanation Model}
\newacro{pj-x}[PJ-X]{Pointing and Justification-based Explanation}
\newacro{icm}[ICM]{Image Captioning Model}
\newacro{ndt}[NDT]{Neural Decision Tree}
\newacro{dndt}[DNDT]{Deep Neural Decision Tree}
\newacro{mgm}[MGM]{Mind the Gap Model}
\newacro{aot}[AOT]{AND-OR Template}

\def\Sec#1{Section~\ref{#1}}

\def\Fig#1{Fig.~\ref{#1}}

\usepackage{pifont}
\newcommand{\xmark}{\ding{55}}%
\newcommand{\cmark}{\ding{51}}%
\usepackage{enumitem}

\usepackage{amssymb}
\usepackage{fancyvrb}

\usepackage{subcaption}
\usepackage{theorem}
\usepackage{dsfont}
\usepackage{adjustbox}
\usepackage{mathtools}
\usepackage{color, colortbl}
\usepackage{listings}
\definecolor{backcolour}{rgb}{0.95,0.95,0.92}
\definecolor{codegreen}{rgb}{0,0.6,0}
\definecolor{codegray}{rgb}{0.5,0.5,0.5}
\definecolor{codepurple}{rgb}{0.58,0,0.82}

\theoremstyle{definition}
\newtheorem{Assumption}{Assumption}
\newtheorem{Theorem}{Theorem}
\newtheorem{Corollary}{Corollary}
\newtheorem{Definition}{Definition}
\newtheorem{Problem}{Problem}
\theoremheaderfont{\bfseries}
\theoremstyle{plain}

\input{./math.tex}

\def\Sec#1{Section~\ref{#1}}

\def\Fig#1{Fig.~\ref{#1}}

\theoremstyle{definition}

\newtheorem{Problem}{Problem}
\theoremheaderfont{\bfseries}
\theoremstyle{plain}

\begin{acronym}
\newacro{alba}[ALBA]{Active Learning-Based Approach}
\newacro{ann}[ANN]{Artificial Neural Network}
\newacro{ava}[AVA]{Aggregate Valuation of Antecedents}
\newacro{anova}[ANOVA]{Analysis of Variance}
\newacro{astrid}[ASTRID]{Automatic Structure Identification}
\newacro{beta}[BETA]{Black Box Explanations through Transparent Approximations}
\newacro{biore}[BIO-RE]{Binarized Input-Output Rule Extraction}
\newacro{birch}[BIRCH]{Balanced Iterative Reducing and Clustering Using Hierarchies}
\newacro{blm}[BLM]{Bayesian List Machine}
\newacro{boa}[BOA]{Bayesian Or's of And's}
\newacro{brianne}[BRIANNE]{Building Representations for Artificial Intelligence Using Neural Networks}
\newacro{brl}[BRL]{Bayesian Rule List}
\newacro{brs}[BRS]{Bayesian Rule Set}
\newacro{bur}[BUR]{Bust Unordered Rule}
\newacro{cart}[CART]{Classification and Regression Trees}
\newacro{cf}[CF]{Clustering Feature}
\newacro{ci}[CI]{Confidence Interval}
\newacro{clear}[CLEAR]{Class-Enhanced Attentive Response}
\newacro{clurefi}[CluReFI]{Cluster-Representatives' Feature Importance}
\newacro{cnn}[CNN]{Convolutional Neural Network}
\newacro{cred}[CRED]{Continuous Rule Extractor via Decision Tree Induction}
\newacro{deepred}[DeepRED]{Deep Neural Network Rule Extraction via Decision Tree Induction}
\newacro{df}[df]{Degrees of Freedom}
\newacro{dhc}[DHC]{Hill Climbing with Downhill Moves}
\newacro{dnn}[DNN]{Deep Neural Network}
\newacro{dream}[DReaM]{Discriminative Rectangle Mixture}
\newacro{ererx}[E-Re-RX]{Ensemble-Recursive-Rule Extraction}
\newacro{eu}[EU]{European Union}
\newacro{fab}[FAB]{Factorized Asymptotic Bayesian}
\newacro{fernn}[FERNN]{Fast Extraction of Rules from Neural Networks}
\newacro{gam}[GAM]{Generalized Additive Model}
\newacro{ga2ms}[GA\textsuperscript{2}Ms]{Generalized Additive Models plus Interactions}
\newacro{gdpr}[GDPR]{General Data Protection Regulation}
\newacro{gex}[GEX]{Genetic Rule Extraction}
\newacro{gfa}[GFA]{Gradient Feature Auditing}
\newacro{glm}[GLM]{Generalized Linear Model}
\newacro{gradcam}[Grad-CAM]{Gradient-Weighted Class Activation Mapping}
\newacro{grex}[G-Rex]{Genetic-Rule Extraction}
\newacro{grg}[GRG]{Greedy Rule Generation}
\newacro{gsa}[GSA]{Global Sensitivity Analysis}
\newacro{ice}[ICE]{Individual Conditional Expectation}
\newacro{ids}[IDS]{Interpretable Decision Set}
\newacro{ime}[IME]{Interaction-Based Method for Explanations}
\newacro{kdd}[KDD]{Knowledge Discovery in Databases}
\newacro{kt}[KT]{Knowledgetron}
\newacro{lda}[LDA]{Latent Dirichlet Allocation}
\newacro{lime}[LIME]{Local Interpretable Model-Agnostic Explanations}
\newacro{loco}[LOCO]{Leave-One-Covariate-Out}
\newacro{lrcn}[LRCN]{Long-Term Recurrent Convolutional Networks}
\newacro{lrp}[LRP]{Layerwise Relevance Propagation}
\newacro{lstm}[LSTM]{Long Short-Term Memory}
\acroplural{lstm}[LSTM]{Long Short-Term Memories}
\newacro{maple}[MAPLE]{Model-Agnostic Supervised Local Explanations}
\newacro{mds}[MDS]{Multi-Dimensional Scaling}
\newacro{mes}[MES]{Model Explanation System}
\newacro{mgm}[MGM]{Mind the Gap Mode}
\newacro{mio}[MIO]{Mixed Integer Optimization}
\newacro{mlp}[MLP]{Multi-Layer Perceptron}
\newacro{mmd}[MMD]{Maximum Mean Discrepancy}
\newacro{mse}[MSE]{Mean Squared Error}
\newacro{msg}[MSG]{Mean Square between Groups}
\newacro{msgl}[MSGL]{Multi-Class Sparse Group Lasso}
\newacro{nb}[NB]{Naive Bayes}
\newacro{nn}[NN]{Neural Network}
\newacro{orc}[ORC]{Ordered Rules for Classification}
\newacro{osre}[OSRE]{Orthogonal Search-Based Rule Extraction}
\newacro{otspam}[OT-SpAM]{Oblique Treed Sparse Additive Model}
\newacro{pca}[PCA]{Principal Component Analysis}
\newacro{pdp}[PDP]{Partial Dependence Plot}
\newacro{pilm}[PILM]{Personalized Integer Linear Model}
\newacro{pjx}[PJ-X]{Pointing and Justification}
\newacro{pvm}[PVM]{Prototype Vector Machine}
\newacro{qii}[QII]{Quantitative Input Influence}
\newacro{rbf}[RBF]{Radial Basis Function}
\newacro{refne}[REFNE]{Rule Extraction From Neural Networks Ensemble}
\newacro{relu}[ReLU]{Rectified Linear Unit}
\newacro{rerx}[Re-RX]{Recursive Rule Extraction}
\newacro{rf}[RF]{Random Forest}
\newacro{ripper}[RIPPER]{Repeated Incremental Pruning to Produce Error}
\newacro{rxncm}[RxNCM]{Rule Extraction from Neural Networks using Classified and Miss-Classified Data}
\newacro{rxren}[RxREN]{Rule Extraction by Reverse Engineering}
\newacro{sd}[SD]{Standard Deviation}
\newacro{sem}[SEM]{Standard Error of the Mean}
\newacro{sgl}[SGL]{Sparse Group Lasso}
\newacro{p}[Sig.]{Significance (p-value)}
\newacro{slim}[SLIM]{Supersparse Linear Integer Model}
\newacro{sne}[SNE]{Stochastic Neighbor Embedding}
\newacro{socrat}[SOCRAT]{Structured-Output Causal Rationalizer}
\newacro{som}[SOM]{Self-Organizing Map}
\newacro{srae}[SRAE]{Sparse Reconstruction Autoencoder}
\newacro{stare}[STARE]{Statistics based Rule Extraction}
\newacro{splime}[SP-LIME]{Sub-Modular Pick \ac{lime}}
\newacro{svm}[SVM]{Support Vector Machine}
\newacro{tilm}[TILM]{Threshold-Rule Integer Linear Model}
\newacro{vec}[VEC]{Variable Effect Characteristic}
\newacro{via}[VIA]{Validity Interval Analysis}
\newacro {muse}[MUSE] {Model Understanding through Subspace Explanations}
\newacro{lore}[LORE]{LOcal Rule-Based Explanations}
\newacro{mcr}[MCR]{Model Class Reliance}
\newacro{shap}[SHAP]{SHapley Additive exPlanations}
\newacro{xai}[XAI]{Explainable Artificial Intelligence}
\newacro{sml}[SML]{Supervised Machine Learning}
\newacro{abs}[ABS]{Attention Based Summarization}
\newacro{iabm}[IABM]{Image Attention Based Model}
\newacro{sa}[SA]{Sensitivity Analysis}
\newacro{lrp}[LRP]{Layer-Wise Relevance Propagation}
\newacro{senn}[SENN]{Self-Explaining Neural Networks}
\newacro{vem}[VEM]{Visual Explanation Model}
\newacro{pj-x}[PJ-X]{Pointing and Justification-Based Explanation}
\newacro{icm}[ICM]{Image Captioning Model}
\newacro{ndt}[NDT]{Neural Decision Tree}
\newacro{dndt}[DNDT]{Deep Neural Decision Tree}
\newacro{mgm}[MGM]{Mind the Gap Model}
\newacro{aot}[AOT]{AND-OR Template}
\end{acronym}

\usetikzlibrary{topaths}
\usetikzlibrary{decorations.pathmorphing}
\usetikzlibrary{fit}
\usetikzlibrary{shapes.geometric}
\usetikzlibrary{backgrounds}	
\usetikzlibrary{arrows}
\usetikzlibrary{matrix,calc} 
\usetikzlibrary{patterns}

\tikzstyle{blockdef}=[rectangle, rounded corners, draw=black, thick, text centered]
\tikzstyle{rblock}=[blockdef, inner sep=4pt, text width=1.3cm]
\tikzstyle{dblock}=[shape=diamond, draw=black, thick, text centered, aspect=1.5]
\tikzstyle{cblock}=[blockdef, circle, inner sep=2pt, text width=2mm]
\tikzstyle{ablock}=[thick, -latex, black]
\tikzstyle{aablock}=[ultra thick, latex-latex, black]
\tikzstyle{fblock}=[draw=black, fill]
\tikzstyle{fitblock}=[blockdef, fill=none, dashed, inner sep = 1mm]

\tikzstyle{hlNone}=[anchor=base, rounded corners=3pt]
\tikzstyle{hlFrameNone}=[anchor=base, rounded corners=3pt, thick]
\tikzstyle{hlFrameBlue}=[hlFrameNone, fill=BlockBlue, draw=HilightBlue]
\tikzstyle{hlFrameRed}=[hlFrameNone, fill=BlockRed, draw=HilightRed]
\tikzstyle{hlBlue}=[hlNone, fill=BlockBlue]
\tikzstyle{hlGreen}=[hlNone, fill=BlockGreen]
\tikzstyle{hlRed}=[hlNone, fill=BlockRed]

\everymath{\displaystyle}

\usepackage{amssymb}
\usepackage{fancyvrb}
\usepackage{theorem}
\usepackage{dsfont}
\usepackage{adjustbox}
\usepackage{mathtools}
\usepackage{color, colortbl}
\usepackage{listings}
\definecolor{backcolour}{rgb}{0.95,0.95,0.92}
\definecolor{codegreen}{rgb}{0,0.6,0}
\definecolor{codegray}{rgb}{0.5,0.5,0.5}
\definecolor{codepurple}{rgb}{0.58,0,0.82}

\lstdefinestyle{myCustomStyle}{
  language=Tex,
   basicstyle=\tiny,
backgroundcolor=\color{backcolour},
  stepnumber=1,
  numbersep=12pt,
  basicstyle=\fontsize{11}{13}\selectfont\ttfamily
  tabsize=4,
  showspaces=false,
  showstringspaces=false
  breaklines=true
}
\newcommand{\eg}{, \textit{e}.\textit{g}., }
\addtokomafont{labelinglabel}{\sffamily}
\newacro{test}{Testlauf}

\usepackage[normalem]{ulem}
\usepackage{color}

\def\Fig#1{Figure~\ref{#1}}

\def\Sec#1{Section~\ref{#1}}

\def\vec#1{\mathbf{#1}}
\def\vx{\vec x}

\def\SD{\mathcal{D}}
\def\NewN{\ensuremath{\mathds{N}}}
\def\NewR{\ensuremath{\mathds{R}}}

\jmlrheading{X}{2020}{1-74}{6/19}{X/20}
\ShortHeadings{A Survey on the Explainability of Supervised Machine Learning}
{Burkart and Huber}
\firstpageno{1}

\begin{document}

\title{A Survey on the Explainability of Supervised Machine Learning}

\author{\name Nadia Burkart \email nadia.burkart@iosb.fraunhofer.de \\
      \addr Fraunhofer Institute for Optronics, System Technologies, and Image Exploitation IOSB, \\
      Interactive Analysis and Diagnostics,\\
      Fraunhoferstrasse 1, 76131 Karlsruhe, Germany
    \AND
      \name Marco F. Huber \email marco.huber@ieee.org \\
      \addr University of Stuttgart, Institute of Industrial Manufacturing and Management IFF\\
      \addr Fraunhofer Institute for Manufacturing Engineering and Automation IPA, \\Center for Cyber Cognitive Intelligence (CCI),\\
      Nobelstrasse 12, 70569 Stuttgart, Germany 
    }

\editor{}

\maketitle

\begin{abstract}
Predictions obtained by\eg artificial neural networks have a high accuracy but humans often perceive the models as black boxes. Insights about the decision making are mostly opaque for humans. Particularly understanding the decision making in highly sensitive areas such as healthcare or finance, is of paramount importance. The decision-making behind the black boxes requires it to be more transparent, accountable, and understandable for humans. 
This survey paper provides essential definitions, an overview of the different principles and methodologies of explainable \ac{sml}. We conduct a state-of-the-art survey that reviews past and recent explainable \ac{sml} approaches and classifies them according to the introduced definitions. Finally, we illustrate principles by means of an explanatory case study and discuss important future directions.
\end{abstract}

\begin{keywords}
 Explainability, Interpretability, Supervised Machine Learning, Automated decision-making, Black Box
\end{keywords}


\section{Introduction}
\label{sec:introduction}
The accuracy of current Artificial Intelligence (AI) models is remarkable but accuracy is not the only aspect that is of utmost importance. For sensitive domains, a detailed understanding of the model and the outputs is important as well. The underlying machine learning and deep learning algorithms construct complex models that are opaque for humans. 
\citet{holzinger2019causability} state that the medical domain is among the greatest challenges for AI. For areas such as health care, where a deep understanding of the AI application is crucial, the need for \ac{xai} is obvious. 

Explainability is important in many domains but not in all domains.
We already mentioned areas in which explainability is important such as health care.
In other domains such as  \emph{aircraft collision avoidance}, algorithms have been operating without human interaction without giving explanations for years.
Explainability is required when there is some degree of \textit{incompleteness}.
Incompleteness, to be sure, is not to be confused with uncertainty.
Uncertainty refers to something that can be formalized and handled by mathematical models.
Incompleteness, on the other hand, means that there is something about the problem that cannot be sufficiently encoded into the model (\citet{doshi2017towards}).
For instance, a criminal risk assessment tool should be unbiased and it also should conform to human notions of fairness and ethics.
But ethics is a broad field that is subjective and hard to formalize.
In contrast, airplane collision avoidance is a problem that is well understood and that can be described precisely.
If a system avoids collisions sufficiently well, there are no further concerns about it. No explanation is required.

Incompleteness can stem from various sources.
Another example are safety reasons. 
For a system that cannot be tested in a full deployment environment, there is a certain incompleteness with regards to whether the defined test environment is actually a suitable model for the real world.
The human desire for scientific understanding also adds incompleteness to the task.
The models can only learn to optimize their objective by means of correlation.
Yet, humans strive to discover causal dependencies.
All of the examples given contribute to a lack of understanding but sometimes, as these examples were supposed to illustrate as well, this may not bother us.

Automated decision-making can determine whether someone qualifies for certain insurance policies, it can determine which advertisement one sees, which job interviews one is being invited to, which university position one is being offered or what kind of medical treatment one will receive. 
If a loan is approved and everything accords with one's expectations, probably no one will ever ask for a detailed explanation. However, in case of a rejection, the reasons would be quiet interesting and helpful. 


In most cases, the models are complicated because the problem is complex and it is almost impossible to explain what exactly the models are doing and why they are doing it. Yoshua Bengio \citep{bengio2017}, a pioneer in the field of deep learning research, said ``As soon as you have a complicated enough machine, it becomes almost impossible to completely explain what it does.'' Jason Yosinski from Uber states \citep{voosen2017} ``We build amazing models. But we don't quite understand them. Every year this gap is going to get a little bit larger.'' Paul Voosen \citep{voosen2017} gets to the heart of the issue and questions ``why, model, why?.'' 


According to \citet{lipton2016mythos}, explainability is demanded whenever the goal the prediction model was constructed for differs from the actual usage when the model is being deployed. In other words, the need for explainability arises due to the mismatch between what a model can explain and what a decision maker wants to know. 
Explainability issues also arises for well-functioning models that fail on few data instances. Here, we also demand explanations for why the model did not perform well on these few feature combinations \citep{kabra2015understanding}. 
According to \citet{martens2009decompositional}, explainability is essential whenever a model needs to be validated before it can be implemented and deployed. Domains that demand explainability are characterized by making critical decisions that involve, for example, a human life\eg in health care. 

The renewed EU General Data Protection Regulation (GDPR) could require AI providers to provide users with explanations of the results of automated decision-making based on their personal data. Personal data is defined as information relating to an identified or identifiable natural person \citep{europa}. The GDPR replaced the Data Protection Directive from 1995. This new requirement affects large parts of the industry. The European Parliament has revised regulations that concern the collection, storage, and usage of personal information. The GDPR may make complicated or even lead to the prohibition of the use of opaque models that are used for certain applications, e.g., for recommender systems that work based on personal data.
\citet{goodman} call this the right of explanation for each subject (person). This will most likely affect financial institutions, social networks and the health care industry. Automated decision-making used by financial institutions for monitoring credit risk or money laundering needs to be transparent, interpretable and accountable. 

In January 2017, the Association for Computing Machinery \citep{acm} released a statement on algorithmic transparency and accountability. In this statement, the ACM points out that the usage of algorithms for automated decision-making can result in harmful discrimination. To avoid those problems, the ACM also published a list of certain principles to follow. In May 2017, the DARPA launched the program \ac{xai} \citep{gunning2017explainable} which aims at providing explainable and highly accurate models. \ac{xai} is an umbrella term for any research trying to solve the black-box problem for AI. Since there are a lot of different approaches for solving this problem, each with their own individual needs and goals, there is no single common definition of the term \ac{xai}. The key idea, however, is to enable users to understand the decision-making of any model. 

The words understanding, interpreting and explaining are often used interchangeably when used in the context of explainable AI \citep{doran2017does}. Usually, \emph{interpretability} is used in terms of comprehending how the prediction model works as a whole. \emph{Explainability}, in contrast, is often used when explanations are given by prediction models that are incomprehensible themselves.

Explainability and interpretability are also important aspects for deep learning models, where a decision depends on an enormous amount of weights and parameters. Here, the parameters are often abstract and disconnected from the real world, which makes it difficult to interpret and explain the results of deep models \citep{angelov2019towards}. \citet{samek2017explainable} describe different methods for visualizing and explaining deep learning models like the \ac{sa} or the \ac{lrp}. \citet{biswas2017rule} or \citet{zilke2016deepred} also mention different techniques for decompositional decision rules from ANNs.
Since the focus of this paper lies on explanations for \ac{sml} mainly for tabular data, only a few explanation methods for deep learning models are mentioned.


As the research field of \ac{xai} grows rapidly, there already exist a few survey papers that gather existing approaches. \citet{guidotti2018} provide a survey on approaches for explaining black box models. The authors provide a taxonomy for the approaches that is not directly linked to one certain learning problem. \citet{molnar2018guide} describes several approaches for generating explanations and interpretable models. 
He introduces different data sets and thereby describes some approaches in the main field of interpretable models, model agnostic methods and example-based explanations.
\citet{adadi2018} provide an overview of general \ac{xai} research contributions by addressing different perspectives in a non-technical overview of the key aspects in \ac{xai} and various approaches that are related to \ac{xai}. 
\citet{biran2017explanation} provides an overview of explanations and justifications for different machine learning models like Bayesian networks, recommender systems and other adjacent areas. Explaining a machine learning model means to render the output of a model understandable to a human being. Justification can be generated for a black box model and describes why the generated decision is a meaningful one \citep{biran2017explanation}. 
\citet{montavon2018methods} offers various techniques for interpreting individual outputs from deep neural networks, focusing on the conceptual aspects that make these interpretations useful in practice. 
\citet{gilpin2018explaining} defines the terms \emph{interpretability} and {explainability} and reveals their differences. A new taxonomy which offers different explanation possibilities for machine learning models is introduced. This taxonomy can be used to explain the treatment of features by a network, the representation of features within a network or the architecture of a network. 
Unlike many other papers on explainable machine learning, the paper by \citet{dovsilovic2018explainable} focuses on explanations and interpretations of supervised machine learning methods. The paper describes the \textit{integrated (transparency-based)} and the \textit{post hoc} methods and offers a discussion about the topic of explainable machine learning. 
The paper by \citet{tjoa2019survey} gives a broad overview of interpretation approaches and classifies them. It focuses on machine interpretation in the medical field and discloses the complexity of interpreting the decision of a black box model.

This survey paper provides a detailed introduction to the topic of explainable \ac{sml} regarding the definitions and a foundation for classifying the various approaches in the field. We distinguish between the various problem definitions to categorize the field of explainable supervised learning into interpretable models, surrogate model fitting and explanation generation. The definition of interpretable models focuses on the entire model understanding that is achieved either naturally or by using design principles to force it. The surrogate model fitting approach approximates local or global interpretable models based on a black box. The explanation generation process that directly produces a kind of explanation distinguishes between local and global explainability.

In summary, the paper offers the following contributions:

\begin{itemize}
    \item formalization of five different explanation approaches and a review of the corresponding literature (classification and regression) for the entire explanation chain
    \item reasons for explainability, review of important domains and the assessment of explainability
    \item a chapter that solely highlights various aspects around the topic of data and explainability such as data quality and ontologies
    \item a continuous use case that supports the understanding of the different explanation approaches
    \item a review of important future directions and a discussion.
\end{itemize}


\section{Reasons for Explainability and Demanding Domains}
In this chapter, we will describe reasons for explainability and introduce example domains where XAI is needed.

\subsection{Reasons for Explainability}
Automated decision-making systems are not widely accepted. Humans want to understand a decision or at least they want to get an explanation for certain decisions. This is due to the fact that humans do not trust blindly. Trust, then, is one of the motivating aspects of explainability. Other motivating aspects are causality, transferability, informativeness, fair and ethical decision-making \citep{lipton2016mythos}, accountability, making adjustments and proxy functionality.

\emph{\textbf{Trust:}}
Trust and acceptance of the prediction model are needed for the prediction model's deployment. Understanding and knowing the prediction model's strengths and weaknesses is a prerequisite for human trust and, hence, for model deployment.

\emph{\textbf{Causality:}}
Explainability, e.g. in the form of attribute importance, conveys a sense of causality to the system's target group. This concept of causality can only be grasped when the system points out the underlying input-output relationship.

\emph{\textbf{Transferability:}}
The prediction model needs to convey an understanding of future behavior for a human decision-maker in order to use the prediction model with unseen data. Only when the decision-maker knows that the model generalizes well or when he knows in which context it generalizes well, the prediction model will be put in charge of making decisions.

\emph{\textbf{Informativeness:}}
In order to be deployed as a system, it is necessary to know whether the system actually serves the real world purposes it is designed for instead of merely serving the purposes it was trained for. If this information is given, the system can be deployed.

\emph{\textbf{Fair and Ethical Decision Making:}}
Knowing the reasons for a certain decision is a societal need and most likely it will be an official right for EU-citizens \citep{goodman2016eu}. This right to explanation requires decision-makers to present their results in a comprehensible way in order to perceive conformity to ethical standards. Each person that is affected by an automated decision can make use of this right to explanation.

\emph{\textbf{Accountability:}}
One goal of incorporating explainability into the decision-making process is to make an algorithm accountable for its actions. In order for a system to be accountable, it has to be able to explain and justify its decisions. Furthermore, the data-shift problem can be targeted with interpretable systems, making these more accountable for their actions \citep{freitas2014comprehensible}.

\emph{\textbf{Making Adjustments:}}
Understanding the prediction model and the underlying factors enables domain experts to compare the prediction model to the existing domain knowledge. Explainability is a prerequisite for the ability to adjust the prediction model by incorporating domain knowledge. According to \citet{selvaraju2017visual}, explainability of prediction models can teach humans, especially domain experts using these prediction models, how to make better decisions. Furthermore, when looked at from an algorithmic point of view, explainability enables system designers to make changes to the prediction model by, e.g., adjusting parameters. Explainability is also useful for developers since it can be used to identify failure modes.

\emph{\textbf{Proxy Functionality:}}
When explainability is provided by a system, it can also be examined based on other criteria that cannot be easily quantified such as safety, non-discrimination, privacy, robustness, reliability, usability, fairness, verification and causality \citep{doshi2017towards}. In this case explainability serves as a proxy.

\subsection{Domains demanding Explainability}
As already mentioned at the outset, explainability is not required for every domain. There are domains that use black box prediction models since these domains are either well studied and users trust the existing models or because no direct consequences threaten in case the system makes mistakes \citep{doshi2017towards} \eg in recommendation systems for marketing . According to \citet{lipton2016mythos}, explainability is demanded whenever the goal the prediction model was constructed for differs from the actual usage the model is being deployed for. The need for explainability, then, arises due to a mismatch between what a model can explain and what a decision-maker wants to know. According to \citet{martens2009decompositional}, explainability is important whenever a model needs to be validated before it can be implemented and deployed. Domains that demand explainability are characterized by making critical decisions that involve, e.g., human lives (medicine/healthcare) or a lot of money (banking/finance) \citep{strumbelj2010explanation}. In what follows, we take a closer look at relevant domains and provide a motivating example for the necessity of explainability in these specific domains.

\emph{\textbf{Medical Domain/Health-Care:}} When a medical researcher uses an intelligible system for screening patients with a high risk for cancer, it is not sufficient to identify patients with a high risk in an accurate manner;  but he also understand causes of cancer \citep{henelius2014peek}.

\textbf{Judicial System:} In order to defend an automated decision in court, it is necessary to understand the reasons for a specific prediction \citep{freitas2014comprehensible}.

\emph{\textbf{Banking/Financial Domain:}} It is a legal obligation to be able to explain why a customer was denied a credit \citep{freitas2014comprehensible}. Furthermore, it is of great interest for banks and insurance companies to predict and understand customer churn to be able to develop a reasoned counteracting plan (due to high costs of seeking new customers) \citep{verbeke2011building}.

\emph{\textbf{Bio-informatics:}} If trust can be established in a system, more time and money will be invested in experiments regarding the system's domain  according to \citet{freitas2014comprehensible} and \citet{subianto2007understanding}.

\emph{\textbf{Automobile Industry:}} If there is an autonomously driving car involved in an accident, it is of great interest to the developer, to the people involved and to the legal system to understand the reasons why the accident happened in order to fix the system and to sue the person responsible for the accident.

\emph{\textbf{Marketing:}} Marketing is mainly concerned about distributing the products of a company. A company is in a better position than another company if it can explain why a customer preferred one product over another one since this information can be used, e.g., to equip other products with purchase-relevant attributes.

\emph{\textbf{Election campaigns:}} Similar to customer churn, votes in an election can be influenced if the reasons for voting decisions are better understood. In an election campaign, voters can be targeted with coordinated advertising based on their personal interests. 

\emph{\textbf{Precision Agriculture:}} Due to the use of remote sensors, satellites, and UAVs, information regarding a particular area can be gathered. Through the gathered data and machine learning models, farmers can develop a better understanding of what they need to do in order to increase the harvest benefits \citep{agriculture2017}.

\emph{\textbf{Expert Systems for the Military:}} The military can make use of expert systems, e.g., in the context of training soldiers. In a military simulation environment, a user has to accomplish a certain goal. With the help of explainable machine learning, the user receives meaningful information on how to accomplish the goal more efficiently \citep{van2004explainable}. 

\emph{\textbf{Recommender systems:}} Explainable recommendations help system designers to understand why a recommender system offers a particular product to a particular user group. It helps to improve the effectiveness of a recommender system and the clarity of a decision \citep{zhang2018explainable}.

\section{Concepts of Explainability}
\label{sec:explain}

A variety of different approaches for explaining learned decisions have been proposed \citep{molnar2018guide}. Some try to explain the model as a whole or completely replace it with an inherently understandable model such as a decision tree \citep{freitas2014comprehensible}. Other approaches try to steer the model in the learning process to a more explainable state \citep{schaaf2019enhancing, burkart2019forcing} or focus on just explaining single predictions for example by highlighting important features \citep{Ribeiro2016} or contrasting it to another decision \citep{wachter2017counterfactual}. In the following sections, we structure the area of explainable supervised machine learning. First, we describe the problem definitions and dimensions. Next, we introduce interpretable model types and techniques. Finally, the explanation itself is described.


\subsection{Problem Definition and Dimensions}
\label{sec:problem}
In this subsection, we will give an overview of the problem definitions and the according dimensions in \ac{sml}. This creates the basis for categorizing the procedures. 

\ac{sml} aims at learning a so-called \emph{model} $h(\vx)=y$ that maps a feature vector $\vx \in \SX \subseteq \NewR^d$ to a target $y\in \SY \subseteq \NewR$. For this purpose, a set of training data $\SD = \{(\vx_1, y_1), \ldots, (\vx_n, y_n)\}$ is used for the learning process of the model. \ac{sml} can be divided into the tasks of \emph{classification} and \emph{regression}. For classification, the target $y$ is a discrete value often called a \emph{label}. For instance, if $y\in \{0,1\}$ or $y\in \{-1,1\}$ one speaks about binary classification. The task of regression is to predict a continuous target value $y \in \NewR$. 

In this paper, we differentiate between two different types of models. A model can either be a black box $b:\SX \rightarrow \SY$, $b(\vx)=y$ with $b \in \SB$, where $\SB \subset \SH$ is the hypothesis space of black box models, e.g., $\SB = \{\text{neural networks with one hidden layer}\}$, Or the model can be an interpretable one $w:\SX \rightarrow \SY$, $w(\vx)=y$ with $w \in \SI$, where $\SI \subset \SH$ is the hypothesis space of interpretable models, e.g., $\SI = \{\text{decision trees of depth 3}\}$.

\begin{figure*}
    \centering
    \begin{tikzpicture}[scale=0.95]
        \node[rblock, text width=1.8cm, minimum height=1.2cm] at (0,0) (A) {\small Learning Algorithm};
        \node[rblock, text width=1.8cm, minimum height=1.2cm] at (3,0) (M) {\small Model $h$};
        \node[rblock, text width=1.8cm, minimum height=1.2cm] at (6,0) (P) {\small Prediction};
        \node at ($(A.west)+(-1.3,0)$) {(a)};
        
        \draw[ablock] ($(A.north)+(-1.6,.6)$) |- (A.west) node[midway] (tr) {} node[pos=.03, above] {\small Training Data};
        \draw[ablock] ($(M.north)+(-1.6,.6)$) -- +(0,-.2) node[at start, above] {\small Test Data} -| (M.north) node[at start] (te) {};
        \draw[ablock] (A) -- (M);
        \draw[ablock] (M) -- (P);
        \filldraw (tr) circle (.8mm);
        \filldraw (te) circle (.8mm);
        
        \draw[dashed] (-2.7,-1) -- (13,-1);
        
        \node[rblock, text width=1.8cm, minimum height=1.2cm] at (0,-2) (A) {\small Learning Algorithm};
        \node[rblock, text width=1.8cm, minimum height=1.2cm] at (3,-2) (M) {\small Black Box Model $b$};
        \node[rblock, text width=1.85cm, minimum height=1.2cm] at (9,-2) (E) {\small Explanation $e$};
        \node at ($(A.west)+(-1.3,0)$) {(b)};
        
        \draw[ablock, shorten < = -1mm] (tr) |- (A.west) node[midway] (tr) {};
        \draw[ablock] (A) -- (M);
        \draw[ablock] (M) -- (E);
        \filldraw (tr) circle (.8mm);
        
        \node[rblock, text width=1.8cm, minimum height=1.2cm] at (0,-4) (A) {\small Learning Algorithm};
        \node[rblock, text width=1.8cm, minimum height=1.2cm] at (3,-4) (M) {\small White Box Model $w$};
        \node[rblock, text width=1.85cm, minimum height=1.2cm] at (9,-4) (E) {\small Explanation $e$};
        \node at ($(A.west)+(-1.3,0)$) {(c)};
        
        \draw[ablock, shorten < = -1mm] (tr) |- (A.west) node[midway] (tr) {};
        \draw[ablock] (A) -- (M);
        \draw[ablock] (M) -- (E);
        \filldraw (tr) circle (.8mm);
        
        \node[rblock, text width=1.8cm, minimum height=1.2cm] at (0,-6) (A) {\small Learning Algorithm};
        \node[rblock, text width=1.8cm, minimum height=1.2cm] at (3,-6) (M) {\small Black Box Model $b$};
        \node[rblock, text width=1.8cm, minimum height=1.2cm] at (6,-6) (S) {\small Surrogate $w$};
        \node[rblock, text width=1.85cm, minimum height=1.2cm] at (9,-6) (E) {\small Explanation $e$};
        \node at ($(A.west)+(-1.3,0)$) {(d)};
        
        \draw[ablock, shorten < = -1mm] (te) -- +(0,-6) node[at end] (te) {} -| (M.north) node[midway] (tmp) {};
        \draw[ablock, shorten < = -1mm] (tmp) -| (S.north);
        \draw[ablock, shorten < = -1mm] (tr) |- (A.west) node[midway] (tr) {};
        \draw[ablock] (A) -- (M);
        \draw[ablock] (M) -- (S);
        \draw[ablock] (S) -- (E);
        \filldraw (tr) circle (.8mm);
        \filldraw (te) circle (.8mm);
        \filldraw (tmp) circle (.8mm);
        
        \node[rblock, text width=1.8cm, minimum height=1.2cm] at (0,-8) (A) {\small Learning Algorithm};
        \node[rblock, text width=1.8cm, minimum height=1.2cm] at (3,-8) (M) {\small  Model $h$};
        \node[rblock, text width=1.8cm, minimum height=1.2cm] at (6,-8) (P) {\small Prediction};
        \node[rblock, text width=1.85cm, minimum height=1.2cm] at (9,-8) (E) {\small Explanation $e$};
        \node at ($(A.west)+(-1.3,0)$) {(e)};
        
        \draw[ablock, shorten < = -1mm] (te) -- +(0,-2) node[at end] (te) {} -| (M.north);
        \draw[ablock, shorten < = -1mm] (tr) |- (A.west) node[midway] (tr) {};
        \draw[ablock] (A) -- (M);
        \draw[ablock] (M) -- (P);
        \draw[ablock] (P) -- (E);
        \filldraw (tr) circle (.8mm);
        \filldraw (te) circle (.8mm);
        
        \node[rblock, text width=1.8cm, minimum height=1.2cm] at (0,-10) (A) {\small Learning Algorithm};
        \node[rblock, text width=1.8cm, minimum height=1.2cm] at (3,-10) (M) {\small Black Box Model $b$};
        \node[rblock, text width=1.8cm, minimum height=1.2cm] at (6,-10) (P) {\small Prediction};
        \node[rblock, text width=1.8cm, minimum height=1.2cm] at (9,-10) (S) {\small Surrogate $w$};
        \node[rblock, text width=1.85cm, minimum height=1.2cm] at (12,-10) (E) {\small Explanation $e$};
        \node at ($(A.west)+(-1.3,0)$) {(f)};
        
        \draw[ablock, shorten < = -1mm] (te) -- +(0,-2) -| (S.north);
        \draw[ablock, latex-] (M.north) -- +(0,.4) node (te) {};
        \draw[ablock, shorten < = -1mm] (tr) |- (A.west) node[midway] (tr) {};
        \draw[ablock] (A) -- (M);
        \draw[ablock] (M) -- (P);
        \draw[ablock] (P) -- (S);
        \draw[ablock] (S) -- (E);
        \filldraw (tr) circle (.8mm);
        \filldraw (te) circle (.8mm);
    \end{tikzpicture}
    \caption{From an opaque supervised model to an explanation. (a) Standard supervised machine learning without explanation. (b)-(d) Model/global explanations: (b) post-hoc black box model explanation, (c) interpretable by nature, i.e., white box model explanation, and (d) explaining a black box model by means of a global surrogate model. (e)-(f) Instance/local explanations: (e) directly or (f) with a local surrogate.}
    \label{fig:relations}
\end{figure*}
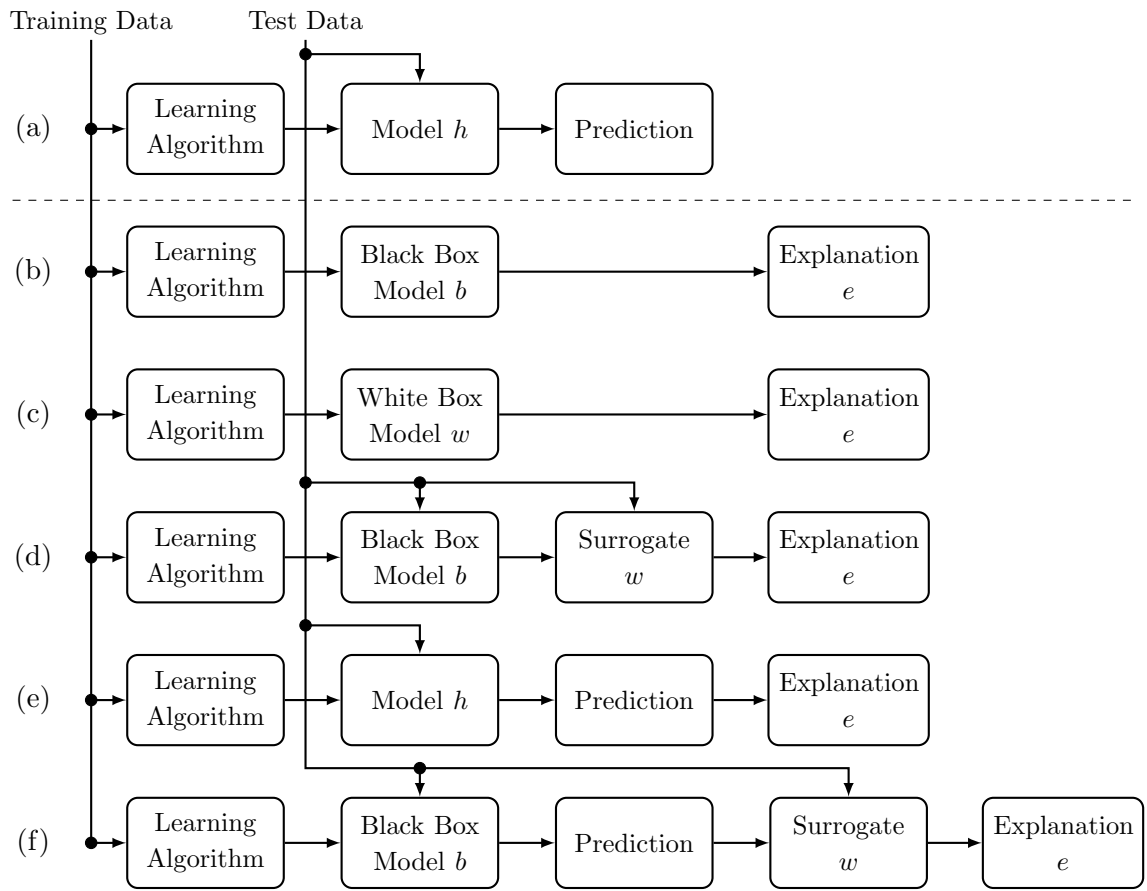

To evaluate the prediction performance of the trained model, we use the \emph{error measure} $S:\SY \times \SY \rightarrow \NewR$. A common example of an error measure in binary classification with $y\in \{-1,1\}$ is the \emph{hinge loss} $S(h(\vx), y) = \max\{0, 1- h(\vx) \cdot y\}$, which is zero when the true label $y$ and the prediction $h(\vx)$ are identical. In regression problems, the \emph{squared deviation} $S(h(\vx), y) = (h(\vx) - y)^2$ is a common error measure. Given an error measure, \ac{sml} can be formulated as an optimization problem. 

\begin{Problem}[Supervised Machine Learning]
\label{prob:sml}
Given training data $\SD$, the SML aims for solving the optimization problem 
	\begin{equation}
		\label{eq:sml}
		h^* = \argmin{h\in \SH}\ \frac{1}{n}\sum_{i=1}^n S\big(h(\vx_i), y_i\big)~,
	\end{equation}
	where the averaged error over all training instances is minimized and where 	$h^*$ is the resulting model.
\end{Problem}

In case of a parametric model $h(\vx;\vthe)$ with parameter vector $\vthe$ as in a neural network, \eqref{eq:sml} is equivalently formulated as 
\begin{equation}
		\label{eq:smlparametric}
		\vthe^* = \argmin{\vthe}\ \frac{1}{n}\sum_{i=1}^n S\big(h(\vx_i;\vthe), y_i\big)~.
\end{equation}
In many cases, the optimization problem \eqref{eq:sml} or \eqref{eq:smlparametric} cannot be solved analytically exactly. One of the rare exceptions where this is possible is linear regression. Thus, it is common to end up with a sub-optimal solution that is found numerically, as is the case when deploying deep neural networks where the parameter vector $\vthe$ is determined by means of gradient descent. 

It is important to differentiate between the \emph{learning algorithm} and the SML model: The learning algorithm tries to solve the optimization problem \eqref{eq:sml}, either directly or implicitly. The result of the optimization problem and, thus, the output of the learning algorithm, is the actual model, which then can be applied to an unseen data instance $\vx$ in order to obtain a prediction $y=h^*(\vx)$. \Fig{fig:relations}(a) illustrates this SML learning pipeline.

\subsection{Five Ways to Gain Interpretability}
\label{sec:explain_ways}
Solving Problem~\ref{prob:sml} usually leads to black box models requiring further processing in order to obtain the desired explanations. In what follows, we define five interpretable SML problems by modifying or extending \eqref{eq:sml}. In general, these problems can be grouped into \emph{model explanation} approaches (cf. \Fig{fig:relations}(b)--(d)) and \emph{instance explanation} approaches (cf. \Fig{fig:relations}(e)--(f)). Model explanation approaches generate the explanations on the basis of an SML model trained on $\SD$ and aim at providing insights about the inner workings of the entire model. 
In contrast, instance explanation approaches merely try to explain a model prediction $y$ for a single data instance $\vx$. Thus, generated explanations are only valid for $\vx$ and its close vicinity. Model and instance explanation approaches are also referred to as \emph{global} and \emph{local} explanation approaches, respectively. 

In what follows, the different ways to gain interpretability are defined formally. In addition, we will briefly mention some well-known examples.

\subsubsection{Explanation Generation}
\label{sec:explain_ways_explanations}
We first need to define the explanation generation process itself. 
While learning algorithms provide the SML model and the model, in turn, provides predictions given unseen data instances, the so-called \emph{explanator} is concerned with deriving a human-comprehensible explanation for the model and/or the prediction. 
Thus, in any case some explanator is mandatory to gain understanding of the model used or the prediction obtained. As indicated in \Fig{fig:relations}(c), even interpretable models\footnote{The terms ``white box model'' and ``interpretable model'' are used synonymously throughout this paper.} need an additional component that extracts an explanation from the model. 
Here, an explanator provides additional means for the model's comprehensibility such as feature statistics, feature importance or visualizations. 
In turn, explanators are no predictors. They always rely on a learned model.


\begin{Problem}[Explanation Generation]
	An explanator function $e$ is defined as
	\begin{equation}
		\label{eq:explanation}
		e: (\SX \rightarrow \SY) \times (\SX\times\SY) \rightarrow \SE
	\end{equation}
	which takes an SML model (black box or interpretable) and a specific data set as input and provides an explanation belonging to the set $\SE$ of all possible explanations as an output. There are two possible explanation generation problems:
    \begin{description}
        \item[Global] Extracting a global explanation from a model that is representative for some specific data set $\SD'$, i.e., $e(b, \SD')$ in case of a black box model (see \Fig{fig:relations}(b)) or $e(w, \SD')$ for interpretable models (see \Fig{fig:relations}(c) and (d)).  
		\item[Local] Instance explanators extract an explanation for a single test input $\vx$ and the corresponding prediction $y$, i.e., $e(b, (\vx, y))$ or $e(w, (\vx, y))$.  
	\end{description}
\end{Problem}


Given this definition, we can introduce the first type of explanation of an SML model, namely the direct interpretation of a given black box model in a \emph{post-hoc} fashion as depicted in \Fig{fig:relations}(b). This is achieved by means of global explanators, often being model-agnostic. A well-known example are partial dependency plots (cf. \Sec{sec:explanations_global_feature} and \citep{goldstein2015peeking}). 

Global explanators in general require no predictions and solely rely on the learned model (black box or interpretable) and some set of feature vectors---often the training data itself. Thus, they are sometimes also referred to as \emph{static} explanators. 
In some cases, not even the data set $\SD'$ is required to generate an explanation from a learned model, i.e., it follows $e(b, \varnothing)$ and $e(w, \varnothing)$. A corresponding example are the attribute weights extracted from linear regression models. 

Local explanators allow for the second way of generating explanations as depicted in \Fig{fig:relations}(e). Given a prediction of a model (black box or interpretable), the local explanator provides insights that are only valid for the particular instance and cannot be generalized for the entire model. For this reason, they are often also named \emph{dynamic} or \emph{output} explanation generation. Examples are
counterfactual explanations (see Section~\ref{sec:explanation_local_counterfacturals}), Shapley values (cf. \Sec{sec:explanations_local_saliency} and \citep{vstrumbelj2014explaining}), or decision paths when classifying $\vx$ with a decision tree.

\subsubsection{Learning Interpretable Models}
\label{sec:explain_ways_whitebox}
Next, we consider models that are interpretable on their own, i.e., models that are easily comprehensible for humans.

\begin{Problem}[White Box Model Learning]
\label{prob:interpretable}
Modifying Problem~\ref{prob:sml} leads to the optimization problem
\begin{equation}
		\label{eq:interprete}
		w^* = \argmin{w \in \SI}\ \frac{1}{n}\sum_{i=1}^n S\big(w(\vx_i), y_i\big)~,
	\end{equation}
	with $(\vx_i, y_i) \in \SD$. 
\end{Problem}
That is, by solving Problem \ref{prob:interpretable}, one aims for learning a white box model from the hypothesis space of interpretable models $\SI$, which is the third way of gaining explainability. This corresponds to the pipeline depicted in \Fig{fig:relations}(c) and is often also called \emph{ante-hoc} interpretability. Typical examples are learning small decision trees or using linear models for regression problems.

\subsubsection{Surrogate Model Learning}
\label{sec:explain_ways_surrogate}
While using interpretable models might be appropriate for some learning problems, they come at the cost of flexibility, accuracy, and usability according to \citep{ribeiro2016model}. Here, so-called \emph{post-hoc} interpretability can be of help, where the black box model is still used for predictions and, thus, one can rely on the potentially high accuracy of this model. In addition, a white box surrogate model of the black box model is generated to gain interpretability. In what follows, the remaining two ways for obtaining a surrogate model are introduced. 

\begin{Problem}[Surrogate Model Fitting]
\label{prob:surrogate}
Surrogate Model Fitting is the process of translating a black box model into an approximate interpretable model by solving
\begin{align}
	\label{eq:surrogate}
	w^* = \argmin{w \in \SI}&\ \frac{1}{|\SX|}\sum\nolimits_{\vx\in\SX} S\big(w(\vx), b(\vx)\big)~.
\end{align}
Here, we differentiate between two different kinds of surrogate models. 
\begin{description}
    \item[Global] The surrogate model $w$ approximates the black box model $b$ on the whole training data, i.e., $\SX = \{\vx_1,\ldots, \vx_n\}$ is taken from the training data set $\SD$. Alternatively, $\SX$ is a sample data set which represents the input data distribution of the model $b$ sufficiently well.
    \item[Local] The surrogate model $w$ approximates the black box model around a test input $\vx$ defined as $\SX = \{\vx'| \vx' \in N(\vx)\}$, where $N$ is some neighborhood of $\vx$. 
\end{description}
\end{Problem}

It is worth mentioning that in \eqref{eq:surrogate}, the measure $S$ acts as a \emph{fidelity} score, i.e., it quantifies how well the surrogate $w$ approximates or agrees with the black box model $b$. 

Global and local surrogates are reflected in \Fig{fig:relations}(d) and (f), respectively. 
The global surrogate tries to simulate all predictions of the black box model with high accuracy, which makes it possible to understand the black box.
A simple example of a global surrogate model is to learn a decision tree on a data set $\SD' = \{(\vx_1, b(\vx_1)), \ldots, (\vx_n, b(\vx_n))\}$, i.e., the surrogate is trained on the predictions $b(\vx_i)$ of the black box model, where $\vx_i$ are the feature vectors from the training data.

A local surrogate is only valid near the current prediction. Therefore, only a local understanding of the black box model can be gained. 
Well-known representatives of the class of local surrogates are LIME \citep{ribeiro2016should} or SHAP \citep{lundberg2017unified}.

\begin{figure*}[t]
    \centering
    \begin{tikzpicture}
        \node[rblock, text width=2cm, minimum height=1.7cm] at (0,0) (sml) {Model Learning};
        \node[rblock, text width=2cm, minimum height=1.7cm] at (5,0) (surrogate) {Surrogate Model Fitting};
        \node[rblock, text width=2cm, minimum height=1.7cm] at (10,0) (explain) {Explanation Generation};
        
        \filldraw ($(surrogate.west)+(-.5,0)$) circle (.8mm) node (circ) {};
        \filldraw ($(surrogate.east)+(.5,0)$) circle (.8mm) node (circ2) {};
        \draw[ablock, dashed] (circ) -- +(0,1.5) -| (explain.north) node[near start, above] {post-hoc};
        \draw[ablock, dashed] (sml) -- +(0,-1.5) -| (circ2) node[near start, above] {ante-hoc};
        \draw[ablock, latex-] (sml.west) -- +(-.5,0) node[left] {\rotatebox{90}{Data}};
        \draw[ablock] (sml) -- (surrogate) node[pos=.4, above=-2] {black box} node[pos=.4, below=-2] {model};
        \draw[ablock] (surrogate) -- (explain) node[pos=.6, above=-2] {interpret.} node[pos=.6, below=-2] {model};
        \draw[ablock] (explain.east) -- +(.5,0) node[right] {\rotatebox{90}{Explanation}};
        
    \end{tikzpicture}
    \caption{Summary of model/global explanation approaches. Strictly following the solid arrows corresponds to global surrogate fitting as depicted in \protect\Fig{fig:relations}(d). The ante-hoc path corresponds to learning an interpretable/white box model (cf. \protect\Fig{fig:relations}(c)), while the post-hoc path aims for black box model explanation (cf. \protect\Fig{fig:relations}(b)).}
    \label{fig:model-explainability}
\end{figure*}
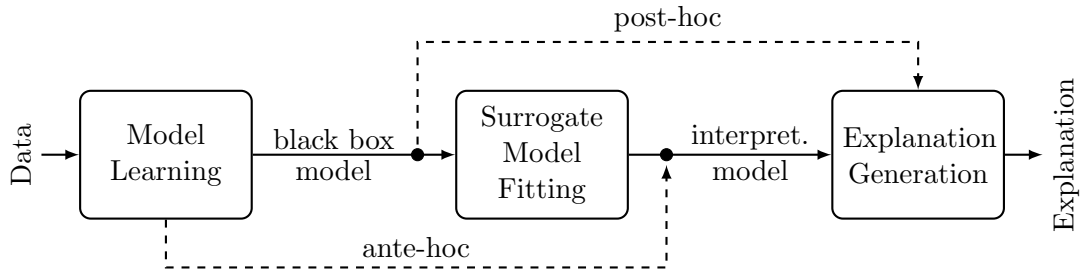

\subsubsection{Summary}
\label{sec:explain_ways_summary}
Table \ref{DimSML} summarizes various dimensions regarding the problem definitions in the field of explainable SML:
\begin{description}[leftmargin=10em,style=nextline, align=right]
    \item[ante-hoc] Interpretability is built-in from the beginning of the model creation
    \item[post-hoc] Interpretability is created after model creation
    \item[instance/local] Interpretability only holds locally for a single data instance and its close vicinity
    \item[model/global] Interpretability holds globally for the entire model
    \item[specific] The mechanism for gaining interpretability works only for a specific model class
    \item[agnostic] The mechanism for gaining interpretability is generally applicable for many or even all model classes
    \item[data independent] The mechanism for gaining interpretability works without additional data
    \item[data dependent] The mechanism for gaining interpretability requires data
\end{description}

Interpretable/white box models are inherently ante-hoc and specific. There is no agnostic interpretable model learning approach. One aim behind using an interpretable model is to have build-in (by nature) model explainability. Local interpretability is merely possible when combined with a local explanation approach. The data independence/dependence dimensions are not appropriate for white box models since they alone cannot provide an explanation. 

Surrogate models are created in a post-hoc fashion from a given black box model and describe a certain instance or the entire model (see Problem~\ref{prob:surrogate}). The surrogate model can be applied either to a specific model class or it can be agnostic for many classes of models. In order to create a surrogate model, it is inevitable to rely on data, at least to measure and optimize the fidelity between surrogate and black box model or to create a local surrogate for a single data instance.

We can see that interpretable model learning and surrogate model fitting are mutually exclusive when it comes to ante-hoc or post-hoc explainability. This is emphasized in \Fig{fig:model-explainability}.

\begin{table}[tb]
\caption{Dimensions of explainable supervised learning. \cmark\ indicates fulfillment of the dimension, while \xmark\ states that the dimension is not applicable. Further, -- indicates that the dimension is not appropriate.}

\begin{adjustbox}{width=\linewidth}
 \begin{tabular}{| l || c | c | c| c| }
    \hline
 \textbf{Dimensions} &  \textbf{Interpretable Models} &  \textbf{Surrogate Models}& \textbf{Explanation Generation} \\ \hline \hline
  ante-hoc & \cmark  & \xmark & \xmark  \\
   post-hoc   & \xmark  & \cmark  &\cmark \\ 
 instance/local & \xmark & \cmark & \cmark \\
    model/global & \cmark & \cmark & \cmark  \\
      specific & \cmark & \cmark & \cmark  \\
       agnostic & \xmark & \cmark & \cmark  \\
       data independent & -- & \xmark & \cmark \\
       data dependent & -- & \cmark & \cmark \\
    
    \hline
    \end{tabular}
    \end{adjustbox}
    
    \label{DimSML}
\end{table}

As is the case with fitting surrogate models, the generation of an explanation is strictly post-hoc as well. Besides that, for both it is possible to find approaches that are used in instance or model explanations that are model specific or model agnostic and that are data independent---and thus global---or data dependent.


\subsection{Interpretable Model Types}
\label{sec:explain_models}
According to Section~\ref{sec:explain_ways}, some models belong to the hypothesis space $\SI$. These interpretable models provide interpretability by themselves and, thus, can be considered ``interpretable by nature''. The models are understood in their entirety by the model's target group (see Section~\ref{sec:target group}). The following incomplete list names some interpretable models:

\emph{Linear models:} Linear models consist of input features and weights for each of them. The models are built by\eg logistic regression in such a way as to assign a weight to every feature used by the model. The assigned weight indicates the feature's contribution to the prediction. Linear models are also easily adopted to yield a continuous value instead of a discrete class label; hence, they are useful for regression. A special group of linear applications are scoring systems. Scoring systems assign each feature or feature interval a specific weight. A final score is evaluated in the same way as the prediction of a linear model is. However, an instance is classified after comparing the final score with a defined threshold. 

\emph{Decision trees:} Decision trees are trained on the labeled input features. The constructed trees consist of tree nodes and leaf nodes. For an exemplary decision tree see \Fig{fig:tree2} and \ref{fig:tree3}. Tree nodes are assigned a splitting feature and a splitting value. Leaf nodes are assigned a class label in case of classification and the averaged value in case of regression. The process for a test instance $\vx$ starts at the top tree node---the root node---and traverses downwards until it reaches a lead node. At every intermediate node, a particular feature value is compared to the splitting value. Depending on the outcome of this comparison, traversing continues with the left or the right path. Decision trees are usually constructed in a top-down and greedy manner such that once a feature and a feature's value are selected as the splitting criterion, they cannot be switched by another splitting value or splitting feature \citep{letham2012building}. Every decision tree can be transformed into a rule-based model but not vice versa. 

\emph{Rule-based models:} Decision rules have the structure of IF \textit{condition} THEN \textit{label} ELSE \textit{other label}. There are different specific rule-based approaches such as simple decision rules, decision sets, decision tables and m-of-n rules.
To add a rule to an already existing one, one can either build a list by adding a rule with \textit{else if} or create a set by adding another rule without any additional syntax. A condition in a rule consists either of a single feature, operator, value triple---in the literature sometimes called literal \citep{wang2015bayesian} or predicate \citep{lakkaraju2016interpretable, lakkaraju2017interpretable}---or it consists of a conjunction or disjunction of predicates. The most common case is to use conjunctions of predicates as the condition \citep{huysmans2011empirical}. 



\emph{Naive Bayes:} The Bayesian classification is a statistical classification method that predicts the probability of an object belonging to a particular group. Naive Bayesian classifiers simplify this problem by assuming the validity of the independence assumption. This assumption states that the effect of a characteristic in the classification is independent of the expressions of other feature values. This assumption is naive insofar as it rarely applies to reality. However, it significantly reduces the complexity of the problem.

\emph{k-nearest neighbours:} Nearest neighbor models do not explicitly build a model from the training data but instead use a similarity measure to compute the nearest neighbors of the test data instance. In case of classification, the label $y$ of the test instance is determined by means of majority voting. The predicted value $y$ for regression problems is obtained via averaging over the neighbor's values.

\emph{Interactive models:} 
Interactive learning combines human feedback with the machine learning concepts of active learning and online learning \citet{holzinger2019interactive}. Thus, interactive learning is considered a subset of the so-called ``human-in-the-loop'' algorithms. Learning algorithms are built interactively through end-user input. End-users can review the model output, make corrections in the learning algorithm, and provide feedback about the model output to the learning algorithm. \citet{holzinger2017glass} summarized the aim of this by concluding to make use of the human cognitive abilities when machines fail.

\emph{Bayesian Networks:} Bayesian networks are acyclic graphs in which the variables are represented as nodes and the relationship between the variables as directed edges. These links between variables express a conditional probability. They are designed to model causal relations in the real world. Bayesian networks do not provide a logical but a probabilistic output \citep{charniak1991bayesian}.

\subsection{General-Purpose Techniques for Interpretability by Design }
\label{sec:PreTeqInterpretableByDesign}
Interpretability can be forced into a model. This means that we can clearly choose between a more interpretable model or a more accurate model. However, this does not necessarily imply a trade-off between interpretability and accuracy. Both goals can be achieved if they are specifically examined during the generation procedure.
It is believed that linear models are better interpretable than deep models, where most popular notions of interpretability depend on the comprehensibility of the features \citep{lipton2016modeling}.
However, if we compare linear models to deep models, deep neural networks tend to operate on raw or lightly processed features while linear models often have to operate on heavily hand-engineered features in order to reach a similar performance level\citep{lipton2016mythos}. Those raw features used by deep models are intuitively meaningful, whereas the hand-engineered features used in linear models are not easy decomposable \citep{lipton2016mythos}. Moreover, linear models which rely on the application of specific features tend to be less likely to also be useful when deployed for other tasks. Conversely, deep neural networks seem to generalize better on different tasks \citep{lipton2016modeling}.

Rudin \citep{stoprudin} claims that this is often not true especially when considering structured data with meaningful features. Interpretability can be enhanced into linear models or Bayesian networks via constraints by demanding\eg sparsity or monotonicity; it can be incorporated into rule- and tree-based prediction models by restricting the number of rules or by limiting the tree size. Other approaches use a specific penalizing procedure that make use of a combination of the methods just described.

\emph{Sparsity:} Sparsity needs to be introduced into the model due to the limited cognitive capacity of humans \citep{ustun2014methods}. However, one has to be careful with prediction models that are too simple \citep{freitas2014comprehensible}. Such models tend to oversimplify the underlying relationship between input features and output. This category supports the generation of local or global explainability.

\emph{Monotonicity:}
Monotonicity constraints are concerned with the relationship between input features and output. If this relationship is monotone, then the connection between input features and output can be more easily grasped by the user. A monotone relationship between an input and output is characterized by an increase in the input value leading to either an increase or a decrease in the output value. The monotonicity constraint can be incorporated into the system as either a hard or a soft constraint \citep{freitas2014comprehensible}. Furthermore, \citet{martens2011performance} distinguish between the phases in the data mining process that incorporate the monotonicity constraint into the system. This category supports the generation of both local and global explainability.

\emph{Ontologies:}
An ontology is defined as a specification of a conceptualization in the context of knowledge sharing \citep{gruber1993translation}.  
Ontologies can be applied to any type of interpretable model (surrogate or white box) and the corresponding data set. The ontology provides an extension or simplification for each explanation belonging to a set of all possible explanations as an output (either local or global). Ontologies are described in detail in Section~\ref{sec:Ontolgies}.

\subsection{The Explanation}
\label{sec:explanations_definition}
Miller states that an explanation is the answer to a why question \citep{miller2017explanation}. 
\citet{gkatzia2016natural} found out that human decision-making can be improved by \textit{Natural Language Generation (NLG)}, especially in the area of uncertain data. 
According to \citet{Kass1988}, the quality of an explanation depends on three different criteria. These criteria are relevance, persuasiveness and comprehensibility. An explanation is relevant if it responds to the current goals and needs of the user. On the one hand, the explanation should provide as much information as possible to achieve these goals. On the other hand, it should be as short as possible in order to avoid providing additional information that is not necessary and could distract the user.
An explanation is said to be convincing if the user accepts it. The user is convinced by an explanation if it is based on facts that the user believes to be true.
Comprehensibility of an explanation is achieved through different facets. The explanation system should use a specific explanation type that the recipient is able to understand. Beside the facets that an explanation should be short and highlight interesting aspects to the user, the explanation should also be simple so that the recipient does not have to look up too many unfamiliar terms \citep{Kass1988}. 
For a system that creates explanations, it is important to have a certain degree of flexibility and responsiveness. If the user fails to understand an explanation, the system should provide further knowledge in order to satisfy the user's needs \citep{Swartout1991}. 
\citet{Fischer1990} confirm the aspect of \citet{Kass1988} that an explanation should be as brief as possible. They introduce an explanation system which generates minimal explanations. When the system receives feedback from the user that the explanation was not sufficient, it adds further details to the given explanation \citep{Fischer1990}. This approach is capable of satisfying the needs of the user without overburdening him. Furthermore, this approach tries to avoid complex explanations. Such explanations are too detailed or not well structured which makes it difficult for the user to understand them \citep{weiner1980blah}.

The nature of data is also important for a convincing explanation, as some of the proposed dimensions can be influenced by different types of data. This means that, depending on the available data type, different explanatory approaches must be considered and the type of communication may change. For example, using a heat map to visualize the gradient of a model is better suited for image data than for tabular or text data.  
\citet{biran2017human} believes that people will only trust a prediction made by machine learning models if the system can justify its decision.
\citet{lipton2016mythos} describes that explanations should focus on abnormal reasons \citep{tversky1974judgment}.
\citet{lipton2016mythos} states that an interpretable model should be human-simulatable. Human-simulatable means that a user is able to ``take in input data together with the parameters of the model and in reasonable time step through every calculation required to produce a prediction''. To build an explanation, the following building blocks can be considered: 

\emph{\textbf{\textit{What}} to explain (content type), \textbf{\textit{How}} to explain (communication), to \textbf{\textit{Whom}} is the explanation addressed (target group)}

\subsubsection{Content Type}
Depending on the model, different explanations can be generated. In order to explain a decision, we need to choose a certain type of explanation or stylistic element. We differentiate between the following types: 

\textit{Local explanation:} Local explainability is only concerned with an individual's decision \citep{phillips2017interpretable} and provides the reason (or reasons) behind a specific decision \citep{doshi2017towards}. Local area explanations only regard the neighborhood of a data instance to explain the prediction. 

\textit{Counterfactual (local) explanation:} Counterfactual explanations \citep{wachter2017counterfactual} provide data subjects\eg a customer with meaningful explanations to understand a given decision, grounds to contest it, and advice how to change the decision to possibly receive a preferred one (e.g. loan approval). 

\textit{Prototype (local) explanation:} Prototype explainability is provided by reporting similar examples to explain the initial decision. Examples are prototypical instances that are similar to the unseen instance. According to \citet{ruping2006learning}, providing examples helps to equip a model with explainability.

\textit{Criticism (local) explanation:} Criticism \citet{ruping2006learning} supports prototypes since it detects what the prototypical instance did not capture. 

\textit{Global explanation:} Global explainability covers global dependencies to describe what a model focuses on in general. 
The global scope is concerned with the overall actions \citep{phillips2017interpretable} and provides a pattern that the prediction model discovered in general. The system can convey the behavior of a classifier as a whole without regarding predictions of individual instances \citep{lakkaraju2017interpretable}.

 \subsubsection{Communication}
 The communication type determines how the explanation is communicated to the user. 

\textit{Textual Description:} 
Explainability is provided in text form. This mimics humans insofar as humans normally justify their decisions verbally. Examples of textual descriptions are generated captions of an image or explanation sentences that justify why a specific class was predicted.

\textit{Graphics:}
Explainability is provided in visual form. Visualization approaches try to illustrate what a model has learned by, e.g., depicting the parameters of the prediction model.

\textit{Multimedia:} Explainability through multimedia combines different types of content: text, graphics, animation, audio and video.

 \subsubsection{Target Group}
 \label{sec:target group}
There are different groups who seek explainability. They pose different requirements for an explainable system due to the difference in experience and due to the different underlying goals \citep{weller2017challenges}. Therefore, explanations serve different purposes and can have different degrees of complexity.
The user might be a non-expert and completely inexperienced with both machine learning and the domain.
This user needs very simple and easily comprehensible explanations.
Alternatively, the user might be a domain expert and familiar with the peculiarities of the data, although not familiar with machine learning.
Such a user can be presented with more complex explanations and more subtle data.
A domain expert could even feel offended if the explanations presented to him do not have a certain complexity (e.g., a doctor who prefers precise diagnoses over vague descriptions).
Typically, a machine learning engineer will not be familiar with the domain but has a lot of experience with being exposed to complex topics.
An explanation for an engineer can be more technical and may even contain internals of a model.
Also related to the experience of the user is the time frame available to comprehend the explanation.
An explanation that must be read in moments has to have a different look than an explanation that is to be fully understood within days. We consider the following groups:

\textit{Non-Expert:}
The non-expert uses the system. He neither has technical nor domain knowledge but receives the decision. His goal is to understand why a certain prediction was made\eg why his credit was denied. Furthermore, he could be interested in understanding whether the explanation domain is composed of simple or complex input features. 

\textit{Domain-Expert:}
A domain expert also uses the system but he is an expert regarding the domain of application. His goal is to more deeply understand the system and the factors and features that are used by the system in order to incorporate his domain knowledge into the system and finally trust the system and its predictions. The domain expert considers trust an important requirement for system deployment.

\textit{System-Developer:}
A developer or system-designer builds the system and is a technical expert that usually does not have any domain knowledge. His goal is to really understand how the algorithm works and what features it exploits. He is concerned with the overall functionality of the system.

\textit{AI-Developer:}
The AI expert trains the model. He is a technical expert in the field of AI without any or only little domain knowledge. His goal is to understand how the algorithm works and what features it exploits so that he can debug and improve the model from a technical standpoint \eg the accuracy.


\subsection{Assessment of Explainability}
\label{sec:assessmentExplainability}

After we elaborated on the building blocks of an explanation, we now want to discuss some suitable metrics for explainability.
\citet{miller2017explanation} did a survey of psychological studies to find out what humans actually consider a \emph{good} explanation.
His major findings are that an explanation should be crucially contrastive, concise, selected, social and that probabilities are less important. Furthermore, explanations should go beyond statistics and imply causality. They need to make certain assumptions about the target's beliefs.

Contrastive explanations should clarify why an input converted into a specific output and not into some counterfactual contrast output \citep{hilton1990conversational}.
Saying that explanations should be concise means that explanations that are too long are not considered interpretable by humans. 
Explanations are also considered better if they confirm the beliefs of the addressee. 
This is generally known as \emph{confirmation bias}.
Explanations should fit the social context of the person that judges the interpretability. This means that the explanation should not only fit the knowledge of this person but also how she sees herself and her environment.
Furthermore, explanations should not contain probability measures since most humans struggle to deal with uncertainty \citep{miller2017explanation}. 
\citet{tversky1981framing} note that explanations should also focus on abnormal reasons.
Humans prefer rare events as explanations over frequent ones. All of these attributes are hard to measure.
In order to asses explainability, \citet{doshi2017towards} propose three levels for tests: First, experiments on real-world tasks with humans; second, simple, experimental tasks with humans; and third, proxy tasks and metrics where other studies with the above assessment methods validate that they are good surrogates.
When a system is evaluated for interpretability based on a proxy, usually complexity and sparsity measures are used. Especially for rule- and tree-based models, there are different measures. There are measures for sparsity that are concerned with the total number of features used \citep{su2015interpretable}. For decision trees, sparsity measures the total number of features used as splitting features \citep{craven1996extracting}. For rule-based models, the size measures the total number of rules in a decision set \citep{lakkaraju2016interpretable}, \citep{lakkaraju2017interpretable} or decision list \citep{bertsimas2011ordered}, \citep{letham2015interpretable}. The length of a rule measures the total number of predicates used in the condition of the decision set \citep{lakkaraju2016interpretable} or decision list \citep{bertsimas2011ordered}. The length of each single rule can be accumulated to measure the total number of predicates used \citep{lakkaraju2017interpretable}. Furthermore, measures for the total number of data instances that satisfy a rule (called \textit{cover} in \citet{lakkaraju2016interpretable}) and measures for the total number of data instances that satisfy multiple rules (called \textit{overlap} in \citet{lakkaraju2016interpretable}) can be used to measure complexity. \citet{freitas2014comprehensible} suggests a different measure for the complexity of a rule-based model. He argues that the average number of rule conditions which were considered for making predictions is a better complexity measure.
For tree-based models, \textit{size} measures the number of tree nodes \citep{craven1996extracting}. Another measure is their \textit{depth} \citep{ruping2005learning}. Furthermore, feature importance can be extracted from rule- and tree-based models. The total number of instances that use a feature for classification can be used as the feature's importance. \citet{samek2017explainable} uses perturbation analysis for measuring explainability. This method is based on three simple ideas. First, the predictive value of a model suffers more from the disturbance of important input features than from unimportant features. Second, approaches such as Sensitivity Analysis or Layer Wise Relevance Propagation provide a feature score that makes it possible to sort them. Third, it is possible to change the input values iteratively and document the predicted value. The averaged prediction score can be used to measure the explanation quality. If the averaged prediction score fluctuates, it can be an indication for important or unimportant parameters of an explanation.

Other measurement methods are introduced for recommender systems \citep{abdollahi2017using}. The similar based approach analyzes the user's neighborhood. The more users from the neighborhood recommend a product, the better this majority vote can be used as a basis for an explanation. The approach from \citet{abdollahi2016explainable} also uses neighborhood style explanations. An explainability score with a range of zero to one is used to determine whether the user rated a product in a similar way that his neighborhood users did. If the value is zero, the user is not contributing to the user-based neighbour-style explanation. Another important aspect that is needed during the testing of an explanatory process is the ground truth. Therefore, an exploration environment such as the Kandinsky Patterns can be used \citep{holzinger2019kandinsky}. The work of \citet{murdoch2019interpretable} introduces the \textit{Predicitve, Descriptive, Relevant (PDR) framework}. This framework uses three components for the interpretation generation and evaluation: \textit{predictive accuracy}, \textit{descriptive accuracy}, and \textit{relevancy}. \textit{Predictive accuracy} is the error measure for given interpretations. \textit{Descriptive accuracy} describes the relationships that models learn. \textit{Relevancy} is given if the explanation provides meaningful information for a specific audience \citep{murdoch2019interpretable}.

However, measuring explainability is still a complicated task due to its subjectivity. Whether an explanation is considered satisfactory depends on the knowledge, needs and objectives of the addressees. 
\citet{adadi2018} mention another challenge that makes measuring explainability difficult. They describe that ML models often have a complex structure and that this can lead to the fact that for the same input variables with the same target values, the algorithm generates different models because the algorithm passes through different paths during execution. This, in turn, leads to different explanations for the same data, making it difficult to accurately measure the explainability.

Users may have different types of questions they want to ask. In the work of \citet{hoffman2018metrics}, each domain expert had the possibility to rate explanations by a scoring system. The result was summed up and then divided by the total amount of participants. The \emph{Content Validity Ratio (CVR)} score was then transformed to a range between -1 and +1. A score above 0 indicates that the item is considered meaningful according to the explanation satisfaction. 
Furthermore, \citet{hoffman2018metrics} present the \emph{Explanation Goodness Checklist} for measuring the goodness of explanations. Here, the application context differs from the one used for the \emph{Explanation Satisfaction Scale}. While the Explanation Satisfaction Scale collects opinions of participants after they have worked with the XAI system, the Explanation Goodness Checklist is used as an independent evaluation of explanations by other researches.


In what follows, we will survey past and recent explainability approaches in \ac{sml}, mainly in classification. We will conclude each chapter that describes a category according to the definitions in section \ref{sec:problem} with an illustrative example on the well-known IRIS data set for the task of classification \citep{Dua:2019}. 
The instance we use for the local procedures is taken from the class \textit{virginica} with the feature values $[5.8, 2.8, 5.1, 2.4]$.


\begin{figure}
    \centering
    \begin{tikzpicture}[scale=0.95]
        \node[rblock, text width=1.8cm, minimum height=1.2cm] at (0,0) (A) {\small Learning Algorithm};
        \node[rblock, text width=1.8cm, minimum height=1.2cm] at (3,0) (M) {\small White Box Model $w$};
        \node[rblock, text width=1.85cm, minimum height=1.2cm] at (9,0) (E) {\small Explanation $e$};

        \draw[ablock] (A) -- (M);
        \draw[ablock] (M) -- (E);
        \draw[ablock, latex-] (A.west) -- +(-.5,0) node[above, rotate=90, text width = 1.5cm, text centered] {\small Training Data};
    \end{tikzpicture}
    \caption{Learning white box models according to \protect\Fig{fig:relations}(c).}
    \label{fig:interpretable-model}
\end{figure}

\section{Interpretable Model Learning}
\label{sec:interpretable}
While in Section~\ref{sec:explain_models}, we introduced various kinds of interpretable model types, this section focuses on the actual \emph{learning algorithms} that produce these interpretable models. We differentiate between algorithms plainly providing interpretable models---named \emph{interpretable by nature} in the following---and algorithms putting special emphasize on enhancing interpretability---named \emph{interpretable by design}. This section refers to Problem 2 in Section~\ref{sec:problem} and the explanation generation process is outlined in \Fig{fig:interpretable-model}.

\subsection{Interpretable by Nature}
\label{sec:introduction_well-known}
A closer look at some classic SML algorithms shows that they provide models that can already be interpreted without forcing interpretability on them. This means that the training process for the learning algorithms is not explicitly optimized to gain interpretability. Usually, these algorithms are optimized for high accuracy; interpretability is a by-product. It is naturally given and therefore inherent. The resulting models can be explained directly, e.g. through visualizations, in order to gain an understanding of the model's function and behavior. 
However, in order to be easily understood by humans, the models must be sparse, small and simple. The overview of the different approaches is listed in Table \ref{tab:interpretbynature}.

\subsubsection{Decision Trees}
The \emph{\ac{cart}} \citep{breiman2017classification} algorithm is a divide and conquer algorithm that builds a decision tree from the features of the training data. Splitting features and their values are selected using the Gini index as the splitting criterion. The \emph{ID2of3} \citep{craven1996extracting} algorithm is based on M-of-N splits learned for each node using a hill climbing search process for the construction of the decision tree \citep{craven1996extracting}. The \emph{C4.5} algorithm \citep{quinlan1996bagging} is a divide and conquer algorithm that is based on concepts from information theory. A gain ratio criterion based on entropy is used as the node splitting criterion. The \emph{C5.0T} algorithm is an extension of the C4.5 algorithm \citep{ustun2016supersparse}. The \emph{ID3} algorithm chooses attributes with the highest information gain. 
    
\subsubsection{Decision rules}
\emph{1R} \citep{holte1993very} uses as input a set of training examples and produces as output a 1-rule. 1R generates rules that classify an object on the basis of a single attribute.  
The \emph{AntMiner+} algorithm uses ant-based induction that creates one rule at a time \citep{martens2007comprehensible}. The \emph{cAntMiner\textsubscript{PB}} ordered algorithm uses ant-based induction to create an ordered list of rules that consider interactions between individual rules whereas its unordered version creates an unordered set of rules \citep{otero2016improving}. Others are the \emph{CN2} algorithm \citep{martens2009decompositional}, \emph{RIPPER} algorithm \citep{cohen1995fast}, the \emph{Re-RX} algorithm \citep{setiono2008recursive} and the \emph{C5.0R} algorithm \citep{ustun2016supersparse}. 

\subsubsection{Generalized Additive Models (GAMs)}
GAMs learn a linear combination of shape functions in order to visualize each feature's relationship with the target \citep{lou2012intelligible}. A shape function relates a single feature to the target. Hence, a shape function's visualization illustrates the feature's contribution to the prediction. Shape functions can be regression splines, trees or ensembles of trees. 
However, interactions between features are not considered by GAMs. Therefore, \citet{lou2013accurate} introduce interaction terms to GAMs. The \emph{Generalized Additive Models plus Interactions} GA\textsuperscript{2}Ms considers pairwise interactions of features. These interaction terms add more accuracy to the prediction model while maintaining the prediction model's interpretability. \citet{caruana2015interpretable} proofed the explainability of the model
while remaining the state-of-the-art accuracy on a pneumonia case study. 




\subsubsection{The Use Case}
\begin{figure*}
\centering
   \begin{minipage}{0.48\textwidth}
   \centering
     \includegraphics[scale=0.5]{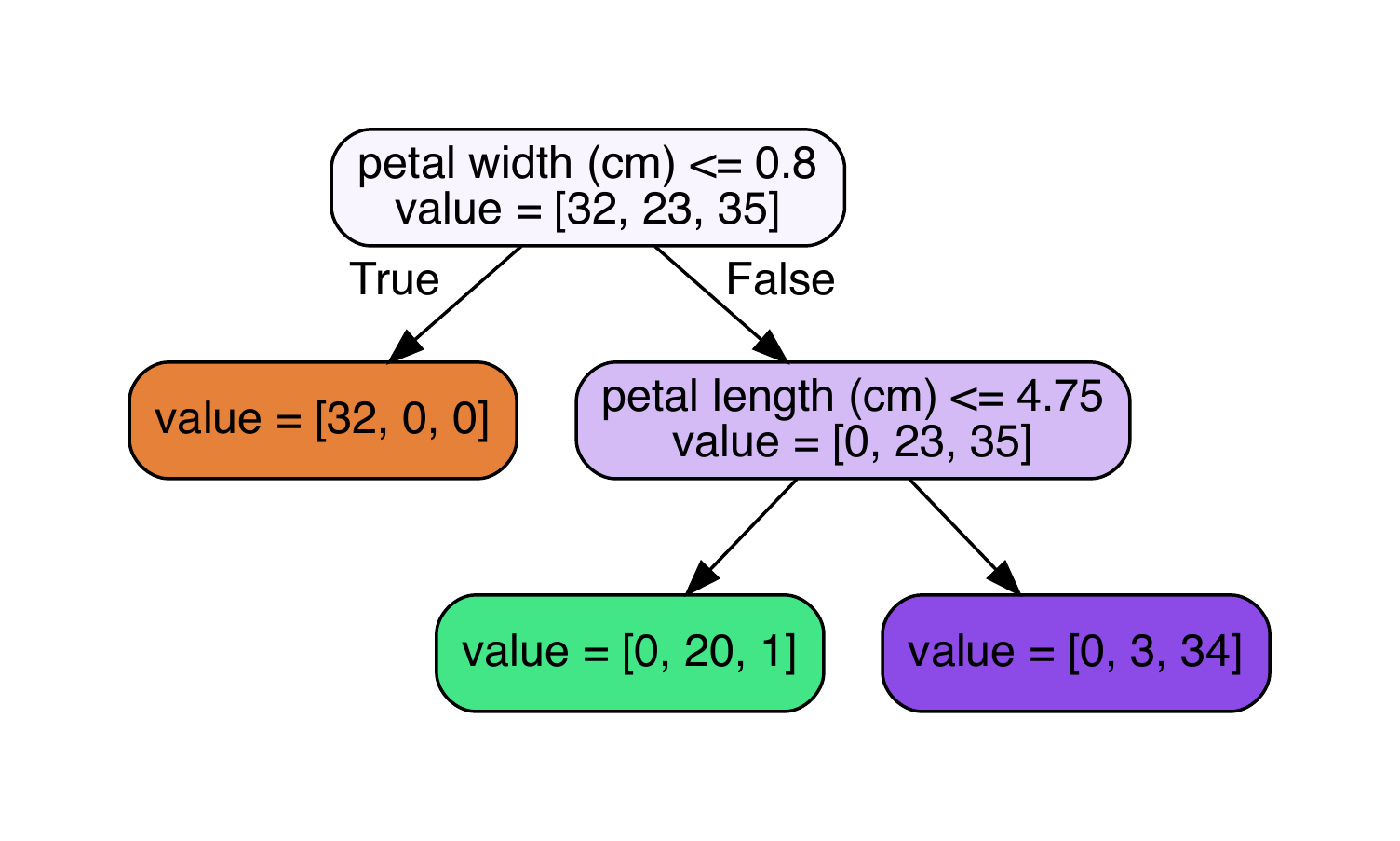}
    \caption{Tree with depth of 2}
    \label{fig:tree2}
     \end{minipage}
 \begin{minipage}{0.48\textwidth}
 \centering
    \includegraphics[scale=0.4]{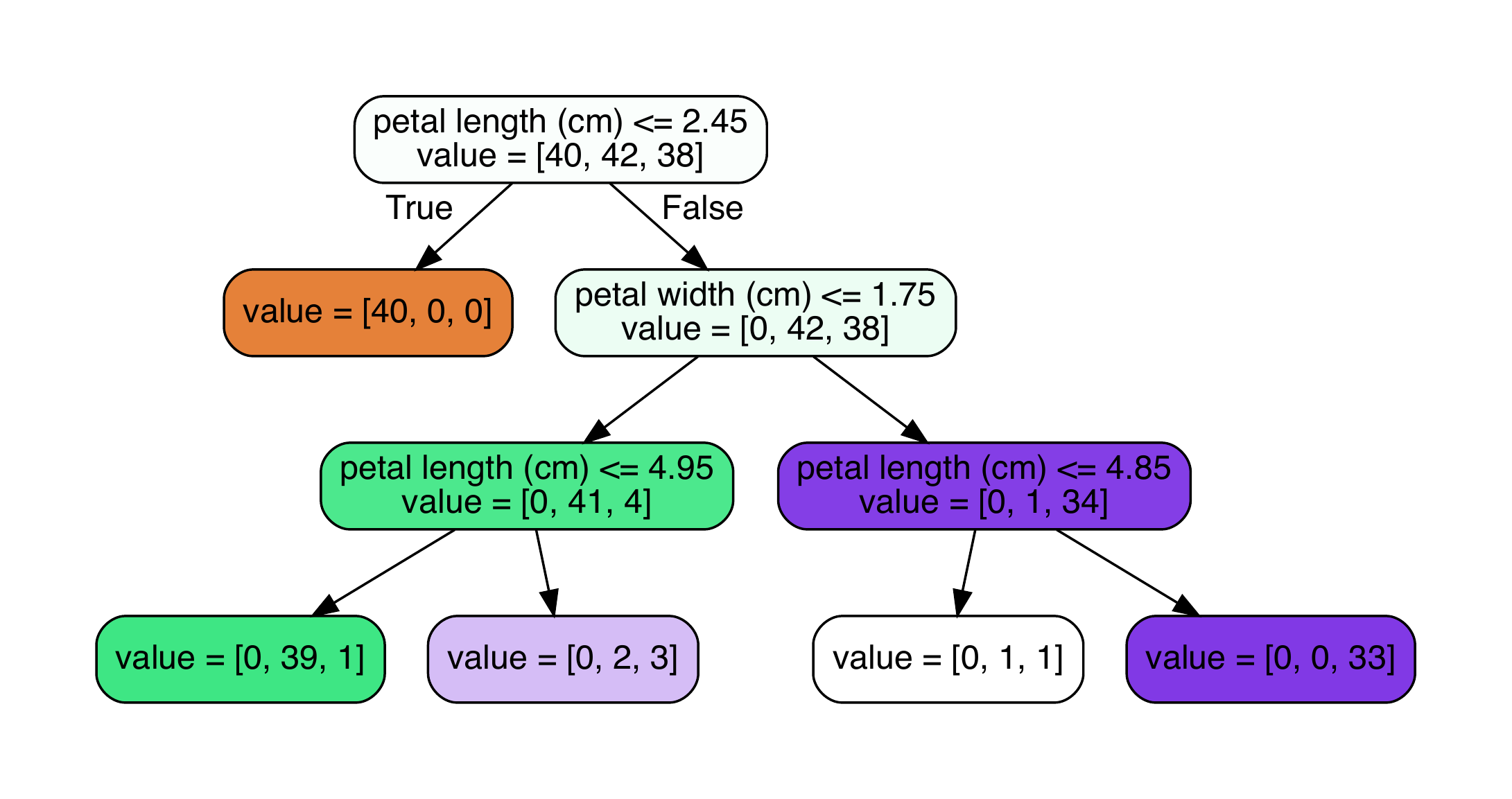}
    \caption{Tree with depth of 3}
    \label{fig:tree3}
      \end{minipage}
\end{figure*}

The decision tree presents the correlations in the data set and provides a simple overview over the contributions that each feature is likely going to have on the data. Decision trees thus allow for ante-hoc model explanations. To train the tree, we need a data set with labeled outputs and the corresponding class labels \textit{virginica, setosa} and \textit{versicolor}. Figure \ref{fig:tree2} illustrates a decision tree with the depth of two. The nodes illustrate which feature is considered or, more precisely, which decision is made to split the data set into the branches. Illustratively, the tree first splits all instances at \textit{petal width 0.8}. Thereby, all instances with petal width smaller than 0.8 are classified as \textit{setosa}, while everything else is further divided by petal length. The user is able to grasp the concept of the decision-making of the decision tree. Figure \ref{fig:tree3} illustrates a decision tree with the depth of three. The tree first splits all instances at \textit{(petal width $\leqq$ 2.45)}, whereas the next split is at \textit{(petal width $\leqq$ 1.75)}. The two last splits are either at \textit{(petal length $\leqq$ 4.95)} or at \textit{(petal length $\leqq$ 4.85)}. Those two examples clearly illustrate that a decision tree with the depth of two or three is still comprehensible.

\subsection{Interpretable by Design}
\label{sec:introduction_enhancing}
An alternative approach to directly train interpretable models is to include interpretability in the design of the training. We call this learning approach ``interpretable by design''. As was the case with the ``by nature'' approaches discussed above, the result usually is a white box model but in contrast, interpretable by design approaches allow the degree of interpretability to be controlled or increased. 
This means that the resulting models are created with the intention of further improving the interpretability for humans. 
Section \ref{sec:PreTeqInterpretableByDesign} lists some general-purpose techniques that can be used to achieve interpretability by design.


\subsubsection{Decision Trees}
\emph{Oblique Treed Sparse Additive Models} (OT-Spam) belong to the category of region-specific predictive models \citep{wang2015trading}. Oblique Trees are used to divide the feature space into smaller regions. Sparse Additive Models called experts are trained for each local region. These experts are assigned as leaf nodes to the oblique tree to make the predictions. For this specific approach, \emph{Factorized Asymptotic Bayesian} (FAB) inference was used to build the prediction model. The size of the tree is also regularized by the FAB inference. 
Another work presents \emph{\ac{ndt}}, an interplay of decision trees with neural networks. This model has the structure of a decision tree where the splitting nodes consist of independent perceptrons. \ac{ndt} can be used for supervised and unsupervised problems and takes the advantages of both methods: the clear model architecture of decision trees and the high-performance capacities of neural networks \citep{balestriero2017neural}. 
\citet{yang2018deep} presents \emph{\ac{dndt}} which are similar to \citet{balestriero2017neural}`s model. This approach is specifically designed for tabular data that learns through backpropagation. \ac{dndt}s differ from \citet{balestriero2017neural}`s method in that each input feature has its own neural network, making it easier to interpret.

\subsubsection{Decision Rules}
The \emph{Bayesian Or's of And's} (BOA) \citep{wang2015bayesian} model is based on a Bayesian approach that allows the incorporation of priors. These priors are used to regulate the size and shape of the underlying rules. Wang introduces the Beta-Binomial prior and the Poisson prior. The Beta-Binomial prior rules the average size of the conditions, whereas the Poisson prior rules the condition length and the overall model size. To learn such rules, association rule mining and simulated annealing or literal-based stochastic local search is performed. 

\emph{Two-Level Boolean Rules} are either of the form AND-of-ORs or of the form OR-of-ANDs \citep{su2015interpretable}. Since a rule consists of two levels, a rule of the form OR-of-ANDs consists of two conditions connecting predicates with ANDs that are connected with an OR. The proposed approach allows for the incorporation of accuracy and sparsity in form of a trade-off parameter. Two optimization based formulations are considered to learn two-level Boolean rules. One is based on the 0-1 loss, the other is based on the Hamming distance. Furthermore, when the 0-1 loss is used, they use a redundancy-aware binarization method for two-level LP-relaxation. When the Hamming distance is used, they use block coordinate descent or alternating minimization. 
The same idea as in \emph{SLIM} \citep{ustun2016supersparse} can be used to construct M-of-N rule tables. \emph{TILM} uses thresholds of feature values and a feature selection process to find binary decision rules \citep{ustun2014methods}.

The \emph{Ordered Rules for Classification} model introduces a decision list classifier that is based on mixed integer optimization (MIO) and association rules \citep{bertsimas2011ordered}. The set of general association rules is found using MIO. MIO is also used for ordering these rules and incorporating other desiderata such as the number of features used in a condition and the total number of rules.

The \emph{Bayesian List Machine} (BLM) model is based on a Bayesian framework that learns a decision list. Sparsity as the trade-off between interpretability and accuracy can be introduced into the Bayesian framework. The learning process consists of first finding the rules and then ordering them \citep{letham2012building}.

The \emph{Bayesian Rule List} (BRL) method differs from BLM just by a different prior \citep{letham2015interpretable}. They use a Bayesian framework and incorporate sparsity into the model. 
As output, BRL provides class probabilities. \emph{Scalable BRL} \citep{hongyu} is an extended version of BRL that provides a posterior distribution of the rule list. 

\emph{Falling Rule Lists} (FRL) are decision lists that are ordered based on class probabilities. With each rule going down the list, the probability of belonging to the class of interest decreases. Furthermore, once an unseen instance matches a rule it is assigned the probability of the rule it matches instantaneously. Since Falling Rule Lists are based on a Bayesian framework, a prior for the size of a decision list can be integrated into the learning process \citep{wang2015falling}.

\citet{malioutov2017learning} propose a rule-based classifier by applying a linear programming (LP) relaxation for interpretable sparse classification rules. \emph{Classification based on Predictive Association Rules} (CPAR) uses dynamic programming to generate a small set of high quality and lower redundancy \citep{yin2003cpar}. \emph{Two-Level Boolean Rules} (TLBR) create classification predictions by connecting features with logical statements in rules \citep{su2016learning}.

\subsubsection{Decision sets} 
The goal of \emph{Interpretable Decision Sets} (IDS) is to learn a set of short, accurate, non-overlapping rules. These learned rules can be used independently of each other. \citep{lakkaraju2016interpretable} use a pre-mined rule space on top of which they apply association rule mining. Only frequent item-sets are considered and mined. In the process of selecting a set of rules, different objectives that include accuracy and interpretability are optimized. 

\emph{Bayesian Rule Set} (BRS) is another approach that introduces interpretable sets of rules \citep{wang2016bayesian}. This approach is based on a Bayesian framework that incorporates the total number of rules as a sparsity prior. Furthermore, for each rule in the set, the minimum support requirement has to be met. 

\subsubsection{Linear Models}
\citet{ustun2014methods} introduces an \emph{integer programming framework}. In addition to the creation of various interpretable models, scoring systems can also be created by this framework. 
Scoring systems assign each feature or feature interval a weight. A final score is evaluated in the same way as the prediction of a linear model. However, an instance is classified only after comparing the final score with a defined threshold.

\emph{Supersparse linear integer models} (SLIM) is a method used for making scoring systems more interpretable \citep{ustun2014methods}. The authors introduce a pareto-optimal trade-off between accuracy and sparsity. Here, the model is considered to be interpretable when it is sparse. In order to achieve this trade-off, an integer program that encodes 0-1 loss for accuracy and L\textsubscript{0} seminorm for sparsity together with a regularization parameter for the trade-off is used to learn SLIM. PILM is a generalization of SLIM. 

\subsubsection{Surrogate Fitting}
The concept of interpretable by design can be extended so as to also include approaches that learn a black box model but that are optimized towards improving surrogate model fitting. 
That is, the learning algorithm of the black box model already is designed in such a way that fitting the surrogate to the black box model is simplified or results in a higher fidelity. 
\citet{wu2018beyond} for instance achieve this by using regularization during training a neural network that forces the network to allow a small decision tree to be fitted as a global surrogate model. A similar approach is considered in \citet{schaaf2019enhancing}, where the used L\textsubscript{1}-orthogonal regularization allows significantly faster training. They state that this preserves the accuracy of the black box model while it can still be approximated by small decision trees. \citet{burkart2019forcing} use a rule-based regularization technique to enforce interpretability for neural networks.

\subsubsection{The Use Case}
\ac{brl} construct rules to be accurate but still interpretable for the users. The approach resembles simple logical decision-making. In this case, the rules are model-specific and support the model interpretability. 
The algorithm requires the labeled data to mine a decision list. Listings \ref{List1}, \ref{List2}, \ref{List3} illustrate the results for classification discriminating each class separately. The rules can be read fairly comfortably and they can tell the user the direct relationship of a feature value with the classification probability that results from that value. These rules apply to all instances alike. The user can quickly learn the knowledge the model has deducted from the data set and thereby grasp the decision-making process. For example, the \textit{setosa} classification rules apply to \textit{setosa} and \textit{not setosa} classification.

\begin{lstlisting}[basicstyle=\small,label=bla2, caption=Setosa Classification, captionpos=b, label=List1]
IF petal width: 0.8 to inf THEN probab. of setosa: 1.2%
ELSE IF petal length: -inf to 2.45 THEN probab. of setosa: 97.4%
ELSE probab. of setosa: 50.0%
\end{lstlisting}

\begin{lstlisting}[basicstyle=\small, caption=Versicolor Classification, captionpos=b, label=List2]
IF petal length: 2.45 to 4.75 THEN probab. of versicolor:97.3%
ELSE IF petal width: 0.8 to 1.7 THEN probab. of versicolor:42.9%
ELSE probab. of versicolor: 2.4%
\end{lstlisting}

\begin{lstlisting}[basicstyle=\small, caption=Virginica Classification, captionpos=b, label=List3]
IF petal length: 5.15 to inf THEN probab. of virginica: 96.6%
ELSE IF petal length:-inf to 4.75 THEN probab. of virginica:2.6%
ELSE IF petal width:-inf to 1.75 THEN probab. of virginica:25.0%
\end{lstlisting}

\begin{table*}
\centering
\caption{Overview of interpretable by nature approaches }
\begin{adjustbox}{width=5in}
\renewcommand{\arraystretch}{1.25}
\begin{tabular}{|l||l|l|l|}
\hline
\textbf {Approach}   &
\textbf{Learning Task}  & \textbf{Model} & \textbf{References}  \\ \hline \hline
GAMs  & Classification & Linear Model     & \citet{lou2012intelligible} \\ \hline
CN2  & Classification & Rule-based   & \citet{clark1989cn2} \\ \hline
RIPPER & Classification & Rule-based   & \citet{cohen1995fast} \\ \hline
 Re-RX  & Classification & Rule-based   & \citet{setiono2008recursive} \\ \hline
C5.0R  & Classification & Rule-based   & \citet{ustun2016supersparse} \\ \hline
AntMiner+  & Classification  & Rule-based   & \citet{martens2007classification} \\ \hline
cAntMinerPB  & Classification & Rule-based    & \citet{otero2016improving} \\ \hline
CART  & Classification \&  Regression & Tree-based  & \citet{breiman2017classification} \\ \hline
ID2of3  & Classification \&  Regression&  Tree-based & \citet{craven1996extracting} \\ \hline
ID3  &Classification \&  Regression &  Tree-based& \citet{quinlan1986induction}  \\ \hline
C4.5  & Classification \&  Regression &  Tree-based & \citet{quinlan2014c4} \\ \hline
C5.0T  &Classification \&  Regression &  Tree-based & \citet{ustun2016supersparse}  \\ \hline
Bayesian Network  & Classification & Bayesian Network & \citet{friedman1997bayesian}\\ \hline
Linear regression (Lasso)  & Regression & Linear model & \citet{tibshirani1996regression}\\ \hline
Linear regression (LARS) &  Regression & Linear model & \citet{efron2004least}\\ \hline 
Logistic regression  & Classification &Linear model & \citet{berkson1953statistically} \\ \hline
KNN & Classification \&  Regression & Nearest Neighbor & \citet{freitas2014comprehensible} \\ \hline
MGM & Classification & Clustering & \citet{kim2015mind} \\ \hline
AOT & Classification & Tree-based & \citet{si2013learning} \\ \hline
\end{tabular}
\end{adjustbox}

\label{tab:interpretbynature}
\end{table*}

\begin{table*}
\centering
\caption{Overview of interpretable by design approaches }
\begin{adjustbox}{width=5in}
\begin{tabular}{|l||l|l|l|}
\hline
\textbf {Approach}   &
\textbf{Learning Task}  & \textbf{Model} &  \textbf{References}  \\ \hline \hline
 SLIM & Classification & Rule-based & \citet{ustun2014methods} \\ \hline
TILM &            Classification  & Rule-based & \citet{ustun2016supersparse}\\ \hline
PILM &        Classification  & Linear model & \citet{ustun2016supersparse}\\ \hline
RiskSLIM&         Classification     & Rule-based & \citet{ustun2017optimized}\\ \hline
Two-level Boolean Rules      & Classification      & Rule-based & \citet{su2015interpretable}\\ \hline
BOA &      Classification      & Rule-based & \citet{wang2015bayesian} \\ \hline
ORC& Classification      & Rule-based & \citet{bertsimas2011ordered}\\ \hline
BLM& Classification      & Rule-based & \citet{letham2012building} \\ \hline
BRL & Classification      & Rule-based & \citet{letham2015interpretable}    \\ \hline
(S)BRL & Classification      & Rule-based &  \citet{hongyu}\\ \hline
FRL & Classification      & Rule-based & \citet{wang}    \\ \hline
IDS & Classification      & Rule-based & \citet{lakkaraju2017interpretable}  \\ \hline
BRS& Classification      & Rule-based & \citet{wang2016bayesian}  \\ \hline
OT-SpAM& Classification      & Tree-based & \citet{wang2015trading} \\ \hline
Tree Regularization &   Classification   &  Tree-based     & \citet{wu2018beyond} \\ \hline
NDT & Classification & Tree-based & \citet{balestriero2017neural} \\ \hline
DNDT & Classification & Tree-based & \citet{yang2018deep} \\ \hline
LP relaxation&  Classification       &   Rule-based  & \citet{malioutov2017learning} \\ \hline
1R & Classification & Rule-based & \citet{holte1993very} \\ \hline
TLBR & Classification & Rule-based &  \citet{su2016learning}\\ \hline
CPAR & Classification & Rule-based & \citet{yin2003cpar}\\ \hline
\end{tabular}
\end{adjustbox}

\label{tab:interpretbydesign}
\end{table*}

\section{Surrogate Models}
\label{sec:surrogate}
This section is concerned with discussing various approaches on fitting global or local surrogates to a black box model or a model prediction, respectively. A surrogate model translates the model into an approximate model \citep{henelius2014peek} either local or global. This technique is applied whenever the model is not interpretable by itself, i.e., whenever it is a black box. An interpretable model is build on top of the black box. Separating the prediction model from its explanation introduces flexibility, accuracy and usability \citep{ribeiro2016model}.

\subsection{Global Surrogates}
\label{sec:surrogate_global}
This section describes global surrogate models that learn interpretable models to mimic the predictions of a black box model. The overview of the different approaches are listed in Table \ref{tabl:globalsurrogate} and \ref{overviewrule}. The global surrogate fitting process is depicted in \Fig{fig:global-surrogate}.

\begin{figure}[t]
    \centering
    \begin{tikzpicture}[scale=0.95]
        \node[rblock, text width=1.8cm, minimum height=1.2cm] at (0,0) (A) {\small Learning Algorithm};
        \node[rblock, text width=1.8cm, minimum height=1.2cm] at (3,0) (M) {\small Black Box Model $b$};
        \node[rblock, text width=1.8cm, minimum height=1.2cm] at (6,0) (S) {\small Surrogate $w$};
        \node[rblock, text width=1.85cm, minimum height=1.2cm] at (9,0) (E) {\small Explanation $e$};
        
        \filldraw ($(M.north)+(0,.4)$) circle (.8mm) node (c) {};

        \draw[ablock] (A) -- (M);
        \draw[ablock] (M) -- (S);
        \draw[ablock] (S) -- (E);
        \draw[ablock, latex-] (A.west) -- +(-.5,0) node[above, rotate=90, text width = 1.5cm, text centered] {\small Training Data};
        \draw[ablock, latex-] (M.north) -- +(0,.6) node[above] {\small Test Data};
        \draw[ablock, shorten <= -2] (c) -| (S.north);
    \end{tikzpicture}
    \caption{Fitting a global surrogate according to \protect\Fig{fig:relations}(d).}
    \label{fig:global-surrogate}
\end{figure}

\subsubsection{Linear Models}
The \emph{sub-modular pick algorithm} (sp-Lime) provides global explanations by using sub-modular picks to find representative instances of the underlying prediction model \citep{ribeiro2016should}. The explanations offered by Lime for these representative instances are combined and considered as the global explanation of the black box. Another variation is k-Lime, where a clustering algorithm is applied to the black box model to find \textit{k} clusters \citep{hall2017machine}. These clusters represent local regions. For each local region, a \emph{Generalized Linear Model} (GLM) is learned. A global surrogate GLM is used as an explanation when the unseen data instance does not fall into a local region.

\subsubsection{Decision Trees}
 \citet{andrzejak2013interpretable} introduce an approach to merge two decision trees into a single one based on distributed data. The procedure utilizes an efficient pruning strategy that is based on predefined criteria. 
 
 The \emph{Decision Tree Extraction} \citep{bastani} approach extracts a new model from a given black box model. The resulting surrogate model is then used as an approximation of the original black box model in the form of a decision tree. First, the algorithm generates a Gaussian Mixture Model (GMM) to cluster the data points in the training data set. The second step comprises the computation of class labels for the clustered data points by utilizing the black box model to approximate. 
 
 \citet{hinton2017distilling} illustrate a soft decision tree which is trained by stochastic gradient descent using the predictions of the neural net. The tree uses learned filters to make hierarchical decisions based on an input example.
 
 \citet{yang2018global} propose a binary decision tree that represents the most important decision rules. The tree is constructed by recursively partitioning the input variable space by maximizing the difference in the average contribution of the split variable. 
 
 \citet{schetinin2007confident} describe the probabilistic interpretation of Bayesian decision tree ensembles. Their approach consists of the quantitative evaluation of uncertainty of the decision trees and allows experts to find a suitable decision tree. 
 
 \citet{hara2016making} approximate a simple model from tree ensembles by deriving the expectation maximization algorithm minimizing the Kullback-Leibler divergence. 

\begin{table}
\centering
\caption{Overview of global surrogate approaches}
\label{tabl:globalsurrogate}
\begin{adjustbox}{width=\linewidth}
\renewcommand{\arraystretch}{1.25}
\begin{tabular}{|l||l|l|l|l|}
\hline
\textbf {Approach}   &
\textbf{Learning Task}  &
\textbf{Input Model} &
\textbf{Model} & \textbf{References}  \\ \hline \hline

SP-Lime& Classificiation & Agnostic &Linear Model & \citet{ribeiro2016should} \\ \hline

k-Lime & Classificiation & Agnostic & Linear Model & \citet{hall2017machine} \\ \hline

Tree Merging &  Classificiation   & Specific (DT) &Tree-based& \citet{andrzejak2013interpretable} \\ \hline
Decision Tree extract &  Classificiation  & Specific (DT) & Tree-based& \citet{bastani}\\ \hline
Soft Decision Tree &  Classificiation &  Agnostic & Tree-based& \citet{hinton2017distilling}\\ \hline
Binary Decision Tree & Classificiation & Agnostic & Tree-based& \citet{yang2018global} \\ \hline
Probabilistic interpretation &  Classificiation & Specific (DT) & Tree-based& \citet{schetinin2007confident} \\ \hline
EM min. Kullback&   Classificiation & Specific (DT) & Tree-based& \citet{hara2016making} \\ \hline
\end{tabular}
\end{adjustbox}
\end{table}

\subsubsection{Rule-based} 
\begin{figure}[t]
    \centering
    \includegraphics[width=\columnwidth]{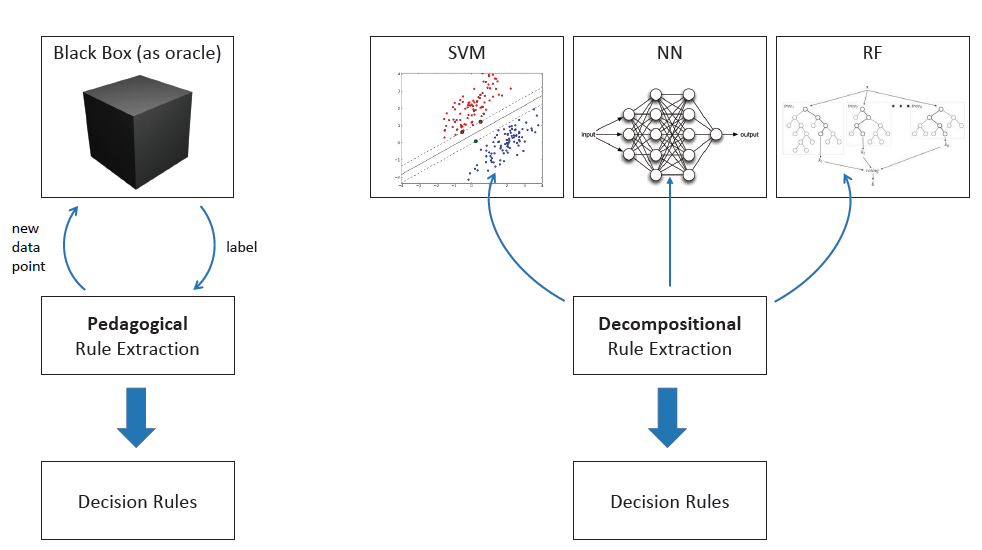}
    \caption{Difference between pedagogical and decompositional rule extraction approaches}
    \label{fig:ruleextract}
\end{figure}

Rule extraction approaches learn decision rules or decision trees from the predictions made by black boxes. They are divided into pedagogical, decompositional (see Figure \ref{fig:ruleextract}) and eclectic approaches \citep{andrews1995survey, martens2011performance}. The pedagogical approach (model-agnostic) perceives the underlying prediction model as a black box and uses the provided relation between input and output. The decompositional approach (model-specific) makes use of the internal structure of the underlying prediction model. The eclectic or hybrid approach combines the decompositional and the pedagogical. The aforementioned approaches regarding interpretable decision rules and decision trees can also be applied to black boxes. The approaches are listed in Table \ref{overviewrule}.

The rule extraction techniques have an advantage over interpretable decision rules or trees. Although a combination of features is not present in the training data, the rule extraction technique can still work since its underlying prediction model labels any feature combination. Decision rules depend on the available training data and cannot augment their training set \citep{craven1996extracting}. 

\begin{table*}
\centering
\caption{Overview of surrogate rule extraction approaches and their properties of explainability}
\begin{adjustbox}{width=\textwidth}
\renewcommand{\arraystretch}{1.25}
\begin{tabular}{|l||l|l|l|l|l|}
\hline
 \textbf{Approach}           & \multicolumn{1}{l|}{\textbf{\begin{tabular}[c]{@{}l@{}}Extraction \\ Approach\end{tabular}}} & \multicolumn{1}{l|}{\textbf{\begin{tabular}[c]{@{}l@{}}Learning\\ Task\end{tabular}}} & \multicolumn{1}{l|}{\textbf{\begin{tabular}[c]{@{}l@{}}Model\\ Scope\end{tabular}}}  & \multicolumn{1}{l|}{\textbf{\begin{tabular}[c]{@{}l@{}}Model\end{tabular}}} & \multicolumn{1}{l|}{\textbf{\begin{tabular}[c]{@{}l@{}}References\end{tabular}}} 

\\ \hline \hline
\textbf{BRAINNE} & Feature Selection & Classification & Pedagogical& Rule-based & \citet{sestito1992automated}                                                                          \\ \hline
\textbf{VIA}            & Sensitivity analysis                                                                                                                                  & Classification                                                       & Pedagogical& Rule-based                                                                       &  \citet{zilke2016deepred}                                                                            \\ \hline
\textbf{BIO-RE}         & Feature Selection                                                                                                & Classification                                                       & Pedagogical & Rule-based                                                                                        &  \citet{taha1996three}                                               \\ \hline
\textbf{G-Rex}          & Genetic algorithm                                                                                                    & \begin{tabular}[c]{@{}l@{}}Classification \\ Regression\end{tabular} & Pedagogical & Rule-based                                                                 &  \citet{johansson2004truth}                                                                            \\ \hline
\textbf{STARE}          & Feature Selection                                                                                                 & Classification                                                       & Pedagogical & Rule-based                                                                             &  \citet{zhou2000statistics}                                                                \\ \hline
\textbf{REFNE}          & Ensemble concept                                                                                                            & Classification                                                       & Pedagogical& Rule-based                                                                        &  \citet{zhou2003extracting}                                                                         \\ \hline
\textbf{BUR}            & Ensemble concept                                                                                                       & Classification                                                       & Pedagogical & Rule-based                                                                       &  \citet{ninama2013ensemble}                                                                    \\ \hline
\textbf{Iter}           & Sequential Covering                                                                                                 & \begin{tabular}[c]{@{}l@{}}Classification\\ Regression\end{tabular}                                            & Pedagogical & Rule-based                                                                       &  \citet{huysmans2006iter}                                                                  \\ \hline
\textbf{OSRE}           & Orthogonal search-based                                                                                                     & Classification                                                       & Pedagogical & Rule-based                                                                        &  \citet{etchells2006orthogonal}                                                                           \\ \hline
\textbf{Minerva}        & Sequential Covering                                                                                                           & \begin{tabular}[c]{@{}l@{}}Classification\\ Regression\end{tabular}                                            & Pedagogical & Rule-based                                                                                                                         &  \citet{martens2008rule}                \\ \hline
\textbf{RxREN}          & Reverse engineering                                                                                                             & Classification                                                       & Pedagogical& Rule-based                                                                      &  \citet{augasta2012reverse}                                                                           \\ \hline
\textbf{RxNCM}          & Correctly and misclassified data ranges                                                                                                           & Classification                                                       & Pedagogical & Rule-based                                                                     &    \citet{biswas2017rule}                                                                      \\ \hline
\textbf{BETA}          & Reverse engineering                                                                                                           & Classification                                                       & Pedagogical & Rule-based                                                                        &  \citet{lakkaraju2017interpretable}                                                                          \\ \hline
\textbf{KDRuleEX}       & Genetic algorithm                                                                                                             & Classification                                                       & Pedagogical & Rule-based                                                                      &  \citet{sethi2012extended}                                                                 \\ \hline
\textbf{TREPAN}       &  Symbolic Learning                                                                                                    & Classification                                                       & Pedagogical & Tree-based                                                                                              &      \citet{craven1996extracting}                                   \\ \hline
\textbf{ANN-DT}         & Interpolated sample outputs                                                                                                          & \begin{tabular}[c]{@{}l@{}}Classification\\ Regression\end{tabular}                                            & Pedagogical & Tree-based                                                                      &  \citet{schmitz1999ann}                                                                 \\ \hline
\textbf{DecText}     &  \begin{tabular}[c]{@{}l@{}}Improved splitting (SetZero) \\ and  discretization\end{tabular}                                                                                                                 & Classification                                                       &  Pedagogical & Tree-based                                                                        &  \citet{boz2002extracting}                                                                 \\ \hline
\textbf{REx}            & Genetic algorithm                                                                                                          & Classification                                                       & Pedagogical & Rule-based                                                                     &  \citet{kamruzzaman2010rex}                                                                           \\ \hline
\textbf{GEX}            & Genetic algorithm                                                                                                            & Classification                                                       & Pedagogical& Rule-based                                                                          &  \citet{markowska2004discovering}                                                                        \\ \hline
\textbf{DeepRED}      &  Tree induction                                                                                               & Classification                                                       & Decompositional [ANN] & Rule-based                                                                        &  \citet{zilke2016deepred}                                                                           \\ \hline
\textbf{SUBSET}        & Feature Importance                                                                                                & Classification                                                       &Decompositional [SVM]  & Rule-based                                                                      &  \citet{biswas2017rule}                                                                           \\ \hline
\textbf{MofN}           & Feature Importance                                                                                                & Classification                                                       & Decompositional& Rule-based                                                                       &  \citet{setiono2014mofn}                                \\ \hline
\textbf{Knowledgetron}             & Feature Selection (heuristic search)                                                                                                   & Classification                                                       & Decompositional [ANN]& Rule-based                                                                     &   \citet{fu1994rule}                                                                        \\ \hline
\textbf{NeuroRule}      & Feature Importance                                                                                                & Classification                                                       & Decompositional [ANN] & Rule-based                                                                            &  \citet{setiono1995neurorule}                                                                         \\ \hline
\textbf{RX}             & Feature Importance                                                                                                & Classification                                                       & Decompositional [ANN]& Rule-based                                                                                                &  \citet{biswas2017rule}                                                     \\ \hline
\textbf{NeuroLinear}    & \begin{tabular}[c]{@{}l@{}}Discretization of hidden \\ unit activation values \end{tabular}                                                                                                 & Classification                                                       & Decompositional [ANN]& Rule-based                                                                         &  \citet{setiono1997neurolinear}                                                                            \\ \hline
\textbf{full-RE}        & Feature selection                                                                                                 & Classification                                                       & Decompositional [ANN]& Rule-based                                                                        &  \citet{biswas2017rule}                                                                       \\ \hline
\textbf{FERNN}          & Feature Importance                                                                                                & Classification                                                       & Decompositional [ANN]& Tree  \&  Rule-based                                                                                    &  \citet{biswas2017rule}                                                        \\ \hline
\textbf{CRED}           & Tree induction                                                                                                & Classification                                                       & Electic [ANN] & Tree  \&  Rule-based                                                               &  \citet{zilke2016deepred}                                                                           \\ \hline
\textbf{ANNT}           & Tree induction                                                                                                  & Classification                                                       & Decompositional [ANN] & Rule-based                                                                          &  \citet{biswas2017rule}                                                                          \\ \hline
\textbf{GRG}            & Clustering                                                                                                & Classification                                                       & Model-specific & Rule-based                                                                        &  \citet{odajima2008greedy}                                                                          \\ \hline
\textbf{E-Re-RX}        & Ensemble concept                                                                                         & Classification                                                       & Decompositional [ANN]& Rule-based                                                                      &    \citet{hayashi2013neural}                                                                          \\ \hline
\textbf{X-TREPAN}       & Tree induction                                                                                                & \begin{tabular}[c]{@{}l@{}}Classification\\ Regression\end{tabular}                                            & \begin{tabular}[c]{@{}l@{}}Model-specific \\ (Decompositional [ANN]) \end{tabular} & Tree-based                                                                      &  \citet{biswas2017rule}                                                                    \\ \hline
\textbf{RuleFit}        & Predictive Learning                                                                                                & \begin{tabular}[c]{@{}l@{}}Classification\\ Regression\end{tabular}                                            &Decompositional [RF] & Linear Model \& Rule-based                                                           &  \citet{friedman2008predictive}                                                                             \\ \hline
\textbf{Node harvest}   & Feature Selection                                                                                                 & \begin{tabular}[c]{@{}l@{}}Classification\\ Regression\end{tabular}                                            & Decompositional [RF] & Rule-based                                                                        &  \citet{meinshausen2010node}                                                                         \\ \hline
\textbf{DHC}            & Scoring Function                                                                                               & \begin{tabular}[c]{@{}l@{}}Classification\\ Regression\end{tabular}                                            & Decompositional [RF] & Rule-based                                                                     &  \citet{mashayekhi2017rule}                                                                        \\ \hline
\textbf{SGL}            & Feature Importance                                                                                                & \begin{tabular}[c]{@{}l@{}}Classification\\ Regression\end{tabular}                                            & Decompositional [RF] & Rule-based                                                                                   &  \citet{mashayekhi2017rule}                                                                  \\ \hline
\textbf{multiclass SGL}           & Scoring Function                                                                                               & Classification                                                       & Decompositional [RF]& Rule-based                                                                           &  \citet{mashayekhi2017rule}                                                                          \\ \hline
\textbf{SVM+Prototypes} &  Clustering                                                                                                  & Classification                                                       & Decompositional [SVM] & Rule-based                                                                        &  \citet{martens2007comprehensible}                                           \\ \hline
\textbf{Fung}           & Clustering                                                                                                 & Classification                                                       & Decompositional [SVM]  & Rule-based                                                                       &   \citet{fung2008rule}                                                                             \\ \hline
\textbf{SQRex-SVM}      & Sequential Covering                                                                                                 & Classification                                                       & Decompositional [SVM] & Rule-based                                                                      &   \citet{barakat2007rule}                                                                               \\ \hline
\textbf{ALBA}           & Active Learning                                                                                                 & Classification                                                       & Decompositional [SVM] & Rule-based                                                                        &  \citet{martens2009decompositional}                                                            \\ \hline
\end{tabular}
\end{adjustbox}

\label{overviewrule}
\end{table*}

\subsubsection{Pedagogical Decision Rules}
The \emph{\ac{brianne}} algorithm was originally proposed for \ac{ann}. The algorithm measures the relevance of a specific feature for some output. This measure is then used to build rule conditions. The measure determines the features and the features' values of a condition \citep{biswas2017rule}.

The \emph{\ac{via}} \citep{zilke2016deepred} algorithm aims at finding provably correct rules. This is done by applying an approach similar to sensitivity analysis. 
The \emph{\ac{biore}} \citep{biswas2017rule} algorithm applies a sampling based approach that generates all possible input combinations and asks the black box for their predictions. Based on that, a truth table is build on top of which an arbitrary rule-based algorithm can be applied to extract rules. 

There is a class of rule extraction approaches utilizing genetic programming, which is motivated by Darwin's theory on survival of the fittest. 
The \emph{\ac{grex}} algorithm for instance chooses the best rules from a pool of rules generated by the black box and combines them with a genetic operator \citep{martens2009decompositional, martens2007comprehensible}. The \emph{REX} algorithm was originally proposed to extract rules from \acp{ann}. However, it can be applied to any black box. It uses genetic programming to extract fuzzy rules \citep{ninama2013ensemble}. The \ac{gex} algorithm is a genetic programming algorithm that uses sub-populations based on the number of different classes that are present in the training data. The algorithm extracts propositional rules from the underlying black box \citep{ martens2009decompositional}.

The \emph{\ac{stare}} algorithm is based on breadth first search. Furthermore, the input data is permuted in order to produce a truth table that is used for rule extraction.
The \emph{\ac{refne}} algorithm uses \ac{ann} ensembles to generate instances that are then used to build decision rules.

The \emph{\ac{bur}} algorithm is based on the gradient boosting machine. The algorithm first learns and then prunes the decision rules \citep{ninama2013ensemble}. \emph{Iter} \citep{martens2009decompositional} is a sequential covering algorithm that learns one rule at a time. It randomly generates extra data instances and uses the black box as an oracle to label these data instances. 
The \emph{\ac{osre}} algorithm was originally proposed for \acp{ann} and \acp{svm}. It converts the given input to a desired format and uses activation responses to extract rules \citep{etchells2006orthogonal}. \emph{Minerva} \citep{martens2008rule} is a sequential covering algorithm that uses iterative growing to extract decision rules from a black box. 

The \emph{\ac{rxren}} \citep{biswas2017rule} algorithm uses reverse engineering for input feature pruning. Input features are pruned when their temporary omission does not change the classification output significantly. 
The \emph{\ac{rxncm}} algorithm is based on a modification applied to \ac{rxren}. \ac{rxncm} uses classified and misclassified data to figure out per class ranges for significant features. It also differs from \ac{rxren} in the black box pruning step. Input features are only pruned when their absence increases prediction accuracy. Based on that, rules are extracted from the black box \citep{biswas2017rule}. 

The \emph{\ac{beta}} algorithm applies two level decision sets on top of black box predictions. The first level specifies the neighborhood of a rule (subspace descriptor) and the second level introduces decision rules that are specific for each region \citep{lakkaraju2017interpretable}.

\subsubsection{Decompositional Decision Rules from \acp{svm}}
There are techniques that use a \ac{svm} as the underlying prediction model and that extract rules from the \ac{svm}. The \emph{SVM+Prototypes} \citep{martens2007comprehensible} approach first separates the two classes using a \ac{svm}. Per subset, the algorithm uses clustering and finds prototypes for each cluster. It further uses support vectors to create rule defining regions. 

The rule extraction technique proposed by Fung is limited to linear \acp{svm} and extracts propositional rules \citep{martens2009decompositional}. \emph{SQRex-SVM} uses a sequential covering algorithm that is only interested in correctly classified support vectors. The approach is limited to binary classification since only rules for the desired class are extracted \citep{martens2009decompositional}. The \emph{\ac{alba}} \citep{martens2009decompositional} extracts rules from \acp{svm} by using support vectors. The algorithm re-labels the input data based on the predictions made by the \ac{svm} and generates extra data close to the support vectors that are also labeled by the underlying \ac{svm}. 

\subsubsection{Decompositional Decision Rules from \acp{ann}}
There are techniques that use \acp{ann} or \acp{dnn} as the underlying prediction model and extract rules from the \ac{ann}.
The \emph{SUBSET} method finds subsets of features that fully activate hidden and output layers. These subsets are then used for rule extraction from \acp{ann} \citep{biswas2017rule}. The \emph{MofN} algorithm is an extension of the SUBSET algorithm. The goal is to find connections with similar weights. This can be done using clustering. The extracted rules label an unseen data instance with the rule label when M of the N conditions of the rule are met \citep{biswas2017rule} \citep{zilke2016deepred}. The \emph{\ac{kt}} algorithm extracts rules for each neuron based on the input neurons that are responsible for another neuron's activation. This is the basis for the final rule extraction step \citep{biswas2017rule} \citep{zilke2016deepred}. The \emph{NeuroRule} algorithm first applies pruning to the \ac{ann} in order to remove irrelevant connections between neurons. An automated rule generation method is then used to generate rules that cover the maximum number of samples with the minimum number of features in the condition of the rule. However, continuous data needs to be discretized first \citep{biswas2017rule}. The \emph{RX} algorithm can only be applied to \acp{ann} with just one hidden layer. First, the \ac{ann} is pruned to remove irrelevant connections between neurons. Then, neuron activations are used to generate rules. Another limitation of this approach is the restriction to discrete data. The \emph{NeuroLinear} algorithm extracts oblique decision rules from the underlying \ac{ann} \citep{biswas2017rule}. The \emph{full-RE} algorithm first learns intermediate rules for hidden neurons. These intermediate rules are linear combinations of the input neurons that activate a specific hidden neuron. Then a linear programming solver is used to learn the final rules \citep{biswas2017rule}. The \emph{\ac{fernn}} algorithm aims at identifying relevant hidden and input neurons. It uses C4.5 to construct a decision tree on top of the \ac{ann}. This decision tree is the basis for the rule extraction step\citep{biswas2017rule}. The \emph{\ac{cred}} algorithm first builds a decision tree for each output neuron by just using the hidden neurons. Then, input neurons are also incorporated into these decision trees. Finally, rules are extracted from these decision trees \citep{zilke2016deepred}. The \emph{ANNT} algorithm aims at reducing the number of rules that are extracted \citep{biswas2017rule}. The \emph{\ac{grg}} algorithm is restricted to discrete input features and \acp{ann} consists of only one hidden unit \citep{biswas2017rule}. The \emph{\ac{ererx}} algorithm uses multiple\eg two \acp{ann} to extract decision rules \citep{biswas2017rule}. 

\subsubsection{Decompositional Decision Rules from \acp{rf}}
There are approaches that use a \ac{rf} as the underlying prediction model and that solely extract rules from the \ac{rf}. The \emph{RuleFit} algorithm uses predictive learning to build a linear function on top of the rules extracted from the \ac{rf}. The linear function consists of rule conditions and input features. The algorithm uses lasso penalty for the coefficients and the final linear combination \citep{mashayekhi2017rule}. 

The \emph{Node harvest algorithm} starts with an initial set of rules and adds additional rules to this set based on two criteria. In order for a rule to be added to the set, a rule has to satisfy a threshold set for the number of features used in the condition of the rule and a threshold set for the number of samples it covers. The algorithm further finds weights for all the selected rules while minimizing a loss function \citep{mashayekhi2017rule}. 

The \emph{\ac{dhc}} algorithm removes rules from the set of rules generated by the \ac{rf} based on a scoring function. The scoring function assigns each rule a score based on specific accuracy measures and complexity measures for interpretability that were defined beforehand \citep{mashayekhi2017rule}. The \emph{\ac{sgl}} method first groups rules based on their underlying tree in the \ac{rf}. It further finds a weight vector over the generated rules from the \ac{rf}. This weight vector is sparse. Groups are assigned a weight as well. Based on the weights, single rules or whole groups are eliminated from the final decision rules \citep{mashayekhi2017rule}. The \emph{\ac{msgl}} method functions in the same way as the \ac{sgl} method but uses multi-class sparse group lasso instead of sparse group lasso. In contrast to all the previously mentioned decompositional \ac{rf} methods, this method can only be used for classification tasks but not for regression tasks \citep{mashayekhi2017rule}.

\subsubsection{Pedagogical Decision Trees} \emph{TREPAN} uses a hill climbing search process for its tree construction. It further uses a gain ratio criterion to find the best M-of-N splits for each node. Whenever there are not enough training instances available for a certain split, additional data instances for training are generated. However, TREPAN is limited to binary classification \citep{craven1996extracting}. 

The \emph{ANN-DT} algorithm was initially proposed for tree induction from \acs{ann}. However, binary decision trees can be extracted from any block box using ANN-DT, which is an algorithm similar to \ac{cart}. It uses a sampling strategy to create artificial training data that receives its labels from the underlying black box \citep{ninama2013ensemble}. 

\emph{DecText} was originally proposed for \acp{ann} but it can be applied to any black box model. The method uses a novel decision tree splitting criterion for building the decision tree \citep{ninama2013ensemble}.

\subsubsection{Decompositional Decision Trees}
\citet{barakat2004learning} use the C4.5 algorithm and apply it to \acp{svm} in order to extract decision trees from \acp{svm}. \emph{X-TREPAN} \citep{biswas2017rule} is an extension of the TREPAN algorithm in order to extract decision trees from \acp{ann}.

\subsubsection{Pedagogical Decision Tables}
The \emph{KDRuleEX} \citep{biswas2017rule} method uses a genetic algorithm to create new training data when not enough data instances are given for a certain split. The final output is a decision table.

\subsubsection{The Use Case} 
\begin{figure*}
\centering
   
     \includegraphics[width=4.5in]{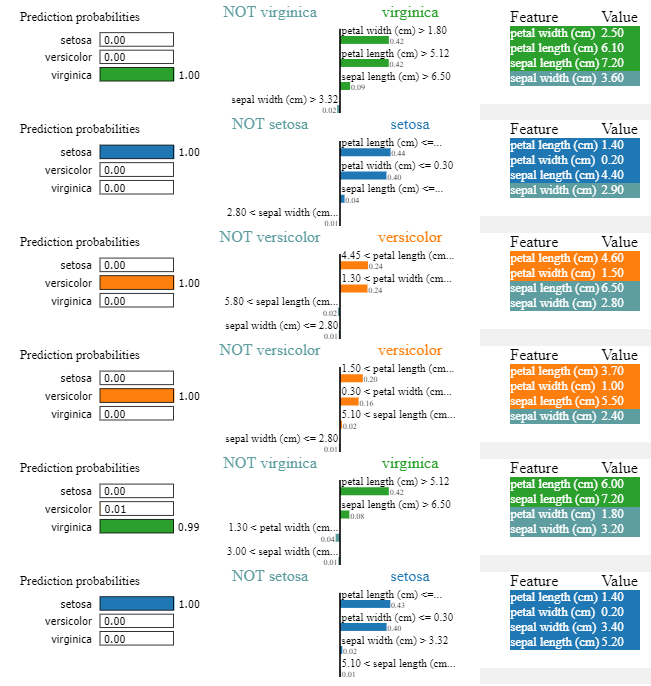}
    \caption{Six explanations for the three classes (virginica, setosa and versicolor) that were picked by sp-LIME. On the right side, the features and the feature's values are displayed.}
    \label{fig:spa}

\end{figure*}

Submodular-pick LIME is model-agnostic, post-hoc and a global surrogate which expands the LIME method by choosing a few instances that explain the global decision boundaries of the model best. Sp-Lime needs the black box model and a sample of the input data set that represents the entire data space. The approach outputs a single feature contribution overview for each of the chosen instances to be representative for the feature space. Thereby, the user can evaluate each of them separately and deduct from the set of explanations the understanding of the model globally. From the small number of local explanations, the user gets an impression of the decision process of the model for different feature distributions. We chose six explanations of the three classes (see Figure \ref{fig:spa}). From the instances explained, one can see which features affect which class in which direction (negative or positive), an assertion that holds globally since the task at hand is simple. This explanation seems to be kind of global since it provides only a small sample of the global behaviour depending on how many explanations are chosen. The more explanations are chosen, the more difficult it may be for the user to understand the model. 

\begin{figure}[t]
    \centering
    \begin{tikzpicture}[scale=0.95]
        \node[rblock, text width=1.8cm, minimum height=1.2cm] at (0,0) (A) {\small Learning Algorithm};
        \node[rblock, text width=1.8cm, minimum height=1.2cm] at (3,0) (M) {\small Black Box Model $b$};
        \node[rblock, text width=1.8cm, minimum height=1.2cm] at (6,0) (P) {\small Prediction};
        \node[rblock, text width=1.8cm, minimum height=1.2cm] at (9,0) (S) {\small Surrogate $w$};
        \node[rblock, text width=1.85cm, minimum height=1.2cm] at (12,0) (E) {\small Explanation $e$};
        
        \filldraw ($(M.north)+(0,.4)$) circle (.8mm) node (c) {};

        \draw[ablock] (A) -- (M);
        \draw[ablock] (M) -- (P);
        \draw[ablock] (P) -- (S);
        \draw[ablock] (S) -- (E);
        \draw[ablock, latex-] (A.west) -- +(-.5,0) node[above, rotate=90, text width = 1.5cm, text centered] {\small Training Data};
        \draw[ablock, latex-] (M.north) -- +(0,.6) node[above] {\small Test Data};
        \draw[ablock, shorten <= -2] (c) -| (S.north);
    \end{tikzpicture}
    \caption{Fitting a local surrogate according to \protect\Fig{fig:relations}(f).}
    \label{fig:local-surrogate}
\end{figure}
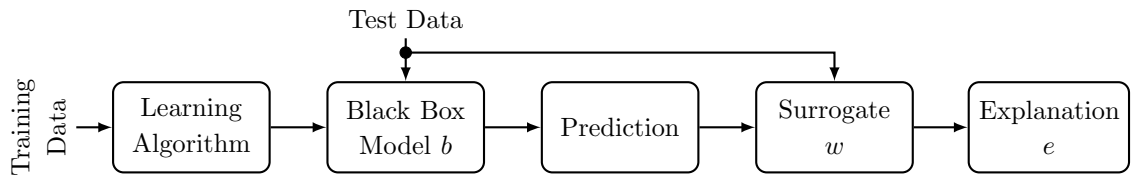

\subsection{Local Surrogates}
In this section, we focus on approaches that provide a local surrogate model being extracted or being created from a given SML model. In contrast to the global surrogates discussed above, the local surrogate is valid only for a specific data instance and its close vicinity. 
LIME---one of the currently most popular explainability approaches---belongs to the class local surrogates. The overview of local surrogate models is listed in Table \ref{localSurrogates}. 

\subsubsection{Decision Rules}
\citet{ribeiro2018anchors} introduce anchorLIME. The approach generates explanations in the form of independent IF-THEN rules like IDS \citep{phillips2017interpretable}. \ac{muse} \citep{lakkaraju2019faithful} is a model-agnostic framework that provides understanding of a black box model by explaining how it behaves in the sub-spaces defined by the features of interest. The framework learns decision sets for a specific region in the feature space.
\ac{lore} \citep{lore} learns a local interpretable model on a synthetic neighborhood generated by a genetic algorithm. It derives a meaningful explanation from the local interpretable predictor which consists of a decision rule set of counterfactual rules.

\subsubsection{Linear models}
The explanation frameworks \textit{Model Explanation System} (MES) and \textit{Local Interpretable Model-Agnostic Explanations} (as well as its variations) are local model approximation approaches. 
MES \citep{turner} is a general framework which explains the predictions of a binary black box classifier. MES explains the outcome (predictions) made by a binary classifier. \ac{maple} \citep{maple} uses a local linear modeling approach with a dual interpretation strategy of random forests. LIME \citep{ribeiro2016should} describes a particular prediction made by any black box classifier by taking samples from the locality (neighborhood) of a single prediction. The explanation is valid only on a local level. The samples are presented in the form of an explanation model\eg linear model, decision tree, or rule list.
\textit{explainVis} interprets the individual predictions as local gradients which are used to identify the contribution of each feature. With \textit{explainVis}, it is possible to compare different prediction methods \citep{robnik2008explaining}.
Another approach for interpreting predictions is SHAP (SHapley Additive exPlanation). SHAP explains a particular output by computing an interpretation contribution to each input of a prediction. SHAP fulfils the three characteristics \textit{Local accuracy}, \textit{Missingness} and \textit{Consistency}. \textit{Local accuracy} determines the same output for an approximated model and the original model if they receive the same input. Missingness relates to the fact of missing features. If a feature is absent, it should not have any impact on the output. \textit{Consistency} ensures that if a feature increases or stays the same, the impact of this feature should not decrease. With the help of SHAP, there is a unique solution in the class of meaningful properties \citep{lundberg2017unified}.

\begin{table}

\centering
\caption{Overview of local surrogate approaches}
\begin{adjustbox}{width=5in}
\renewcommand{\arraystretch}{1.25}
\begin{tabular}{|l|l|l|c|}
\hline
\textbf {Approach}   &
 \textbf{Learning Task}  &
\textbf{Model} & 
\textbf{References}  \\ \hline \hline
LIME& Classification/Regression&Linear Model & \citet{ribeiro2016should}\\ \hline
aLime & Classification&Linear Model & \citet{ribeiro2018anchors}\\ \hline
MES & Classification&Linear Model & \citet{turner}\\ \hline
MUSE& Classification&Linear Model & \citet{lakkaraju2019faithful}\\ \hline
LORE& Classification&Linear Model & \citet{lore}\\ \hline
MAPLE& Classification&Linear Model & \citet{maple}\\ \hline
Kernel SHAP &Classification & Linear Model & \citet{lundberg2017unified} \\ \hline
Linear SHAP &Classification/Regression & Linear Model & \citet{lundberg2017unified} \\ \hline

\end{tabular}
\end{adjustbox}
\label{localSurrogates}
\end{table}

\subsubsection{Use Case}
\label{sec:casestudy_localsurrogate}
\begin{figure*}[h!]
    \centering
    \includegraphics[width=\textwidth]{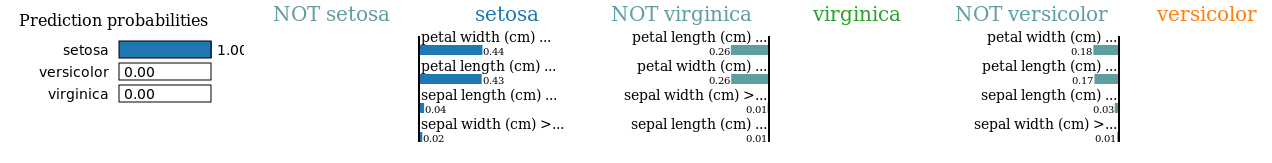}
    \caption{LIME explanation for the \textit{virginica} instance. The instance was classified as \textit{virginica} and the most influential features therefore is that \textit{petal width} is smaller than 0.30 cm}
    \label{fig:LIMEClass}
\end{figure*}

\acs{lime} approximates the decision of the black box linearly in a local neighborhood of a given instance. As input \acs{lime} requires a black-box model and the instance of interest. Figure \ref{fig:LIMEClass} illustrates the explanation for a \textit{setosa} instance that is generated by LIME. This overview illustrates for the user which features influenced the likelihood of each of the three possible classes in this specific case. The user can see clearly why this instance was classified the way it was based on the specific feature values.

\begin{figure*}[h!]
    \centering
    \includegraphics[width=\textwidth]{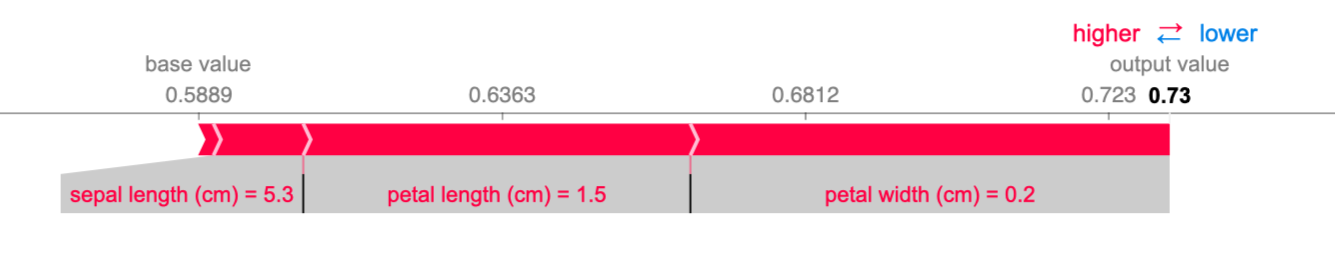}
    \caption{Shapley values plotted for setosa output: The plots illustrate that, starting from a base value of 0.5889 for the probability of class \textit{setosa}, the features \textit{petal length} and \textit{petal width} contributed the most to the final probability of 0.73 (output value). A blue arrow in the other direction would tell us that a feature contributed negatively to the classification probability. 
}
    \label{fig:shap}
\end{figure*}

Another approach is SHAP (Kernel SHAP or Linear SHAP) \citep{lundberg2017unified}. It is a local and post-hoc approach which extracts the feature importance for a given prediction. 
As input we take the trained model and the desired instance for which we want an explanation. The SHAP approach relies on the fact that a prediction can be written as the sum of bias and single feature contributions. The feature contributions (shapley values) are extracted by marginalizing over every feature to analyze how the model behaves in its absence. This yields the result of an explanation in an overview (see Figure \ref{fig:shap}).

\section{Explanation Generation}
\label{sec:explanations_global}

In this section, we describe approaches that directly can generate an explanation (either local or global). The difference between the surrogate models is that the explanation is directly inferred from the black box model. 

\begin{table}[ht]
\centering
\caption{Overview of global explanation generation methods}
\label{overview_global_induction}
\begin{adjustbox}{width=\textwidth,center=\textwidth}
\begin{tabular}{|l|l|l|l|l|l|l|l|l|l|l|}
\hline
\textbf{Approach} & 
\multicolumn{1}{l|}{\textbf{\begin{tabular}[c]{@{}l@{}}Learning \\ Task\end{tabular}}} &
\multicolumn{1}{l|}{\textbf{\begin{tabular}[c]{@{}l@{}}Input Model\end{tabular}}} & \multicolumn{1}{l|}{\textbf{\begin{tabular}[c]{@{}l@{}}Output Model\end{tabular}}}  &  
\multicolumn{1}{l|}{\textbf{\begin{tabular}[c]{@{}l@{}}References\end{tabular}}}\\ \hline

RF Feature Importance & Classification & Model-specific & Tree-based & \citet{hall2017machine}\\ \hline

Sparsity Constraints  & - & - & - &  \citet{ustun2014methods}  \\ \hline

Correlation Graph  & -  & Model-agnostic & - & \citet{hall2017ideas}\\ \hline

Residual Analysis  & - & Model-agnostic & - &  \citet{hall2017ideas}\\ \hline

Autoencoder  & - & Model-agnostic & - & \citet{hall2017ideas}\\ \hline

PCA  & - & Model-agnostic & - &  \citet{hall2017ideas}\\ \hline

MDS & - & Model-agnostic & -  &  \citet{hall2017ideas}\\ \hline

t-SNE  & - & Model-agnostic & - & \citet{hall2017ideas}\\ \hline

Nomograms & Classification & Model-agnostic & Linear Model   & \citet{robnik2007explaining}\\ \hline

SOM  & - & Model-agnostic & - & \citet{martens2008rule}\\ \hline

Quasi Regression & Classification & Model-agnostic & - & \citet{jiang2002quasi}\\ \hline

EXPLAINER global & Classification & Model-agnostic & - & \citet{subianto2007understanding}\\ \hline

GSA & Classification & Model-agnostic & - & \citet{cortez2011opening}\\ \hline

GOLDEN EYE & Classification & Model-agnostic & - & \citet{henelius2014peek}\\ \hline

GFA & Classification & Model-agnostic &  - & \citet{adler2016auditing}\\ \hline

ASTRID & Classification & Model-agnostic &  - & \citet{henelius2017interpreting}\\ \hline

PDP & Classification & Model-agnostic &  - & \citet{goldstein2015peeking}\\ \hline

IME  & \begin{tabular}[c]{@{}l@{}}Classification\\ Regression\end{tabular} & Model-agnostic & -    & \citet{bohanec2017explaining}\\ \hline

Monotonicity Constraints & - & - & - &  \citet{freitas2014comprehensible}\\ \hline

Prospector & Classification & Model-agnostic & -   & \citet{krause2016interacting}\\ \hline

Leave-One-Out & Classification & Model-agnostic & -   & \citet{strumbelj2010explanation}\\ \hline

\end{tabular}
\end{adjustbox}
\end{table}

\begin{figure}
    \centering
    \begin{tikzpicture}[scale=0.95]
        \node[rblock, text width=1.8cm, minimum height=1.2cm] at (0,0) (A) {\small Learning Algorithm};
        \node[rblock, text width=1.8cm, minimum height=1.2cm] at (3,0) (M) {\small Black Box Model $b$};
        \node[rblock, text width=1.85cm, minimum height=1.2cm] at (9,0) (E) {\small Explanation $e$};

        \draw[ablock] (A) -- (M);
        \draw[ablock] (M) -- (E);
        \draw[ablock, latex-] (A.west) -- +(-.5,0) node[above, rotate=90, text width = 1.5cm, text centered] {\small Training Data};
    \end{tikzpicture}
    \caption{Generating global explanations according to \protect\Fig{fig:relations}(b).}
    \label{fig:global-explanator}
\end{figure}
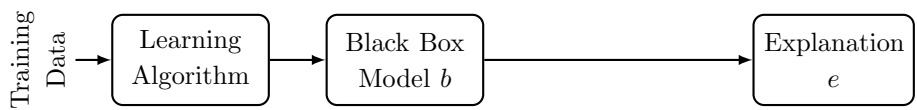

\subsection{Global Explanation Generation}
In this section, global explanators are reviewed. As depicted in \Fig{fig:global-explanator}, they are independent of certain model predictions and try to reveal certain properties of the black box model. The global explanation generation approaches are listed in Table \ref{overview_global_induction}.

\subsubsection{Feature Importance}
\label{sec:explanations_global_feature}
\emph{\ac{rf} Feature Importance} \citep{hall2017machine} is limited to decision trees and ensembles of decision trees. Per split in the decision tree, the information gain is assigned to the splitting feature as its importance measure. This importance measure can be accumulated per feature over all trees. Other feature importance measures for tree-based methods are the feature's depth in the tree or the total number of instances that are used for classification \citep{freitas2014comprehensible}.

\emph{Quasi Regression} \citep{jiang2002quasi} can be used to visualize the contributions to a black box function of different subsets of input features. The black box function is expanded in an orthonormal basis with an infinite number of coefficients. These coefficients can be estimated using the Monte Carlo method.

\emph{Explainer (global)} \citep{subianto2007understanding} is an approach that provides global insights for the importance of a feature. Each feature is assigned a weight that reflects the feature's overall influence. However, this approach is restricted to discrete data.

\emph{\ac{gsa}} \citep{cortez2011opening} introduces a visualization approach to explain decisions made by black box models that is based on a sensitivity analysis method. It measures the effects on the output when the input features are varied through their range of values. \ac{gsa} can consider multiple features at a time. Sensitivity measures used are range, gradient, and variance. Feature importances are plotted in a bar plot and the contribution of a given input feature or a pair of input features is plotted in a \ac{vec} Plot or a \ac{vec} Surface and Contour Plot.

The \emph{Golden Eye} \citep{henelius2014peek} method first finds groupings of features that indicate important feature interactions and associations between features that are exploited by the underlying classifier. This is done without considering the internal structure of the classifier or the distribution of the input data. The classifier is considered a black box. Further Golden Eye uses randomization of input features to figure out feature importance.

The \emph{\ac{gfa}} approach uses an obscuring technique to find indirect influence of the input features without retraining the underlying black box model. It outputs a plot of features and the feature's influences as well as a feature ranking based on their influence on the target \citep{adler2016auditing}.

\emph{\ac{astrid}} finds groupings of feature interactions that describe the underlying data distribution. It is a top-down greedy algorithm based on statistical significance testing to find the maximum cardinality grouping. These groupings reveal associations between features and feature interactions used by the underlying classifier. Features in the same group interact; features in different groups do not interact \citep{henelius2017interpreting}.

\emph{\ac{pdp}s} display the average prediction of the black box when an individual feature is varied over its range. The underlying idea of partial dependence aims at showing how an individual feature affects the prediction of the global model. An individual feature's relationship with the target is visualized in \acp{pdp} \citep{goldstein2015peeking}.

\emph{Prospector} uses an extension of \ac{pdp} to visualize how features affect a prediction \citep{krause2016interacting}. \ac{pdp} are extended by a partial dependence bar that shows a colored representation of the prediction value over the range of input values that a certain feature can take. Prospector also uses a novel feature importance metric that outputs a feature's importance number on an instance basis. Furthermore, Prospector provides actionable insight by supporting tweaking of feature values. However, it is restricted to single class predictions and also limited insofar as it does not take interactions between features into account.

The \emph{\ac{ime}} method aims to find the feature importance while also considering feature interactions. Interactions are considered by randomly permuting some feature values and measuring the prediction difference. \ac{ime} further uses the Shapley value from coalitional game theory to assign each feature a contribution value \citep{bohanec2017explaining}. 

The \emph{Leave-One-Out} approach learns a feature's contribution by omitting the feature from the prediction process. It considers all different subsets of features which enables the consideration of feature interactions. Each feature is assigned a local contribution value that can be aggregated and averaged to also assign each feature a global contribution value. However, the approach only works with low dimensional data or when a feature selection process is applied beforehand \citep{strumbelj2010explanation}.

\citet{vstrumbelj2014explaining} describe a general approach for explaining how features contribute to classification and regression models' predictions by computing the situational importance of features (local or global). 

\emph{iForest} \citep{zhao2019iforest} is an interactive visualization system that helps users interpret random forests model by revealing relations between input features and output predictions, hence enabling users to flexibly tweak feature values to monitor prediction changes. \emph{RuleMatrix} is an interactive visualization technique that supports users with restricted expertise in machine learning to understand and validate black box classifiers. \emph{\ac{ava}} \citep{bhatt2019towards} is a value attribution technique that provides local explanations but also detects global patterns.


   
\subsubsection{The Use Case}  
\begin{figure*}[h!]
\centering
   \begin{minipage}{0.47\linewidth}
   \centering
     \includegraphics[scale=0.35]{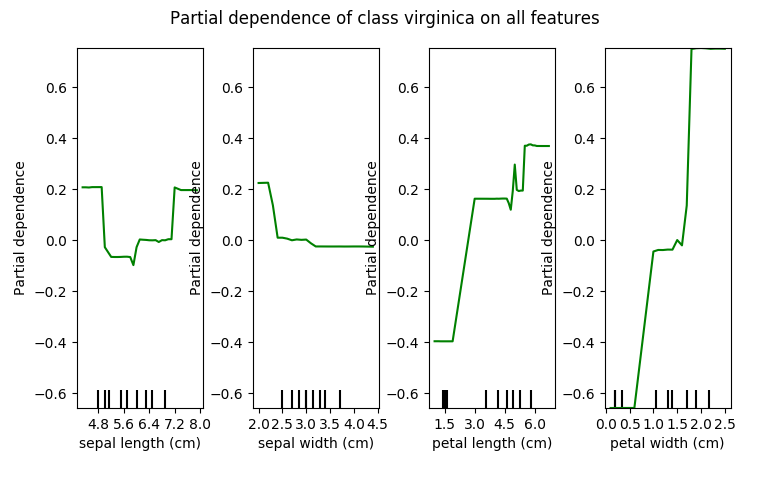}
    \caption{Explanations per feature}
    \label{fig:pdp1}
     \end{minipage}
 \begin{minipage}{0.47\linewidth}
 \centering
    \includegraphics[scale=0.35]{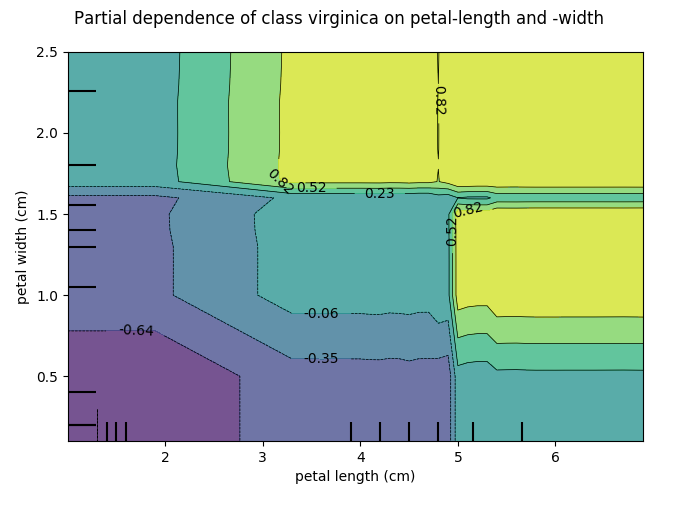}
    \caption{Explanation against two features}
    \label{fig:pdp2}
      \end{minipage}
        \vspace{4.00mm}

\end{figure*}

\ac{pdp} is a global, model-agnostic and post-hoc explanation generation approach. PDPs can be implemented on all trained models, more or less efficiently. Additionally, we need the data set to marginalize over the features we are not interested in. PDPs are presented as plots in which we either plot a classification against its partial dependence on one feature (see Figure \ref{fig:pdp1}) or two features against each other with a map plot (see Figure \ref{fig:pdp2}).
From a PDP, the correlation a feature has to the classification of a certain class can be extracted. The map plot illustrates the virginica probability and the interaction of petal width and petal length. The plot shows the increase in virginica probability if petal length is greater than 3 cm and petal length is greater than 1.8 cm.

Another approach in this category is the Global Feature Importance. It is a model-agnostic, post-hoc explanation generation approach similar to the treeinterpreter \citep{treeinterpreter}, but global. Global Feature Importance simply returns a list assigning an importance score to each feature that, when taken together, sum up to 1. This importance does not tell the user much about the effect of the features but merely whether or not the model considers them strongly. This is a first good entry for explainability but it needs to be combined with other explainability approaches to gain more detailed insights.

\begin{table} 
    \caption{Global Feature Importance}
        \centering
    \begin{tabular}{| c | c |}
    \hline
    \textbf{Feature} & \textbf{Importance} \\ \hline \hline
    sepal length (cm) & 0.47174951 \\ \hline
    sepal width (cm)  & 0.40272703 \\ \hline
    petal length (cm) & 0.10099771 \\ \hline
    petal width (cm)  & 0.02452574 \\ \hline
    \end{tabular}
        \label{GlobalFeatureImportance}
\end{table}

Table \ref{GlobalFeatureImportance} illustrates that the feature importance provides that \textit{sepal-length} and \textit{-width} are the most decisive features which the model considers more often than the other two. 




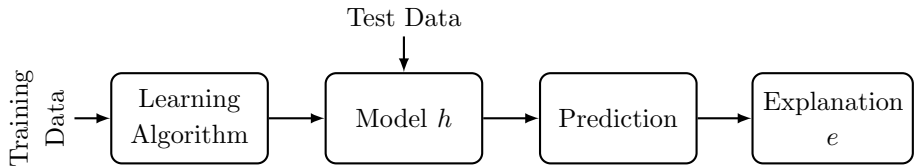
\begin{figure}[b]
    \centering
    \begin{tikzpicture}[scale=0.95]
        \node[rblock, text width=1.8cm, minimum height=1.2cm] at (0,0) (A) {\small Learning Algorithm};
        \node[rblock, text width=1.8cm, minimum height=1.2cm] at (3,0) (M) {\small Model $h$};
        \node[rblock, text width=1.8cm, minimum height=1.2cm] at (6,0) (P) {\small Prediction};
        \node[rblock, text width=1.85cm, minimum height=1.2cm] at (9,0) (E) {\small Explanation $e$};

        \draw[ablock] (A) -- (M);
        \draw[ablock] (M) -- (P);
        \draw[ablock] (P) -- (E);
        \draw[ablock, latex-] (A.west) -- +(-.5,0) node[above, rotate=90, text width = 1.5cm, text centered] {\small Training Data};
        \draw[ablock, latex-] (M.north) -- +(0,.5) node[above] {\small Test Data};
    \end{tikzpicture}
    \caption{Generating local explanations according to \protect\Fig{fig:relations}(e).}
    \label{fig:local-explanator}
\end{figure}

\subsection{Local Explanation Generation}
\label{sec:explanations_local}
Local explanations are only valid in the vicinity of a certain prediction as indicated in \Fig{fig:local-explanator}. This section reviews commonly used local explanators. The local explanation generation approaches are listed in Table \ref{overview_local_induction}.

\subsubsection{Saliency Methods}
\label{sec:explanations_local_saliency}
Saliency methods in general relate the model prediction to the feature vector by ranking the explanatory power, i.e. the salience of the individual features.

\paragraph{Feature Attribution:} 
An important class of saliency methods are feature attribution approaches that explicitly try to quantify the contribution of each individual feature to the prediction. 

\emph{Explainer} (local) provides local insight by finding a set of features that is minimal in the sense that changing the values of those features changes the prediction outcome \citep{subianto2007understanding}. This concept can be formalized and used as the basis for finding minimal sets of features. However, this approach is restricted to discrete data. 

The \emph{Local Gradients} approach learns a local explanation vector consisting of local gradients of the probability function. The local explanation vector can be mapped onto the input feature vector such that the sign of the values in the local explanation vector indicates the direction of influence of the underlying feature towards the instance and the absolute value indicates the amount of influence. When this gradient information cannot be obtained from the used prediction model, a probabilistic approximate such as parzen window, logistic regression or Gaussian Process Classification is employed \citep{baehrens2010explain}. 

The \emph{\ac{loco}} approach trains two different models, one with all input features and one with the feature of interest left out. These two models can then be compared with each other using their computed residuals. If their difference is above a certain threshold for some values of the feature, the feature is considered important in that range. This approach can be applied to the entire data (global) or just one instance (local) \citep{lei2018, hall2017machine}. 

The \emph{\ac{qii}} \citep{datta2016algorithmic} method introduces different measures for reporting the influence of input features on the prediction outcome. In order to do so, feature values are randomly perturbed and their change in the prediction outcome is measured. In order to consider correlations between input features, the causal \ac{qii} measure called Unary \ac{qii} is used. 

\emph{\ac{mcr}} derives connections between permutation importance estimates for a single prediction model, U-statistics, conditional causal effects, and linear model coefficients. Furthermore, they give probabilistic bounds for MCR by using a generalizable technique \citep{fisher2018model}.

\emph{\ac{ice}} plots are an extension of \acp{pdp}. They visualize the relationship between the target and an individual feature on an instance basis and not for the whole model. When \ac{ice} plots and \acp{pdp} are plotted in the same graph, the difference between the individual behavior of a feature and its average behavior can be revealed. Two further extensions of normal \ac{ice} plots, centered and derivative \ac{ice} plots, can discover heterogeneity and explore the presence of interacting effects, respectively \citep{goldstein2015peeking}. 
The treeinterpreter \citep{hall2017ideas} can only be applied to decision trees. It breaks down the final prediction for a specific instance into bias and feature contribution and augments the decision tree's nodes as well as the paths with this additional information.

\emph{Shapley values} \citep{shapley1951notes} are created by means of a method from coalitional game theory assuming that each feature value of the instance is a player in a game where the prediction is the payoff. Shapley values illustrate how to dispense the payoff among the features.


\emph{\ac{senn}} contain the interpretation functionality in their architecture. These models fulfill three characteristics for interpretability: explicitness, faithfulness, and stability. \textit{Explicitness} deals with the question of how comprehensible the explanations are. \textit{Faithfulness} checks whether the meaningful features predicted by the model are really meaningful. \textit{Stability} ensures the coherency of explanations for similar input data \citep{melis2018towards}.
The \emph{\ac{vem}} from \citet{hendricks2016generating} describes an image and provides information why this image got classified in this way. Therefore, the model includes a \textit{relevance loss} and a \textit{discriminative loss}. The \textit{relevance loss} is used to generate a description of an image based on visual components. The \textit{discriminative loss} uses reinforcement paradigms to focus on category specific properties.
The \emph{\ac{icm}} is an approach which generates a discriminative description for a target image. To generate the description, the model explains why the visual target belongs to a certain category.
Afterwards, the model compares the target image with a similar distractor image and attempts to describe the differences. The differences are included in the description to make it more precise \citep{vedantam2017context}. 

In case of image data being processed by neural networks, a common approach is to highlight the contribution of individual pixels for a prediction by means of a heat map. For instance, \emph{Layerwise Relevance Propagation} computes scores for pixels and regions in images by backpropagation.  
Another approach is \emph{Grad-CAM}. This approach can be applied to any CNN model to produce a heat map that highlights the parts of the image that are important for predicting the class of interest. The algorithm further focuses on being class-discriminative and having a high resolution \citep{selvaraju2017visual}.

\emph{DeepLIFT (Deep Learning Important FeaTures)} is an approach where the prediction of a black box model is backpropagated to the input feature in order to generate a reference activation. By comparing the reference activation with the predicted activation of the neural network, DeepLIFT assigns scores according to their difference \citep{shrikumar}. \emph{SmoothGrad} uses copies of a target image, adds noise to those copies and creates sensitivity maps. The average of those sensitivity maps is used to explain the classification process for the target image \citep{smilkov2017smoothgrad}. \emph{Integrated Gradients} (IG) uses interior gradients of deep networks to capture the importance of the different input values \citep{sundararajan2016gradients}.
\citet{montavon2017explaining} introduces \emph{Deep Taylor}, an explanation approach for non-linear models. In this approach, each feature is visually analyzed in heat maps which clearly show which input features led to the classification of the output.
\emph{\ac{lrp}} uses the prediction of a model and applies redistribution rules to assign a relevance score to each feature \citep{samek2017explainable}. In his article, \citet{samek2017explainable} also explains the output of black box models with the Sensitivity Analysis (SA) approach. Here, the output of a model is explained with the model's gradients. The constructed heat map visualizes which features need to be changed in order to increase the classification score.

\begin{table}[ht]
\centering
\caption{Overview of local explanation generation methods}
\label{overview_local_induction}
\begin{adjustbox}{width=\textwidth,center=\textwidth}
\begin{tabular}{|l|l|l|l|l|l|l|l|l|l|l|}
\hline
\textbf{Approach} & 
\multicolumn{1}{l|}{\textbf{\begin{tabular}[c]{@{}l@{}}Learning Task\end{tabular}}} &
\multicolumn{1}{l|}{\textbf{\begin{tabular}[c]{@{}l@{}}Input Model\end{tabular}}} &
\multicolumn{1}{l|}{\textbf{\begin{tabular}[c]{@{}l@{}} Model\end{tabular}}} &
\multicolumn{1}{l|}{\textbf{\begin{tabular}[c]{@{}l@{}}References\end{tabular}}}\\ \hline

ICE Plots & Classification & Model-agnostic &-  & \citet{goldstein2015peeking}\\ \hline

EXPLAINER local & Classification & Model-agnostic  & -  & \citet{subianto2007understanding}\\ \hline

Border Classification & Classification &\begin{tabular}[c]{@{}l@{}}Model-specific\end{tabular} & Non-linear Model  & \citet{barbella2009understanding}\\ \hline

Cloaking & Classification & Model-agnostic & - & \citet{chen2015enhancing} \\ \hline

Tweaking Recommendation & Classification &\begin{tabular}[c]{@{}l@{}}Model-specific\end{tabular} & Non-linear Model & \citet{tolomei2017interpretable} \\ \hline

Nearest Neighbor & Classification & Model-agnostic & - & \citet{freitas2014comprehensible}\\ \hline

SVM Recommendations & Classification & Model-specific & Non-linear Model & \citet{barbella2009understanding}\\ \hline

PVM & Classification & Model-agnostic & - & \citet{bien2011prototype}\\ \hline

Bayesian Case Model & Classification & Model-agnostic & - & \citet{kim2014bayesian}\\ \hline

MMD-critic & Classification & Model-agnostic & - & \citet{kim2016examples}\\ \hline

Influence Functions & Classification & Model-agnostic & - & \citet{koh2017understanding}\\ \hline

Local Gradients & Classification & Model-agnostic & - & \citet{baehrens2010explain}\\ \hline

LOCO & Classification & Model-agnostic & - &  \citet{lei2018}\\ \hline

QII & Classification & Model-agnostic & - & \citet{datta2016algorithmic}\\ \hline

SENN & Classification & - & Linear Model  & \citet{melis2018towards} \\ \hline

VEM & Classification & Model-specific & Non-linear Model   & \citet{hendricks2016generating} \\ \hline

Treeinterpreter & \begin{tabular}[c]{@{}l@{}}Classification\\ Regression\end{tabular} & Model-specific & Tree-based   & \citet{hall2017ideas}\\ \hline

ICM & Classification & Model-specific & Non-linear Model   & \citet{vedantam2017context} \\ \hline

DeepLIFT & Classification& Model-specific & Non-linear Model & \citet{shrikumar} \\ \hline
SmoothGrad & Classification& Model-specific & Non-linear Model & \citet{smilkov2017smoothgrad} \\ \hline
Interior Gradients & Classification & Model-specific &Non-linear Model & \citet{sundararajan2016gradients} \\ \hline
explainVis & Classification & Model-specific  & Linear Model & \citet{robnik2008explaining} \\ \hline
Deep Taylor  & Classification & Model-specific& Non-linear Model & \citet{montavon2017explaining} \\ \hline
LRP & Classification & Model-specific& Non-linear Model & \citet{samek2017explainable} \\ \hline
SA & Classification & Model-specific &Non-linear Model & \citet{samek2017explainable} \\ \hline

\end{tabular}
\end{adjustbox}
\end{table}

\paragraph{Attention Based Models:}

Attention based models are used to highlight the most promising parts of input features that lead to a certain output for a given task. Therefore, they make use of context vectors. A context vector is generated by mapping input values with an annotation sequence that contains promising information about preceding and following input features \citep{bahdanau2014neural}. 
However, \citet{jain2019attention} believe that attention based models do not provide meaningful explanations for model prediction. For a given output, they tried to find out whether the input features with high attention weights were responsible for the predicted outcome. Their work shows that there exists only a weak correlation between feature importance measures and learned attention weights.  
\citet{wiegreffe2019attention} question the assumptions in \citet{jain2019attention}'s work and believe that it depends on how you define an explanation. In various experiments, they show that attention approaches can be used to make model predictions interpretable and explainable.

The \emph{\ac{abs}} system summarizes a certain input text and captures the meaningful paragraphs. This data-driven approach combines a neural language model with an encoder \citep{rush2015neural}.
\emph{\acp{iabm}} describe the content of an image. Furthermore, these networks use additional layers for image interpretation. These interpretations illustrate how the description of the image was created \citep{xu2015show}.
The Encoder-Generator Framework extracts justifications for a specific decision from an input text. The generator selects possible justifications and the encoder maps these justifications to the target values \citep{lei2016rationalizing}. \emph{\ac{pj-x}} is a multi-modal explanation approach for visual decision tasks. It provides convincing explanations for an image by simultaneously highlighting the relevant parts and generating a textual justification \citep{park2016attentive}.
\citet{luong2015effective} introduces two attention based models for translation tasks: The \textit{global} and the \textit{local} approach. The \textit{global} attention mechanism takes all available words into account whereas the \textit{local} attention approach works with a small window of words for its next prediction.

\begin{table}[ht]
\centering
\caption{Overview of attention based models}
\label{overview_induction}
\begin{adjustbox}{width=\textwidth,center=\textwidth}
\begin{tabular}{|l|l|l|l|l|l|l|l|l|l|l|}
\hline
\textbf{Approach} & 
\multicolumn{1}{l|}{\textbf{\begin{tabular}[c]{@{}l@{}}Learning Task\end{tabular}}} &
\multicolumn{1}{l|}{\textbf{\begin{tabular}[c]{@{}l@{}}Input \\ Model\end{tabular}}} &
\multicolumn{1}{l|}{\textbf{\begin{tabular}[c]{@{}l@{}} Model\end{tabular}}} &
\multicolumn{1}{l|}{\textbf{\begin{tabular}[c]{@{}l@{}}References\end{tabular}}}\\ \hline

ABS & - & Model-specific & - & \citet{rush2015neural}\\ \hline

iABM & - & Model-specific  & - & \citet{xu2015show}\\ \hline

Encoder-Generator & - & Model-agnostic & - & \citet{lei2016rationalizing}\\ \hline

PJ-X & Classification & Model-specific & Non-linear Model & \citet{park2016attentive} \\ \hline

Global Attention Approach & Classification & Model-specific & Non-linear Model & \citet{luong2015effective} \\ \hline

Local Attention Approach & Classification & Model-specific & Non-linear Model   & \citet{luong2015effective} \\ \hline

\end{tabular}
\end{adjustbox}
\end{table}

\subsubsection{Counterfactual Methods}
\label{sec:explanation_local_counterfacturals}
Counterfactuals are defined by the Cambridge Dictionary as: ``thinking about what did not happen but could have happened'' \citep{cambridge}. They can be expressed in a more formal way as follows: If $\vx$ had been $\vx'$, Y would have been $y'$. For example: If the weather had been sunny instead of rainy, I would have gone for a walk instead of staying at home. A counterfactual describes an altered reality in which other facts would have lead to different results. The factual $\vx$ has the consequence $y$. However, if $\vx$ changes to the counterfactual $\vx'$, the consequence changes to $y'$.
Counterfactuals can be used as a special form of explanation for machine learning systems. In this case, $\vx$ and $\vx'$ are inputs for the machine learning model and $y$ or $y'$ are outputs or predictions made by the model. Thus the problem of finding a counterfactual explanation turns into a search problem in the feature space with the goal of finding\eg a nearby instance that leads to the prediction of a different class. A found counterfactual can be presented either by itself or as the difference from its factual to highlight the changes responsible for the difference in classification.
\citet{wachter2017counterfactual} describe counterfactual methods as meaningful data subjects that lead to a specific decision. Furthermore, they provide reasons to challenge this output and also advices as to how to receive the desired decision \citep{wachter2017counterfactual}. Counterfactual methods are very similar to inverse classification methods being discussed in \Sec{sec:explanations_local_inverse}.

The simplest method for counterfactual explanations is the search by trial and error. This approach randomly changes the feature values of data instances and stops when the desired output gets classified \citep{molnar2018guide}. 
\citet{wachter2017counterfactual} generates counterfactual explanations by searching counterfactual data points as close as possible to the original data points so that a new target is chosen. The distance can be measured with the Manhattan-distance which is weighted by the inverse median absolute deviation.
Counterfactual instances described by \citet{van2019interpretable} use a simple loss function and sparsely follow the explanation method of \citet{wachter2017counterfactual}. 
A list of various counterfactual methods is to be found separately in Table \ref{tabl:counterfactuals}.

The problem of counterfactual instances is that they suffer from a low degree of interpretability. That is why the \emph{Counterfactual Prototype} approach adds a loss term to the objective function which results in a less sparse but more interpretable output \citep{van2019interpretable}. 

\emph{SEDC} is a document-classification method which analyzes the data quality and the deficiencies of a model. It provides an improved understanding of the inner workings of a classifier which leads to better model performance and decision making \citep{martens2014explaining}.

The \emph{Growing Sphere} provides post-hoc explanations for a data instance through the comparison of an output with its closest enemy. Therefore, Growing Sphere gains useful information about the relevant features and illustrates which concepts the classifier has learned so far \citep{laugel2017inverse}. 

\emph{Local Surrogate} \citep{laugel2018defining} consists of selecting an instance $x_{border}$ which is close to the nearest black box decision border. The instance $x_{border}$ gives important information about the spatial location of the black box decision border for the instance $x$. 

The \emph{Feature Tweeking Algorithm} takes the trained tree-based ensembled model, a true negative feature vector, a cost function, and a threshold as key input components. The cost function measures the transformation process from a truly negative instance to a positive one. The positive threshold is responsible for the fine-tuning so that every single feature follows the correct path of each tree \citep{tolomei2017interpretable}. 

The \emph{optimal action extraction} (OAE) can be used for random forest classifiers, adaboost and gradient boosted trees. This approach attempts to find a feature vector so that the desired output is achieved at a minimum cost \citep{cui2015optimal}.

The \emph{Feasible and Actionable Counterfactual Explanations} (FACE) approach aims to build coherent and feasible counterfactuals by using the shortest path distances defined via density-weighted metrics \citep{poyiadzi2020face}.

\citet{russell2019efficient} proposes coherent counterfactual explanations for mixed data sets and proposes a concrete method for generating diverse counterfactuals  based upon mixed integer programming.

\begin{table}
\centering
\caption{Overview of counterfactual methods}
\label{tabl:counterfactuals}
\begin{adjustbox}{width=\linewidth}
\renewcommand{\arraystretch}{1.25}
\begin{tabular}{|l||l|l|l|l|}
\hline
\textbf {Approach}   &
\textbf{Learning Task}  &
\textbf{Input Model} & \textbf{Model} & \textbf{References}  \\ \hline \hline

Trial and Error & Classificiation & Model-agnostic & - & \citet{molnar2018guide} \\ \hline


Counterfactual Generation &  Classificiation   & Model-agnostic & - & \citet{wachter2017counterfactual} \\ \hline

Counterfactual Instances &  Classificiation  & Model-agnostic & - & \citet{van2019interpretable}\\ \hline

Class Prototypes &  Classificiation & Model-agnostic &  - & \citet{van2019interpretable}\\ \hline

LORE& Classification & Model-agnostic &- &\citet{lore}\\ \hline

FACE& Classification & Model-agnostic &- & \citet{poyiadzi2020face}\\ \hline

Coherent Counterfactuals& Classification/Regression & Model-agnostic &- & \citet{russell2019efficient}\\ \hline

SEDC & Classificiation & Model-agnostic & Linear/Non-linear Model & \citet{martens2014explaining} \\ \hline

Growing Sphere &  Classificiation & Model-agnostic & - & \citet{laugel2017inverse, laugel2018defining} \\ \hline

Feature Tweeking &  Classificiation & Model-specific & Linear Model & \citet{tolomei2017interpretable} \\ \hline

OAE &  Classificiation & Model-specific & Linear Model & \citet{cui2015optimal} \\ \hline

\end{tabular}
\end{adjustbox}
\end{table}

\subsubsection{Inverse Classification}
\label{sec:explanations_local_inverse}
Inverse classification looks at an individual instance and determines the values of that instance that need to be adjusted in order to change the instance's class \citep{barbella2009understanding}. This category supports the generation of counterfactual explanations.

\emph{Cloaking} is an approach that finds features that can be hidden from the prediction model in order to decrease the probability of belonging to a certain class. These features are called evidence counterfactual. When removed they significantly reduce the probability of an instance belonging to a certain class \citep{chen2015enhancing}. 

The \emph{inverse classification framework} presented by \citet{lash2017budget} can be applied to different classification models. This framework provides meaningful information about adapting the input values in order to change the actual output class. In addition, the framework ensures that these proposed data modifications are realistic. 

\emph{Border Classification} reports an instance's features that need to be changed in order to place that instance on the border (separating surface) between two classes. This approach is restricted to \acp{svm} since it makes use of the support vectors \citep{barbella2009understanding}.
Based on \citet{barbella2009understanding}'s \acs{svm} classification, these recommendations report an instance's most influential support vector. The method utilizes a pull measure that uses the kernel function similarity to measure the contribution of a support vector towards the prediction of an unseen instance \citep{barbella2009understanding}. This approach can also be categorized as a \emph{Prototype} which will be presented in the next section. 



\subsubsection{Prototypes and Criticism}
Nearest neighbor classifiers can be regarded as a method for providing prototypes. They do not explicitly build a model from the training data but instead use a similarity measure to figure out the nearest neighbors of the unseen data instance. These nearest neighbors vote with their own label for the label of the unseen instance. Hence, an explanation for one unseen instance differs from the explanation of another unseen instance. To avoid different explanations depending on the instance to be classified, typical instances can be found and used as prototypes. Another disadvantage of nearest neighbor classifier arises with high dimensional data. The concept of neighborhood has to be reconsidered when the data has a large number of features. One can either only regard certain feature-neighborhoods instead of the whole feature space or incorporate feature weights into the classifier \citep{freitas2014comprehensible}. 

Other approaches improve on the drawbacks of simple nearest neighbor classifiers.
The \emph{\acp{pvm}} \citep{bien2011prototype} finds a small set of prototypes that well represent the underlying data. The prototypes are found using an integer program that is approximated. These prototypes are selected to capture the full variance of the corresponding class while also discriminating from other classes. These prototypes can further be used for classification. An unseen data instance can be labeled based on the closest prototypes \citep{bien2009classification}. 

Another approach involves selecting prototypes that maximize the
coverage within the class, but minimize the coverage across them.
The \emph{Bayesian Case Model} \citep{kim2014bayesian} is an unsupervised clustering technique that learns prototypes and subspaces per cluster. Subspaces contain those features that characterize a cluster and are important to the corresponding prototype.

The \emph{\acs{mmd}-critic} approach uses prototypes together with criticism in order to put the focus on the aspects that are not captured by the model. The approach uses Bayesian model criticism together with \ac{mmd} to select prototypes. Criticism samples are scored with a regularized witness function and then selected \citep{kim2016examples}. 

\emph{ProtoDash} \citep{gurumoorthy2017protodash} is an approach for selecting prototypical examples from complex data sets that extends the work from \citet{kim2016examples} by non-negative weights for the importance of each prototype. 

The \emph{Influence Function} \citep{koh2017understanding} method approximates leaving out one training data instance and retraining the prediction model with influence functions. The method reports those training instances that were most influential for a specific instance and its prediction.

\subsubsection{The Use Case} 


The treeinterpreter approach is model-specific to decision trees and random forests. It is a local, post-hoc inspection approach that retrieves feature contributions to a final prediction in a manner very similar to the Shapley values. Here too, as input we need the trained model and the desired instance for which we want an explanation.

In case of a \textit{setosa} sample, this method yields the following contributions which sum up to $1$ because the random forest used predicts the setosa class with complete certainty. In this particular case, the result of the treeinterpreter highlights that both \textit{petal length} and \textit{petal width} had a positive impact on the decision, while \textit{sepal length} and \textit{sepal width} were almost not considered. The bias of a little more than one-third illustrates that the model is slightly biased towards the \textit{setosa class} compared to the other classes (see Table \ref{treeinterpreter}). 
\begin{table} 
\caption{Feature Importance for a certain instance}
\centering
    \begin{tabular}{| c | c |}
    \hline
    \textbf{Feature} & \textbf{Importance} \\ \hline
    bias & 0.35685 \\  \hline
    petal length (cm) & 0.28069664 \\ \hline
    petal width (cm)  & 0.26613121 \\ \hline
    sepal length (cm) & 0.08566027 \\ \hline
    sepal width (cm)  & 0.00684522 \\ \hline
    \end{tabular}
    \label{treeinterpreter}

\end{table}

\section{Data and Explainability}
\label{sec:Data}

This chapter focuses on the topic of data and explainability. First, we highlight the topic of data quality which is an essential factor for explainability. Furthermore, we describe the topic of ontologies in detail. Ontologies can improve the explainability of any given model by incorporating knowledge either before the model training or after the explanation generation to further improve them. 

\subsection{Data Quality}
\label{sec:DataQuality}

A survey conducted by Kaggle \citep{kaggle2017} revealed that the most significant barrier data scientists face is poor data.
The quality of the underlying data is essential. If a huge amount of incomplete and noisy data is used to train a model, the results will be poor. Data in the real world is always incomplete, noisy and inconsistent. For example, if you need to prevent data quality issues in a Hospital Information System (HIS) where the data is added manually by the hospital personnel, it is mandatory to have the uncleaned data to learn from mistakes the personnel made in order to prevent these mistakes from happening in the future. 
In fields in which persons are being classified to a certain class\eg credit scoring or medical treatment, it is mandatory to have an accurate model that hasn't been trained by inconsistent data. 
It is a commonly held belief that the more data is available, the better the model is. For example, Google's Research Director Peter Norvig \citep{norvig2017} claimed: ``We don't have better algorithms than anyone else; we just have more data." While this isn't wrong altogether, it obscures the crucial point---which is that it isn't enough to have more data but to have more good data \citep{amatriain}. Several experiments were conducted where more data was included in a training step and the results showed that this did not improve the performance of the model \citep{amatriain}. 

Therefore, \citet{gudivada2017data} quite rightly states: ``The consequences of bad data may range from significant to catastrophic''.
In 1996, \citet{wang1996beyond} introduced a typical definition for data quality. They defined data quality as the degree to which the available data meets the needs of the user. As \citet{helfert2016big} have derived from this definition, the concept of data quality is context-dependent and subjective.
There is a variety of data quality dimensions, the six core dimensions of which are Completeness, Uniqueness, Timeliness, Validity, Accuracy, and Consistency. Completeness is the proportion of data actually collected compared to the full amount of data that theoretically could have been collected. Uniqueness measures the degree of identical data instances. Timeliness is responsible for ensuring that the collected data is not outdated. Validity checks whether the syntax of the data matches its definition. Accuracy measures whether the available data describes the given task correctly. Consistency compares the same data representations with their definitions and measures the differences \citep{askham2013six}. 
In the field of machine learning, a high quality of data is essential for the prediction of a correct model result. Here, the assessment of data quality becomes more complex because different performance metrics, the search for the best parameters or the model type can distort the data quality \citep{gudivada2017data}. 
Most derived explanations are based on the outcome of a model and depend on the data that the model uses for its predictions. If the quality of the data is low, the prediction could already be wrong and thus, the explanation could be correct but based on inconsistent data. 

\subsection{Data Visualization}
\label{sec:Visualization}
\begin{figure*}
\centering
  \begin{minipage}{0.5\textwidth}
  \centering
     \includegraphics[scale=0.4]{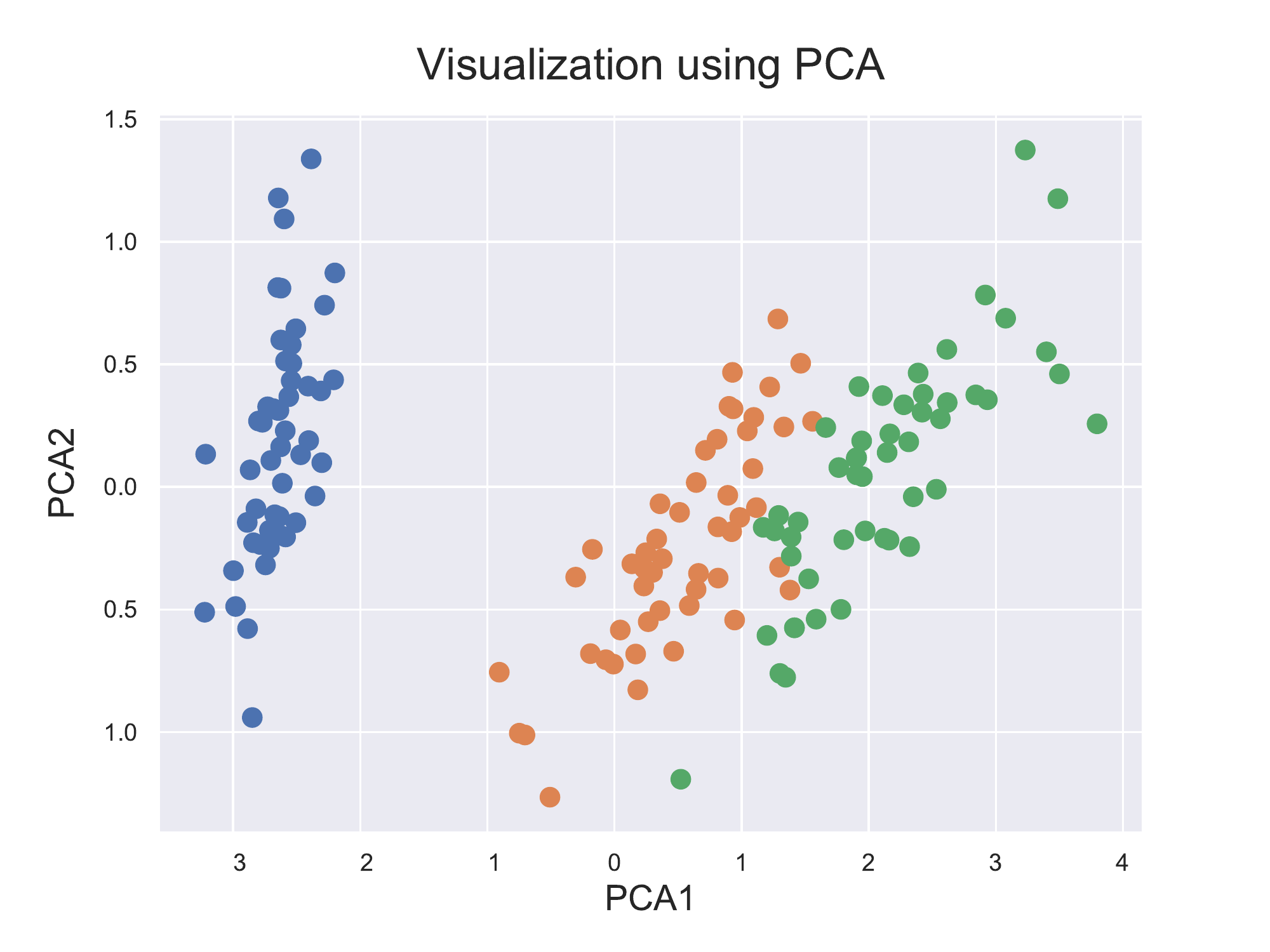}
    \subcaption{PCA (2D) }
    \label{fig:toya}
     \end{minipage}
        \vspace{4.00mm}
     \begin{minipage}{0.5\textwidth}
     \centering
    \includegraphics[scale=0.40]{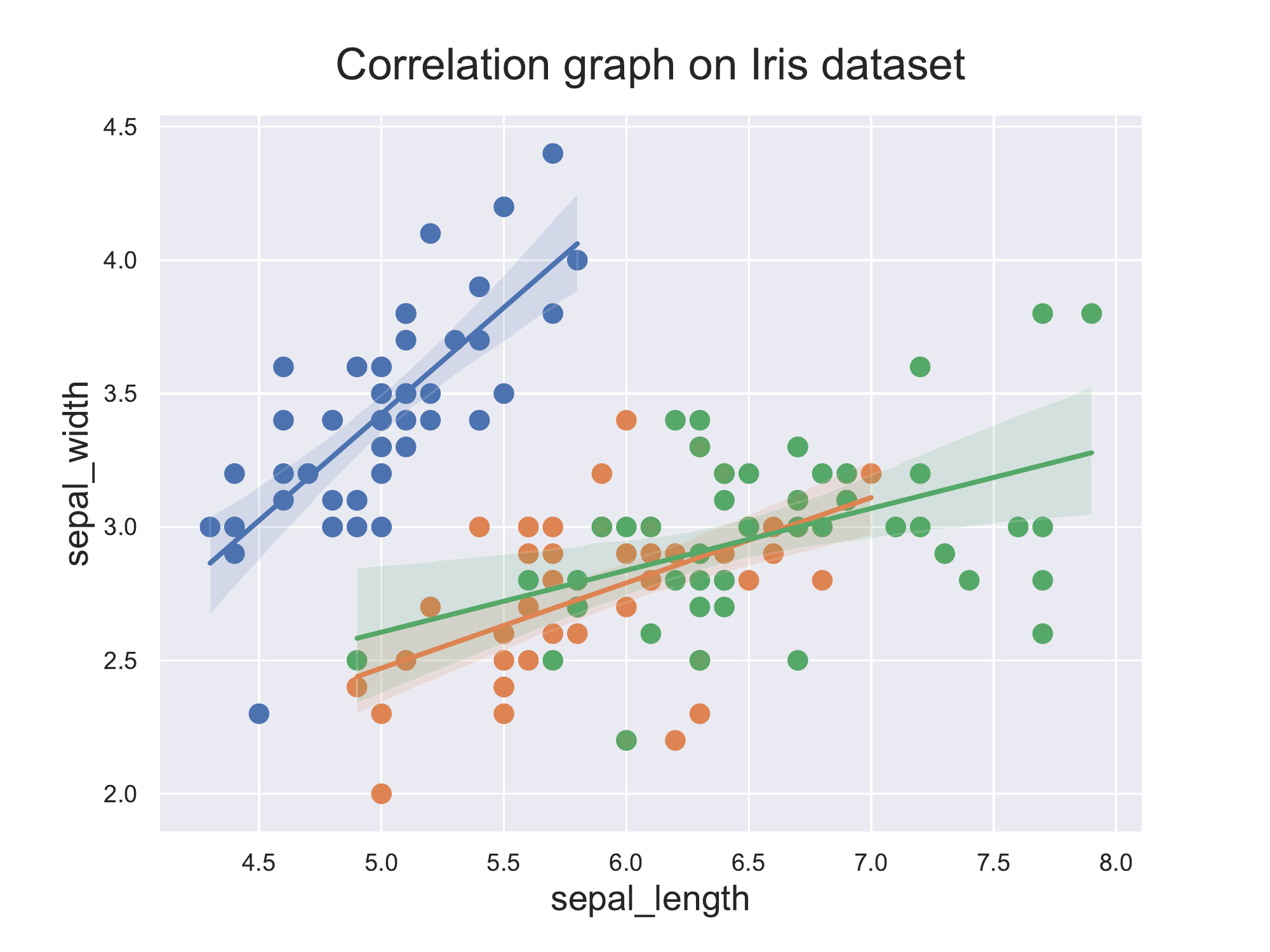}
    \subcaption{Correlation graph (2D)}
    \label{fig:toyc}
      \end{minipage}
\begin{minipage}{0.8\textwidth}
     \centering
    \includegraphics[scale=0.45]{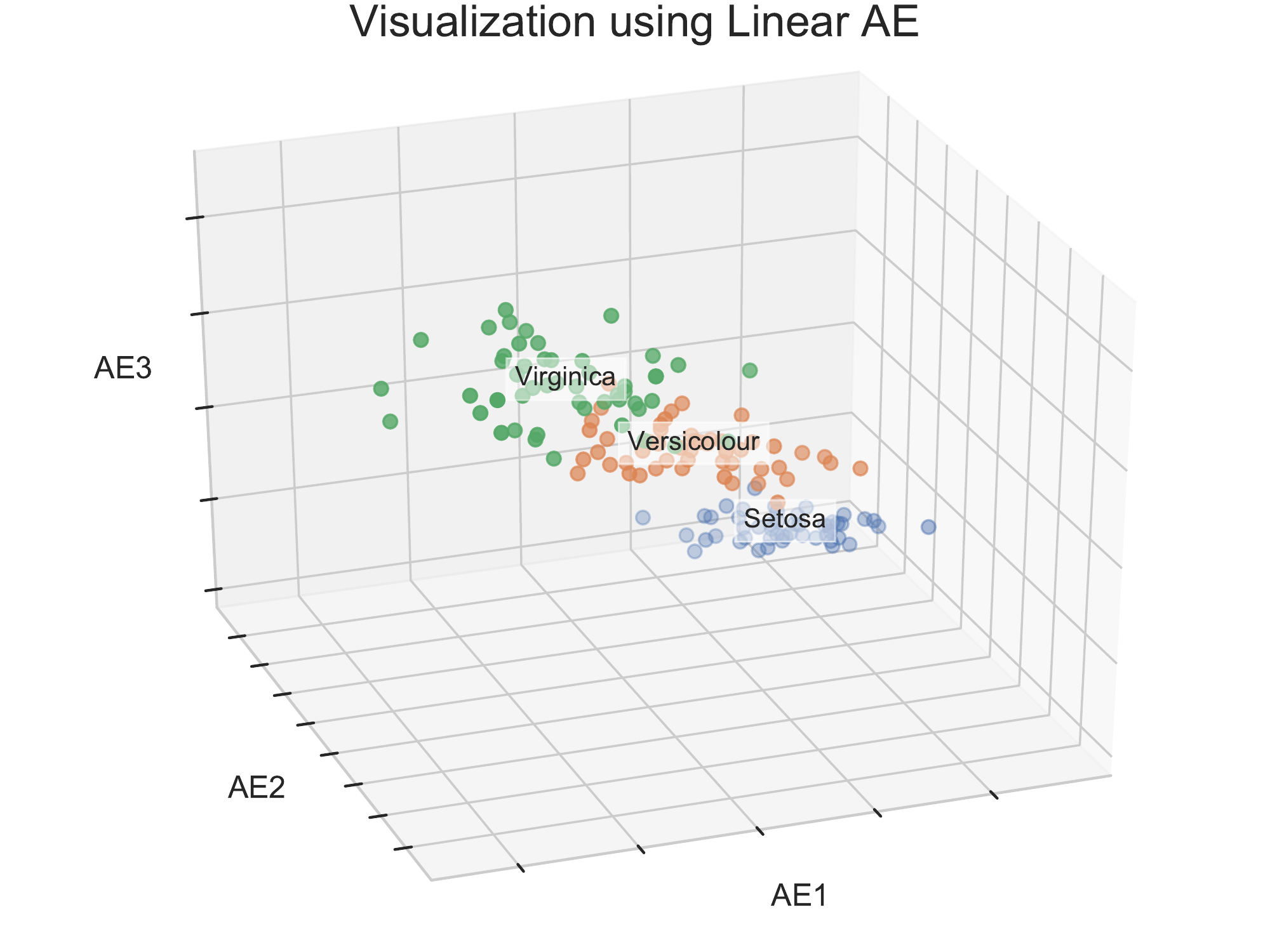}
    \subcaption{Linear AE (3D)}
    \label{fig:toyd}
      \end{minipage}
        
         \caption{Data Visualization techniques}\label{fig:datavis}
\end{figure*}

Data visualization approaches are useful in order to get a first impression of the data.
The goal is to visualize the entire data in just two or three dimensions. Visualization techniques can be used for exploratory data analysis and as a complement to a prediction model. Some of the available approaches are listed in Table \ref{tabl:visapproaches} and are not linked to a specific learning task. \Fig{fig:datavis} illustrates an example of some of the described visualization approaches. 

Nomograms can be used to visualize linear \acp{svm}, logistic regression, and \ac{nb}. They visualize the associated weights for each feature \citep{robnik2007explaining}. \ac{som} are two-layer \acp{ann} that preserve the topology of the underlying data. Similar instances are mapped closely together in a \ac{som}. Furthermore, color is assigned to the trained neurons based on their respective classification \citep{martens2008rule}.
Glyphs represent rows of a data set using color, texture, and alignment \citep{hall2017ideas}.
Correlation graphs visualize the relationships between input features \citep{hall2017ideas}.
Residual values are plotted against the predicted values. A random distribution of plotted points implies a good fit of the underlying prediction model \citep{hall2017ideas}.
An auto-encoder uses an \ac{ann} to learn a representation of the data using fewer dimensions than the original data. These dimension can then be visualized in a scatter plot \citep{hall2017ideas}.
\ac{pca} extracts the principle components from the data and visualizes these components using a scatter plot. The goal is to find linear combinations of input features that represent the underlying structure of the data while reducing the overall number of features \citep{hall2017ideas}. Scatter plots are a good tool for this visualization since they are able to visualize key structural elements such as clusters, hierarchy, outliers, and sparsity. They further project similar aspects close to one another.
\ac{mds} is a linear projection method that is used to map data on approximate Euclidean metric space and visualize it using a scatter plot \citep{hall2017ideas}.
The t-\acs{sne} visualization technique is a non-linear dimensionality reduction technique that maps data in a low dimensional space while preserving the data's structure. It finds two distributions in high and in low dimensional space and minimizes a metric between them. The high dimensional space can be converted into a matrix with pairwise similarities \citep{hall2017ideas}.

Other visualization techniques for visualizing feature importance and relationship were already mentioned. They include \ac{pdp}, \ac{ice} plots, explainVis using bar plots, \ac{vec} plots, \ac{vec} surface and contour plots, and heat maps. Figure \ref{tabl:visapproaches} illustrates some of the described visualization approaches.

\begin{table}
\centering
\caption{Overview of visualization approaches}
\label{tabl:visapproaches}
\begin{adjustbox}{width=\linewidth}
\renewcommand{\arraystretch}{1.25}
\begin{tabular}{|l||l|l|l|l|}
\hline
\textbf {Approach}   &
\textbf{Input Model} & \textbf{Model} & \textbf{References}  \\ \hline \hline

SOM & Model-agnostic & - & \citet{martens2008rule} \\ \hline

Nomogram &  Model-agnostic & Linear Model & \citet{robnik2007explaining}\\ \hline

Glyphs &   Model-agnostic & - & \citet{hall2017ideas} \\ \hline

Correlation Graphs &   Model-agnostic & - & \citet{hall2017ideas}\\ \hline

Autoencoder &   Model-agnostic&  - & \citet{kramer1991nonlinear}\\ \hline

PCA &  Model-agnostic & Linear/Non-linear Model & \citet{wold1987principal} \\ \hline

MDS & Model-agnostic & - & \citet{mead1992review} \\ \hline

t-SNE & Model-specific & Linear Model & \citet{maaten2008visualizing} \\ \hline

\end{tabular}
\end{adjustbox}
\end{table}

\subsection{Ontologies}
\label{sec:Ontolgies}
The application of ontologies can be used to improve the data quality but also to generate better explanations. By including domain knowledge that is approved by experts \eg consistency checks can be done directly on the data. This can improve the classification performance of the model. Other ontologies can be used to improve the explainability by including them before training the model \eg by summarizing features to facilitate the understanding for users. In what follows, we will give an overview of the topic of ontologies.

\subsubsection{Definition of Ontology}

Knowledge combines information, experience and skills to enable a concept of understanding. The basis for that process is data which is currently growing faster and faster. All this data is used to acquire different forms of information and to learn about important connections within the collected data. In the end, this knowledge can be used to make possible new advances and discoveries. But in order to fulfill this purpose, knowledge needs to be shared. 

Acquired knowledge is important in many fields and needs to be spread accordingly. An easy exchange needs a uniform structure and, thus, a common definition. This is even more important if knowledge is to be shared between computational systems like AIs. One approach to enable this exchange of data and knowledge are ontologies.
\citet{gruber1993translation} first defined an \emph{Ontology} as a specification of a conceptualization in the context of knowledge sharing. 
The definition includes essential parts that till this day are the main building blocks of ontologies. An ontology is a vocabulary that represents the categories, properties, and relations of a specific domain in a formal and organized structure. It consists of classes, properties, and individuals which are utilized to define concepts.

Definitions are written in a standardized language with the Web Ontology Language (OWL) \footnote{\url{https://www.w3.org/TR/owl2-overview/}} as a collection of languages used to author ontologies. It builds upon the Resource Description Framework (RDF) \footnote{\url{https://www.w3.org/RDF/}} that was initially conceptualized by the World Wide Web Consortium (W3C) for describing meta data and that is now a center piece of the Semantic Web. This concept defines statements in the form of 3-tuples, consisting of subjects, predicates, and objects which can be represented in a labeled, directed multi-graph.
The knowledge representation inside an ontology follows a description logic which utilizes concepts, roles, individuals, and operators. It is part of first order logic (predicate logic) and able to define formal descriptions of logical contexts. A concept represents a unary predicate, whereas a role represents a binary predicate and an individual is a constant. An operator can combine concepts and roles to form new definitions.

In summary, an ontology describes knowledge as formal definitions of the types, properties, and relations that exist in some domain. A uniform structure enables the exchange of knowledge between various fields of science.
Ontologies can be combined to create a greater knowledge out of smaller concepts, where upper ontologies build the basis for bigger compounds \citep{hoehndorf2010upper}. The Descriptive Logic enables the use of Semantic Reasoners which infer unstated information from the ontology and check the consistency of that knowledge base.

Representing knowledge in this form of ontologies can have many advantages such as the inference of information from relational concepts. But since different needs require different services, there are other approaches to define relational knowledge.

\emph{Taxonomies} are mostly used to demonstrate origins and connections of words. They show relations of usage and their counterparts which makes them very useful for structuring information. Therefore, many dictionaries are represented in the form of a taxonomy. Even ontologies often use taxonomies as a building block to make their knowledge more structured.

Furthermore, ontologies themselves can be used as building blocks to construct even greater knowledge relations. Those concepts known as \emph{knowledge graphs} are high level structures of connected information. Using them to interlink sources of similar topics makes them a great framework for search engines. The largest knowledge graph today is produced by Google and displays additional and related information in a separate area on the side when searching with Google's search engine. It enables quick access to corresponding knowledge by linking similar information and commonly searched additions. 

\subsubsection{Ontologies in Practice}
Besides sharing knowledge in a uniform way across many areas, ontologies are primarily used for information mining. The focus does not lie on the retrieval of raw data but on the cognition of possible connections within a domain.
In the pharmacy industry, searching for the causes of an illness can be done by categorizing identified explicit relationships within a causality relation ontology \citep{Osama2012}. This enables a quicker connection of symptoms and possible reasons, thus making treatments easier and better.
Another prominent example of data mining capabilities of ontologies is IBM's project \emph{Watson} \citep{watson}.
Watson is one of the so-called DeepQA projects. It is a program designed to answer questions in natural language. The power of this system is created by the immense amount of data it can process and by its ability to connect relations via ontologies and similar concepts.
Other use cases of ontologies enrich Semantic Web mining, mining health records for insights, fraud detection, and semantic publishing.

Since ontologies are designed to share knowledge, there are some popular ontologies\footnote{\url{https://www.w3.org/wiki/Good_Ontologies}} that are primarily used in their genuine field of application. An example for this is the \textit{Friend Of A Friend (FOAF)} ontology \citep{FOAF}that is commonly used to describe people and social relationships on the Web. It can be complemented by the \textit{Semantically-Interlinked Online Communities (SIOC)} ontology \citep{SIOC} which extends the FOAF ontology with online communities and the definitions of their content.
The \textit{Music Ontology} \citep{Music, Raimond2007} is the most used knowledge base to describe information related to the music industry. 
Search engines like Yahoo, SearchMonkey and BestBuy utilize the \textit{GoodRelations} \citep{GoodR} vocabulary to describe products sold online and to improve search results when shopping online.

\begin{table}
  \centering
    \caption{Often used ontologies and their use cases}
  \begin{adjustbox}{width=0.6\textwidth,center=\textwidth} 
  \begin{tabular}{|l||l|l|} 
  \hline
    \textbf{Approach}                        & \textbf{Use Case} & \textbf{Reference}\\ \hline 
    FOAF                         & Social media & \citet{FOAF}  \\ \hline
    SIOC                          & Social media  & \citet{SIOC}\\ \hline
    Music Ontology                & Music industry & \citet{Music} \\ \hline
    GoodRelations              & Online shopping& \citet{GoodR}  \\ \hline
  \end{tabular}
  \end{adjustbox}
  \label{tab:ontologies}
\end{table}

In the next sections, we will discuss research that has been conducted in the field of ontologies and their usage within machine learning and explainability approaches. Because these fields are relatively new, most work is directed at basic concepts that are needed for future applications of those systems. Nevertheless, there are some promising steps towards the implementation of domain knowledge in AI and their explanations.

\subsubsection{Improving ontologies}
Due to their already growing use, research focuses greatly on improving ontologies. \citet{Wong2011} provide a method for the automation of ontology generation from domain data. \textit{OntoLearn} \citep{Navigli2004} uses websites and shared documents to extract domain terminology to then arrange it hierarchically and finally create an ontology.
Other researchers go even further and attempt to completely automate the development of ontologies. \citet{Maedche2001} propose semi-automatic import, extraction, pruning, refinement, and evaluation of ontologies, providing the ontology engineer with coordinated tools for ontology modeling.

\begin{table}
  \centering
   \caption{Overview of research that improves ontologies}
  \begin{adjustbox}{width=0.7\textwidth,center=\textwidth} 
\begin{tabular}{|l|l|l|}        
\hline
  \textbf{Approach}                            & \textbf{Description}  & \textbf{References}\\ \hline 
    -                             & Automated generation& \citet{Wong2011}\\ \hline 
    OntoLearn                  & Automated extraction, generation & \citet{Navigli2004}\\ \hline 
    -                          & Semi-Automated engineering & \citet{Maedche2001}\\ \hline 
  \end{tabular}
  \end{adjustbox}
 
  \label{tab:onto-improv}
\end{table}

\subsubsection{Ontologies and Machine Learning}
In many use cases, more than one ontology is needed and therefore multiple ontologies need to be connected. ML strategies could improve the mapping of those multiple ontologies and thereby enable the construction of large knowledge bases. \citet{Doan2004} developed \textit{GLUE}, a system that employs learning techniques to semi-automatically create semantic mappings between ontologies.
Conversely, ontologies are used to refine ML by incorporating them into the training process. An application in bio-medicine \citep{Tysmbal2007} integrates domain specific ontologies to improve decision-making. The existence of large knowledge bases is used to mitigate the complexity of such a high dimensional field of science and to use combined knowledge for an overall better understanding.
A different concept is proposed by \citet{Ning2015}. This concept uses semantic relations among features that are defined inside an ontology to improve random forests for image classification. Splits in each decision tree are determined by semantic relations which automatically include the hierarchical structure of knowledge. This usage of ontologies proves to be very helpful for image recognition, as different detectable things are made up of parts that are all included in the ontology. 

\begin{table*}
\centering
\caption{Overview of ontology approaches that improve ML}
\label{overview_induction}
\begin{adjustbox}{width=0.7\textwidth,center=\textwidth} 
\begin{tabular}{|l|l|l|}
\hline
   \textbf{Approach}                & \textbf{Description}  & \textbf{References}             \\  \hline
    GLUE                  & Improve ontologies with ML  & \citet{Doan2004}      \\ \hline 
    -                    & Improve ML with ontologies   & \citet{Tysmbal2007}   \\ \hline 
    -                    & Improve ML with ontologies   & \citet{Ning2015}      \\ \hline 
  
 \end{tabular}
\end{adjustbox}
\end{table*}

\subsubsection{Ontologies and Explainability}
\label{sec:OntoExplainability}
The goal of explanations for AI systems is to make complex models understandable for human beings. This includes the connection of related facts and their informational importance on the result. Ontologies as concepts that represent knowledge about the different relations of data provide a huge potential in making complex data structures better understandable.

A first step towards a better understanding of these systems is the \textit{ML-Schema} developed by \citet{Publio2018}. This top-level ontology provides a set of classes, properties, and restrictions that can be used to define information on ML algorithms, data sets, and experiments. A uniform description of these elements simplifies the exchange of such collected knowledge.
The next step covers the concept of explanations themselves. \citet{McGuinness2007} showcase \textit{PML 2}, a combination of three ontologies that is designed to serve as an interlingua for the sharing of generated explanations. It consists of the \textit{provenance ontology}, the \textit{justification ontology} and the \textit{trust relation} ontology (also known as PML-P, PML-J and PML-T) which are all used to describe information within system responses that can then be used to generate an explanation. This approach focuses on the information about how a system generates output, on its dependencies and on the possible information that could transport trust. \textit{PML 2} creates a uniform definition to share the building blocks of an explanation.

The structure of an ontology does not only provide a good possibility to share knowledge but it can also be utilized directly. \citet{Roberto2019} present \textit{Trepan reloaded}, a recent approach that uses domain ontologies while also generating an explanation. A decision tree is used to explain the decision process of a neural network. The goal is to benefit from the structured knowledge within an ontology and to apply it to the next generation of the decision tree. A conducted user study illustrates that explanations which follow a structure similar to the human understanding of data are better understandable.
Explanations for neural networks try to make the process from the input to the output understandable. \citet{Sarker2017} analyze this relation with the help of ontologies. Their proof of concept describes the potential of background knowledge for the explainability of such systems.
\citet{Panigutti2020} extend the relational concepts over time and present \textit{Doctor XAI}. Showcased for medical situations, this approach deals with multi-labeled, sequential, ontology-linked data. This means it is able to deduce connections between information that happen over time. It creates a new source of knowledge that can be used to improve the predictions and explanations made in such critical areas.
Supported by the concept of knowledge sharing, transfer learning \citep{west2007} is a sub category of ML and aims at reusing learned information and applying it to new learning problems. \citet{Chen2018} use ontologies to implement explanations for transfer learning. The goal is to create human-centered explanations that enable non-ML experts to detect positive and negative transfers. This way, they can decide what to transfer and when to transfer in order to create an optimized transfer learning setting.
\citet{Geng2019} present an alternative approach for this that utilizes Knowledge Graphs. Both emphasize the potential of knowledge bases in human-centered explanations. The approach developed by \citet{mahajan2019preserving} presents a method that uses (partial) structural causal models to generate actionable counterfactuals. 

\begin{table*}
\centering
\caption{Overview of ontology approaches in XAI}
\label{OntologiesXAI}
\begin{adjustbox}{width=0.7\textwidth,center=\textwidth} 
\begin{tabular}{|l|l|l|}
\hline
    \textbf{Approach}                                 & \textbf{Description} & \textbf{References}\\ \hline
    ML-Schema               & ML described by an ontology  & \citet{Publio2018}   \\ \hline 
    PML 2                   & Explanations described by an ontology & \citet{McGuinness2007}\\ \hline 
    Trepan reloaded           & Explanations supported by an ontology & \citet{Roberto2019} \\ \hline 
    -                         & Explanations supported by an ontology  & \citet{Sarker2017} \\ \hline 
    Doctor XAI             & Explanations supported by an ontology  & \citet{Panigutti2020} \\ \hline 
    -                         & Explanations supported by an ontology & \citet{Chen2018}    \\ \hline
       Feasible counterfactuals                        & Counterfactual Explanations supported by causal constraints & \citet{Chen2018}    \\ \hline
    Thales XAI Platform                       & Explanations supported by an ontology & \citet{Thales2019}    \\ \hline 
 \end{tabular}
\end{adjustbox}
\end{table*}

Other works try to incorporate Knowledge Graphs as well, with the \textit{Thales XAI Platform} \citep{Thales2019} being the first of its kind. One approach for making AI systems applicable to critical situations, this platform provides example-based and feature-based explanations or counterfactuals, using textual and visual representations. In addition, explanations based on semantics are generated with the help of Knowledge Graphs. Semantic-Web tools are used to enrich the data for ML with context information. Explanations are then generated by using a Knowledge Graph and the context information of the ML result to identify representative semantic relations.

\section{Discussion and Open Challenges}
\label{sec:discussion}
There is a lively debate about the need for explainability in machine learning in general and, more specifically, about what the main research focus should lie on. Proponents of explainable artificial intelligence argue that for a model to be reliable, first and foremost it needs to be understandable. Opponents take the view that sometimes, human reasoning is a complete black box as well, so there is no actual need for explainability for every purpose. Maybe someone wants to understand the reasoning of a doctor for preferring a particular medical treatment but cares less about why his loan was approved. Both sides seem to agree that in regulated areas, the need for understandable models are crucial. 
To get a sense of the global model behaviour, the methods of\eg a simple decision tree or \ac{brl} can be applied. While the tree and the rule lists do not vary in the nature of their respective approaches (since a tree can be transformed into a decision list as well), decision lists are more easily understandable and directly convey their supposed meaning to a "lay user". Especially if the model has learned only a few significant features, decision lists approximate their relationships well. If more features are required to make a meaningful distinction, the decision tree has the upper hand while still being small and readable with a reduced amount of training. Both decision trees and rule lists have the characteristic to be reproducible and, thus, being what Lipton \citep{lipton2016mythos} calls \textit{human-simulatable}.
SP-Lime does not allow a "lay user" to make that reproducible judgment because the presentation of various single explanations does not necessarily imply how a new instance will be classified. If the submodular-pick is selected wisely and if the variation between instances is reasonably clear, the user will probably be able to make a good guess, but that also requires the understanding of how the sample explanations were generated. 
Explanation generation methods like PDPs seem to have little value to "lay users". However, for those users who do possess knowledge about the relevant techniques, these methods can yield a very rough approximation of the model judgment based on the features. 
With regards to local surrogate explanations, we applied SHAP and LIME. As \citet{lundberg2017unified} state, both of them follow a similar approach of explanation. However, apart from the actual visual representation, they do not make a real difference for the user. The only difference is the approach by which the results are being produced and that yields marginal different values for single features. 
Since the user probably is interested in both a general direction and the most significant contributions, the underlying method is irrelevant.




When it comes to explainability in particular, dimension reduction methods are mentioned in the literature \citep{kittler}. However, dimension reduction approaches need to be used carefully when the goal is to achieve more explainability because they could conceal explainability. The pre-processing of the data is important because it can already lead to a more understandable model. Whether the goal is to achieve a more interpretable or a more accurate model, it is never a bad idea to mind and mine the quality of the models' input data.
After illustrating different explainability approaches, especially in the local area, it becomes clear that what was shown by \citet{lundberg2017unified} has an influence on the intrinsic quality of explanations as well. Many of the explanations resort to feature importance which could make intuitive sense to a statistician but is likely not the way a "lay user" would like the process to be explained. 
Besides this, there are several more challenges that need to be tackled in the future:

\textit{Interpretable versus black box models - Rethinking the problem in the first place:}
Before we train models that solve the addressed problem, we need to reflect on the question, "What are we actually looking for? Do we really need a black box model?" \citet{stoprudin} describes the explosion of research in the field of explainable ML with surrogate models regarding high stakes decisions as problematic. They state that building interpretable models that are accurate as black boxes should be considered. Therefore, more research in the field where a black box and an interpretable model are competing against each other is needed. If we had interpretable models that prove to be reliable as black boxes, would there still be a need to use them?

\textit{Measuring and comparing explainability:}
According to our research, what is missing is a standard procedure to measure, quantify, and compare the explainability of enhancing approaches that allows scientists to compare these different approaches. Although some research is being done in this  field \ref{sec:assessmentExplainability}, a standardized procedure is needed. The performance of a classifier is evaluated by\eg accuracy, recall, and the F1-score. The need for likewise metrics to evaluate explainability is crucial. \citet{miller2017explanation} states that most of the work relies on the authors' intuition about explainable machine learning. An essential point within the research of XSML is to have metrics that describe the overall explainability and to be able to compare different models regarding their level of explainability. 

\textit{Improving explanations with ontologies:}
Another research area that should be further addressed is the combination of ontologies with explanations. We already addressed this point in section \ref{sec:OntoExplainability} but further research with practical use cases needs to be done. Furthermore, the advantage of the combination with ontologies must be examined by several user studies. 

\textit{Trust in machine learning models:}
What if we measure both the explainability and the trust within a model but both are missing? 
Can we provide more trust just-in-time? 
What are the possibilities to raise the trust in the model? All those questions remain hard to answer without more research and especially user-centered experiments in this field.
\citet{schmidt2019quantifying} introduced a quantitative measure of trust in ML decisions and conducted an experiment. In their experiment, they examined two methods, COVAR, a glass-box method, and LIME, a black box method. They found that COVAR yielded more interpretable explanations. Thereby, they highlighted the usefulness of simple methods. \citet{lipton2017doctor} states that addressing the foundations of the problem by discovering what explainability really is will be crucial to see meaningful progress within this field.

\textit{User studies regarding specific explainability aspects:}
Almost every day, a paper is published that purports to solve the explainability problem algorithmically. However, another important aspect are user studies. There are only a few user experiments in the area of explainability, but much more experiments are needed to cover the topic holistically. \citet{poursabzi2018manipulating} measure trust by determining the difference between the prediction of the model and the participant's prediction. As a use case, they predict housing prices. 
\citet{ribeiro2016should} conducted a user study to measure whether the participants would trust the prediction model. \citet{yin2019understanding} measure the frequency with which they revise their predictions to match the predictions of the model and their self-reported levels of trust in the model. \citet{el2019study} evaluate three different explanation approaches based on the users' trust by a within-subject design study. \citet{lage2019evaluation} conducted a user study to find out what makes explanations human-interpretable by systematically varying properties of explanation to measure the effect of these variations on the performance of several tasks. The tasks were: simulating the system's response, verifying a suggested response, and counterfactual reasoning. One of their findings included that, across all experiments, counterfactual questions showed significantly lower accuracy. 
\citet{herman2017promise} differentiates between a descriptive and persuasive explanation generation task. Whereas the first describes an explanation within a feature space generated by the explainable or interpretable approach, the latter adds cognitive functions, user preferences, and expertise to the explanation.


\section{Conclusion}

The relevance of explainable machine learning can be seen in many areas.  
The high number of published research papers in certain areas can probably be attributed to the fact that there is a fairly high need to provide explanations to users in these areas \eg in the medical domain. For example, explainable ML was already used to learn more about COVID-19 \citep{chen2020interpretable, fan2020investigation, rezaul2020deepcovidexplainer, bao2020triaging}. 

This paper introduced different problem definitions of explainable \ac{sml} and categorized and reviewed past and recent approaches in each field. The use cases mentioned throughput the paper were supposed to illustratively depict each approach according to the problem definitions given. The overall goal of the paper was to gather an explanation that offers an broad overview of different approaches in the field of explainable \ac{sml}. For example, by providing an overview over the most influential features to the decision by considering a local linear approximation of the model, an explanation can be generated. For an intuitive example, we can think of a linear approximation to a complex model as a sort of representative-analysis where we can illustrate for a user how representative the features for the predicted class based on the most important features actually were. This approach can give a reasonable human interpretation that matches with most subconscious processes of decision-making which often rely on the representativeness of a certain instance.
However, when humans try to explain themselves, they use different approaches to knowledge that can hardly be captured in a classifier. This is due to the fact that humans often explains themselves by referring to post-factum coherent stories. Rather than providing two features and an importance of those features with a specified class, the human mind would tend to build a story around those features that explains why it seems obvious that the respective instance belongs to a specific class. In an explanation like this, all sorts of environmental conditions play a role, as the person telling the story seeks to build trust and understanding for her decision. 
Thus, the most we can strive for when explaining a model is a sort of human graspable approximation of the decision process. 

 Diving down into the ethical dilemmas of automated decision-making, especially self-driving cars move into the focus. If a self-driving car needs to decide whether it drives into a crowd of elderly or a crowd of young people, the ethical controversy begins. The moral machine \citep{mit} is an attempt to assess ethical dilemmas of the kind in which self-driving cars need to choose between insoluble situations. Would you be able to choose? If so, could you explain your decision?




\newpage









\vskip 0.2in
\begin{thebibliography}{251}
\providecommand{\natexlab}[1]{#1}
\providecommand{\url}[1]{\texttt{#1}}
\expandafter\ifx\csname urlstyle\endcsname\relax
  \providecommand{\doi}[1]{doi: #1}\else
  \providecommand{\doi}{doi: \begingroup \urlstyle{rm}\Url}\fi

\bibitem[Abdollahi and Nasraoui(2016)]{abdollahi2016explainable}
B.~Abdollahi and O.~Nasraoui.
\newblock Explainable restricted boltzmann machines for collaborative
  filtering.
\newblock \emph{arXiv preprint arXiv:1606.07129}, 2016.

\bibitem[Abdollahi and Nasraoui(2017)]{abdollahi2017using}
B.~Abdollahi and O.~Nasraoui.
\newblock Using explainability for constrained matrix factorization.
\newblock In \emph{Proceedings of the Eleventh ACM Conference on Recommender
  Systems}, pages 79--83, 2017.

\bibitem[ACM(2017)]{acm}
ACM.
\newblock Statement on algorithmic transparency and accountability, June 2017.
\newblock URL
  \url{https://www.acm.org/binaries/content/assets/publicpolicy/2017_usacm_statement_algorithms.pdf}.

\bibitem[Adadi and Berrada(2018)]{adadi2018}
A.~Adadi and M.~Berrada.
\newblock Peeking inside the black-box: A survey on explainable artificial
  intelligence (xai).
\newblock \emph{IEEE Access}, 2018.

\bibitem[Adler et~al.(2016)Adler, Falk, Friedler, Rybeck, Scheidegger, Smith,
  and Venkatasubramanian]{adler2016auditing}
P.~Adler, C.~Falk, S.~Friedler, G.~Rybeck, C.~Scheidegger, B.~Smith, and
  S.~Venkatasubramanian.
\newblock Auditing black-box models for indirect influence.
\newblock In \emph{Data Mining (ICDM), 2016 IEEE 16th International Conference
  on}. IEEE, 2016.

\bibitem[Amatriain(2017)]{amatriain}
X.~Amatriain.
\newblock More data or better models?, November 2017.
\newblock URL
  \url{http://technocalifornia.blogspot.de/2012/07/more-data-or-better-models.html}.

\bibitem[Andrews et~al.(1995)Andrews, Diederich, and Tickle]{andrews1995survey}
R.~Andrews, J.~Diederich, and A.~B. Tickle.
\newblock Survey and critique of techniques for extracting rules from trained
  artificial neural networks.
\newblock \emph{Knowledge-based systems}, 1995.

\bibitem[Andrzejak et~al.(2013)Andrzejak, Langner, and
  Zabala]{andrzejak2013interpretable}
A.~Andrzejak, F.~Langner, and S.~Zabala.
\newblock Interpretable models from distributed data via merging of decision
  trees.
\newblock In \emph{Computational Intelligence and Data Mining (CIDM), 2013 IEEE
  Symposium on}. IEEE, 2013.

\bibitem[Angelov and Soares(2019)]{angelov2019towards}
P.~Angelov and E.~Soares.
\newblock Towards explainable deep neural networks (xdnn).
\newblock \emph{arXiv preprint arXiv:1912.02523}, 2019.

\bibitem[Askham et~al.(2013)Askham, Cook, Doyle, Fereday, Gibson, Landbeck,
  Lee, Maynard, Palmerand, and Schwarzenbach]{askham2013six}
N.~Askham, D.~Cook, M.~Doyle, H.~Fereday, M.~Gibson, U.~Landbeck, R.~Lee,
  C.~Maynard, G.~Palmerand, and J.~Schwarzenbach.
\newblock The six primary dimensions for data quality assessment.
\newblock \emph{DAMA UK Working Group}, pages 432--435, 2013.

\bibitem[Augasta and Kathirvalavakumar(2012)]{augasta2012reverse}
M.~G. Augasta and T.~Kathirvalavakumar.
\newblock Reverse engineering the neural networks for rule extraction in
  classification problems.
\newblock \emph{Neural processing letters}, 2012.

\bibitem[Baehrens et~al.(2010)Baehrens, Schroeter, Harmeling, Kawanabe, Hansen,
  and M.~Zoeller]{baehrens2010explain}
D.~Baehrens, T.~Schroeter, S.~Harmeling, M.~Kawanabe, K.~Hansen, and
  Klaus-Robert M.~Zoeller.
\newblock How to explain individual classification decisions.
\newblock \emph{Journal of Machine Learning Research}, 2010.

\bibitem[Bahdanau et~al.(2014)Bahdanau, Cho, and y.~Bengio]{bahdanau2014neural}
D.~Bahdanau, K.~Cho, and y.~Bengio.
\newblock Neural machine translation by jointly learning to align and
  translate.
\newblock \emph{arXiv preprint arXiv:1409.0473}, 2014.

\bibitem[Balestriero(2017)]{balestriero2017neural}
R.~Balestriero.
\newblock Neural decision trees.
\newblock \emph{arXiv preprint arXiv:1702.07360}, 2017.

\bibitem[Barakat and Diederich(2004)]{barakat2004learning}
N.~Barakat and J.~Diederich.
\newblock Learning-based rule-extraction from support vector machines.
\newblock In \emph{The 14th International Conference on Computer Theory and
  applications ICCTA'2004}. not found, 2004.

\bibitem[Barakat and Bradley(2007)]{barakat2007rule}
N.~H. Barakat and A.~P. Bradley.
\newblock Rule extraction from support vector machines: A sequential covering
  approach.
\newblock \emph{IEEE Transactions on Knowledge and Data Engineering}, 2007.

\bibitem[Barbella et~al.(2009)Barbella, Benzaid, Christensen, Jackson, Qin, and
  Musicant]{barbella2009understanding}
D.~Barbella, S.~Benzaid, J.~M. Christensen, B.~Jackson, X.~V. Qin, and
  D.~Musicant.
\newblock Understanding support vector machine classifications via a
  recommender system-like approach.
\newblock In \emph{DMIN}, 2009.

\bibitem[Bastani et~al.(2017)Bastani, Kim, and Bastani]{bastani}
O.~Bastani, C.~Kim, and H.~Bastani.
\newblock Interpreting blackbox models via model extraction.
\newblock unpublished, 2017.

\bibitem[Bekri et~al.(2019)Bekri, Kling, and Huber]{el2019study}
N.~El Bekri, J.~Kling, and M.~F. Huber.
\newblock A study on trust in black box models and post-hoc explanations.
\newblock In \emph{International Workshop on Soft Computing Models in
  Industrial and Environmental Applications}. Springer, 2019.

\bibitem[Bengio and Pearson(2016)]{bengio2017}
Y.~Bengio and J.~Pearson.
\newblock When ai goes wrong we won't be able to ask it why, July 2016.
\newblock URL
  \url{https://motherboard.vice.com/en_us/article/vv7yd4/ai-deep-learning-ethics-right-to-explanation}.

\bibitem[Berkson(1953)]{berkson1953statistically}
J.~Berkson.
\newblock A statistically precise and relatively simple method of estimating
  the bio-assay with quantal response, based on the logistic function.
\newblock \emph{Journal of the American Statistical Association}, 1953.

\bibitem[Bertsimas et~al.(2011)Bertsimas, Chang, and
  Rudin]{bertsimas2011ordered}
D.~Bertsimas, A.~Chang, and C.~Rudin.
\newblock Ordered rules for classification: A discrete optimization approach to
  associative classification.
\newblock In \emph{SUBMITTED TO THE ANNALS OF STATISTICS}. Citeseer, 2011.

\bibitem[Bhatt et~al.(2019)Bhatt, Ravikumar, and J.~M.
  F.~Moura]{bhatt2019towards}
U.~Bhatt, P.~Ravikumar, and Jose J.~M. F.~Moura.
\newblock Towards aggregating weighted feature attributions.
\newblock \emph{arXiv preprint arXiv:1901.10040}, 2019.

\bibitem[Bien and Tibshirani(2009)]{bien2009classification}
J.~Bien and R.~Tibshirani.
\newblock Classification by set cover: The prototype vector machine.
\newblock \emph{arXiv preprint arXiv:0908.2284}, 2009.

\bibitem[Bien and Tibshirani(2011)]{bien2011prototype}
J.~Bien and R.~Tibshirani.
\newblock Prototype selection for interpretable classification.
\newblock \emph{The Annals of Applied Statistics}, 2011.

\bibitem[Biran and Cotton(2017)]{biran2017explanation}
O.~Biran and C.~Cotton.
\newblock {Explanation and Justification in Machine Learning: A survey}.
\newblock In \emph{IJCAI-17 Workshop on Explainable AI (XAI)}, 2017.

\bibitem[Biran and McKeown(2017)]{biran2017human}
O.~Biran and K.~R. McKeown.
\newblock Human-centric justification of machine learning predictions.
\newblock In \emph{IJCAI}, 2017.

\bibitem[Biswas et~al.(2017)Biswas, Chakraborty, Purkayastha, Roy, and
  Thounaojam]{biswas2017rule}
S.~K. Biswas, M.~Chakraborty, B.~Purkayastha, P.~Roy, and D.~M. Thounaojam.
\newblock Rule extraction from training data using neural network.
\newblock \emph{International Journal on Artificial Intelligence Tools}, 2017.

\bibitem[Bohanec et~al.(2017)Bohanec, Bor{\v{s}}tnar, and
  Robnik-{\v{S}}ikonja]{bohanec2017explaining}
M.~Bohanec, M.~K. Bor{\v{s}}tnar, and M.~Robnik-{\v{S}}ikonja.
\newblock Explaining machine learning models in sales predictions.
\newblock \emph{Expert Systems with Applications}, 2017.

\bibitem[Boz(2002)]{boz2002extracting}
O.~Boz.
\newblock Extracting decision trees from trained neural networks.
\newblock In \emph{Proceedings of the eighth ACM SIGKDD international
  conference on Knowledge discovery and data mining}. ACM, 2002.

\bibitem[Breiman(2017)]{breiman2017classification}
L.~Breiman.
\newblock \emph{Classification and regression trees}.
\newblock Routledge, 2017.

\bibitem[Burkart et~al.(2019)Burkart, Huber, and Faller]{burkart2019forcing}
N.~Burkart, M.~F. Huber, and P.~Faller.
\newblock Forcing interpretability for deep neural networks through rule-based
  regularization.
\newblock In \emph{2019 18th IEEE International Conference On Machine Learning
  And Applications (ICMLA)}, pages 700--705. IEEE, 2019.

\bibitem[Byrum(2017)]{agriculture2017}
J.~Byrum.
\newblock The challenges for artificial intelligence in agriculture, February
  2017.
\newblock URL
  \url{https://agfundernews.com/the-challenges-for-artificial-intelligence-in-agriculture.html}.

\bibitem[Cambridge(2020)]{cambridge}
Cambridge.
\newblock \emph{The Cambridge dictionary of psychology.}
\newblock Cambridge University Press, 2020.

\bibitem[Caruana et~al.(2015)Caruana, Lou, Gehrke, Koch, Sturm, and
  Elhadad]{caruana2015interpretable}
R.~Caruana, Y.~Lou, J.~Gehrke, P.~Koch, M.~Sturm, and N.~Elhadad.
\newblock Intelligible models for healthcare: Predicting pneumonia risk and
  hospital 30-day readmission.
\newblock In \emph{Proceedings of the 21th ACM SIGKDD International Conference
  on Knowledge Discovery and Data Mining}. ACM, 2015.

\bibitem[Charniak(1991)]{charniak1991bayesian}
E.~Charniak.
\newblock Bayesian networks without tears.
\newblock \emph{AI magazine}, 1991.

\bibitem[Chen et~al.(2015)Chen, Fraiberger, Moakler, and
  Provost]{chen2015enhancing}
D.~Chen, S.~P. Fraiberger, R.~Moakler, and F.~Provost.
\newblock Enhancing transparency and control when drawing data-driven
  inferences about individuals.
\newblock \emph{Proceedings of 2016 ICML Workshop on Human Interpretability in
  Machine Learning}, 2015.

\bibitem[Chen et~al.(2018)Chen, L{\'{e}}cu{\'{e}}, Pan, Horrocks, and
  Chen]{Chen2018}
J.~Chen, F.~L{\'{e}}cu{\'{e}}, J.~Z. Pan, I.~Horrocks, and H.~Chen.
\newblock Knowledge-based transfer learning explanation.
\newblock \emph{CoRR}, abs/1807.08372, 2018.
\newblock URL \url{http://arxiv.org/abs/1807.08372}.

\bibitem[Chen et~al.(2020)Chen, Ouyang, Bao, Li, Han, Zhang, Zhu, Xu, Liu, Ge,
  et~al.]{chen2020interpretable}
Y.~Chen, L.~Ouyang, S.~Bao, Q.~Li, L.~Han, H.~Zhang, B.~Zhu, M.~Xu, J.~Liu,
  Y.~Ge, et~al.
\newblock An interpretable machine learning framework for accurate severe vs
  non-severe covid-19 clinical type classification.
\newblock \emph{medRxiv}, 2020.

\bibitem[Clark and Niblett(1989)]{clark1989cn2}
P.~Clark and T.~Niblett.
\newblock The cn2 induction algorithm.
\newblock \emph{Machine learning}, 1989.

\bibitem[Cleland(2011)]{norvig2017}
S.~Cleland.
\newblock Google's 'infringenovation' secrets, 2011.
\newblock URL
  \url{https://www.forbes.com/sites/scottcleland/2011/10/03/googles-infringenovation-secrets/#5c0c6fd230a}.

\bibitem[Cohen(1995)]{cohen1995fast}
W.~Cohen.
\newblock Fast effective rule induction.
\newblock In \emph{Machine Learning Proceedings 1995}. Elsevier, 1995.

\bibitem[Confalonieri et~al.(2019)Confalonieri, del Prado, Agramunt,
  Malagarriga, Faggion, Weyde, and Besold]{Roberto2019}
R.~Confalonieri, F.~Moscoso del Prado, S.~Agramunt, D.~Malagarriga, D.~Faggion,
  T.~Weyde, and T.~R. Besold.
\newblock An ontology-based approach to explaining artificial neural networks.
\newblock \emph{CoRR}, abs/1906.08362, 2019.
\newblock URL \url{http://arxiv.org/abs/1906.08362}.

\bibitem[Cortez and Embrechts(2011)]{cortez2011opening}
P.~Cortez and M.~J. Embrechts.
\newblock Opening black box data mining models using sensitivity analysis.
\newblock In \emph{Computational Intelligence and Data Mining (CIDM), 2011 IEEE
  Symposium on}. IEEE, 2011.

\bibitem[Craven and Shavlik(1996)]{craven1996extracting}
M.~Craven and J.~W. Shavlik.
\newblock Extracting tree-structured representations of trained networks.
\newblock In \emph{Advances in neural information processing systems}, 1996.

\bibitem[Cui et~al.(2015)Cui, Chen, He, and Chen]{cui2015optimal}
Z.~Cui, W.~Chen, Y.~He, and Y.~Chen.
\newblock Optimal action extraction for random forests and boosted trees.
\newblock In \emph{Proceedings of the 21th ACM SIGKDD international conference
  on knowledge discovery and data mining}, 2015.

\bibitem[D.~Brickley(2020)]{FOAF}
L.~Miller D.~Brickley.
\newblock The foaf project, 2020.
\newblock URL \url{http://www.foaf-project.org/}.

\bibitem[Datta et~al.(2016)Datta, Sen, and Zick]{datta2016algorithmic}
A.~Datta, S.~Sen, and Y.~Zick.
\newblock Algorithmic transparency via quantitative input influence: Theory and
  experiments with learning systems.
\newblock In \emph{Security and Privacy (SP), 2016 IEEE Symposium on}. IEEE,
  2016.

\bibitem[Doan et~al.(2004)Doan, Madhavan, Domingos, and Halevy]{Doan2004}
A.~Doan, J.~Madhavan, P.~Domingos, and A.~Halevy.
\newblock \emph{Ontology Matching: A Machine Learning Approach}, pages
  385--403.
\newblock Springer Berlin Heidelberg, Berlin, Heidelberg, 2004.
\newblock ISBN 978-3-540-24750-0.
\newblock \doi{10.1007/978-3-540-24750-0_19}.
\newblock URL \url{https://doi.org/10.1007/978-3-540-24750-0_19}.

\bibitem[Doran et~al.(2017)Doran, Schulz, and Besold]{doran2017does}
D.~Doran, S.~Schulz, and T.~R. Besold.
\newblock What does explainable ai really mean? a new conceptualization of
  perspectives.
\newblock \emph{arXiv preprint arXiv:1710.00794}, 2017.

\bibitem[Doshi-Velez and Kim(2017)]{doshi2017towards}
F.~Doshi-Velez and B.~Kim.
\newblock Towards a rigorous science of interpretable machine learning.
\newblock \emph{arXiv preprint arXiv:1702.08608}, 2017.

\bibitem[Do{\v{s}}ilovi{\'c} et~al.(2018)Do{\v{s}}ilovi{\'c}, Br{\v{c}}i{\'c},
  and Hlupi{\'c}]{dovsilovic2018explainable}
F.~K. Do{\v{s}}ilovi{\'c}, M.~Br{\v{c}}i{\'c}, and N.~Hlupi{\'c}.
\newblock Explainable artificial intelligence: A survey.
\newblock In \emph{2018 41st International convention on information and
  communication technology, electronics and microelectronics (MIPRO)}, 2018.

\bibitem[Dua and Graff(2017)]{Dua:2019}
D.~Dua and C.~Graff.
\newblock {UCI} machine learning repository, 2017.
\newblock URL \url{http://archive.ics.uci.edu/ml}.

\bibitem[Efron et~al.(2004)Efron, Hastie, Johnstone, Tibshirani,
  et~al.]{efron2004least}
B.~Efron, T.~Hastie, I.~Johnstone, R.~Tibshirani, et~al.
\newblock Least angle regression.
\newblock \emph{The Annals of statistics}, 2004.

\bibitem[Etchells and Lisboa(2006)]{etchells2006orthogonal}
T.~A. Etchells and P.~J.~G. Lisboa.
\newblock Orthogonal search-based rule extraction (osre) for trained neural
  networks: a practical and efficient approach.
\newblock \emph{IEEE transactions on neural networks}, 2006.

\bibitem[Europa.eu(2017)]{europa}
Europa.eu.
\newblock Official journal of the european union: Regulations, June 2017.
\newblock URL
  \url{http://ec.europa.eu/justice/dataprotection/reform/files/regulation_oj_en.pdf}.

\bibitem[F.~Bao et~al.(2020)F.~Bao, Liu, Chen, Li, Zhang, Han, Zhu, Ge, Chen,
  et~al.]{bao2020triaging}
and Y.~He F.~Bao, J.~Liu, Y.~Chen, Q.~Li, C.~Zhang, L.~Han, B.~Zhu, Y.~Ge,
  S.~Chen, et~al.
\newblock Triaging moderate covid-19 and other viral pneumonias from routine
  blood tests.
\newblock \emph{arXiv preprint arXiv:2005.06546}, 2020.

\bibitem[Fan et~al.(2020)Fan, Liu, Chen, and Henderson]{fan2020investigation}
X.~Fan, S.~Liu, J.~Chen, and T.~C. Henderson.
\newblock An investigation of covid-19 spreading factors with explainable ai
  techniques.
\newblock \emph{arXiv preprint arXiv:2005.06612}, 2020.

\bibitem[{Fischer} et~al.(1990){Fischer}, {Mastaglio}, {Reeves}, and
  {Rieman}]{Fischer1990}
G.~{Fischer}, T.~{Mastaglio}, B.~{Reeves}, and J.~{Rieman}.
\newblock Minimalist explanations in knowledge-based systems.
\newblock In \emph{Twenty-Third Annual Hawaii International Conference on
  System Sciences}, volume~3, pages 309--317 vol.3, 1990.

\bibitem[Fisher et~al.(2018)Fisher, Rudin, and Dominici]{fisher2018model}
A.~Fisher, C.~Rudin, and F.~Dominici.
\newblock Model class reliance: Variable importance measures for any machine
  learning model class, from the" rashomon" perspective.
\newblock \emph{arXiv preprint arXiv:1801.01489}, 2018.

\bibitem[Freitas(2014)]{freitas2014comprehensible}
A.~Freitas.
\newblock Comprehensible classification models: a position paper.
\newblock \emph{ACM SIGKDD explorations newsletter}, 2014.

\bibitem[Friedman et~al.(2008)Friedman, Popescu,
  et~al.]{friedman2008predictive}
J.~H. Friedman, B.~E. Popescu, et~al.
\newblock Predictive learning via rule ensembles.
\newblock \emph{The Annals of Applied Statistics}, 2008.

\bibitem[Friedman et~al.(1997)Friedman, Geiger, and
  Goldszmidt]{friedman1997bayesian}
N.~Friedman, D.~Geiger, and M.~Goldszmidt.
\newblock Bayesian network classifiers.
\newblock \emph{Machine learning}, 1997.

\bibitem[Fu(1994)]{fu1994rule}
L.~Fu.
\newblock Rule generation from neural networks.
\newblock \emph{IEEE Transactions on Systems, Man, and Cybernetics}, 1994.

\bibitem[Fung et~al.(2008)Fung, Sandilya, and Rao]{fung2008rule}
G.~Fung, S.~Sandilya, and R.~B. Rao.
\newblock Rule extraction from linear support vector machines via mathematical
  programming.
\newblock In \emph{Rule Extraction from Support Vector Machines}. Springer,
  2008.

\bibitem[Geng et~al.(2019)Geng, Chen, Jimenez-Ruiz, and Chen]{Geng2019}
Y.~Geng, J.~Chen, E.~Jimenez-Ruiz, and H.~Chen.
\newblock Human-centric transfer learning explanation via knowledge graph
  [extended abstract], 2019.

\bibitem[Gilpin et~al.(2018)Gilpin, Bau, Yuan, Bajwa, Specter, and
  Kagal]{gilpin2018explaining}
L.~H. Gilpin, D.~Bau, B.~Z. Yuan, A.~Bajwa, M.~Specter, and L.~Kagal.
\newblock Explaining explanations: An overview of interpretability of machine
  learning.
\newblock In \emph{2018 IEEE 5th International Conference on data science and
  advanced analytics (DSAA)}, 2018.

\bibitem[Gkatzia et~al.(2016)Gkatzia, Lemon, and Rieser]{gkatzia2016natural}
D.~Gkatzia, O.~Lemon, and V.~Rieser.
\newblock Natural language generation enhances human decision-making with
  uncertain information.
\newblock \emph{arXiv preprint arXiv:1606.03254}, 2016.

\bibitem[Goldstein et~al.(2015)Goldstein, Kapelner, Bleich, and
  Pitkin]{goldstein2015peeking}
A.~Goldstein, A.~Kapelner, J.~Bleich, and E.~Pitkin.
\newblock Peeking inside the black box: Visualizing statistical learning with
  plots of individual conditional expectation.
\newblock \emph{Journal of Computational and Graphical Statistics}, 2015.

\bibitem[Goodman and Flaxman(2016{\natexlab{a}})]{goodman}
B.~Goodman and S.~Flaxman.
\newblock European union regulations on algorithmic decision-making and a
  "right to explanation".
\newblock In \emph{ICML Workshop on Human Interpretability in Machine
  Learning}, 2016{\natexlab{a}}.

\bibitem[Goodman and Flaxman(2016{\natexlab{b}})]{goodman2016eu}
B.~Goodman and S.~Flaxman.
\newblock Eu regulations on algorithmic decision-making and a ``right to
  explanation''.
\newblock In \emph{ICML workshop on human interpretability in machine learning
  (WHI 2016), New York, NY}, 2016{\natexlab{b}}.

\bibitem[Gruber et~al.(1993)]{gruber1993translation}
Thomas~R Gruber et~al.
\newblock A translation approach to portable ontology specifications.
\newblock \emph{Knowledge acquisition}, 5\penalty0 (2):\penalty0 199--221,
  1993.

\bibitem[Gudivada et~al.(2017)Gudivada, Apon, and Ding]{gudivada2017data}
V.~Gudivada, A.~Apon, and J.~Ding.
\newblock Data quality considerations for big data and machine learning: Going
  beyond data cleaning and transformations.
\newblock \emph{International Journal on Advances in Software}, 10\penalty0
  (1):\penalty0 1--20, 2017.

\bibitem[Guidotti et~al.(2018{\natexlab{a}})Guidotti, Monreale, Ruggieri,
  Pedreschi, Turini, and Giannotti]{lore}
R.~Guidotti, A.~Monreale, S.~Ruggieri, D.~Pedreschi, F.~Turini, and
  F.~Giannotti.
\newblock Local rule-based explanations of black box decision systems.
\newblock \emph{arXiv preprint arXiv:1805.10820}, 2018{\natexlab{a}}.

\bibitem[Guidotti et~al.(2018{\natexlab{b}})Guidotti, Monreale, Ruggieri,
  Turini, Giannotti, and Pedreschi]{guidotti2018}
R.~Guidotti, A.~Monreale, S.~Ruggieri, F.~Turini, F.~Giannotti, and
  D.~Pedreschi.
\newblock A survey of methods for explaining black box models.
\newblock \emph{ACM Comput. Surv.}, 2018{\natexlab{b}}.

\bibitem[Gunning(2017)]{gunning2017explainable}
D.~Gunning.
\newblock Explainable artificial intelligence (xai).
\newblock \emph{Defense Advanced Research Projects Agency (DARPA)}, 2017.

\bibitem[Gurumoorthy et~al.(2017)Gurumoorthy, Dhurandhar, and
  Cecchi]{gurumoorthy2017protodash}
K.~S. Gurumoorthy, A.~Dhurandhar, and G.~Cecchi.
\newblock Protodash: Fast interpretable prototype selection.
\newblock \emph{arXiv preprint arXiv:1707.01212}, 2017.

\bibitem[Hall et~al.(2017{\natexlab{a}})Hall, Gill, Kurka, and
  Phan]{hall2017machine}
P.~Hall, N.~Gill, M.~Kurka, and W.~Phan.
\newblock Machine learning interpretability with h2o driverless ai.
\newblock \emph{H2O.ai}, 2017{\natexlab{a}}.

\bibitem[Hall et~al.(2017{\natexlab{b}})Hall, Phan, and Ambati]{hall2017ideas}
P.~Hall, W.~Phan, and S.~Ambati.
\newblock Ideas on interpreting machine learning, March 2017{\natexlab{b}}.
\newblock URL
  \url{https://www.oreilly.com/ideas/ideas-on-interpreting-machine-learning}.

\bibitem[Hara and Hayashi(2016)]{hara2016making}
S.~Hara and K.~Hayashi.
\newblock Making tree ensembles interpretable.
\newblock \emph{arXiv preprint arXiv:1606.05390}, 2016.

\bibitem[Hayashi(2013)]{hayashi2013neural}
Y.~Hayashi.
\newblock Neural network rule extraction by a new ensemble concept and its
  theoretical and historical background: A review.
\newblock \emph{International Journal of Computational Intelligence and
  Applications}, 2013.

\bibitem[Helfert and Ge(2016)]{helfert2016big}
M.~Helfert and M.~Ge.
\newblock Big data quality-towards an explanation model in a smart city
  context.
\newblock In \emph{proceedings of 21st International Conference on Information
  Quality, Ciudad Real, Spain}, 2016.

\bibitem[Hendricks et~al.(2016)Hendricks, Akata, Rohrbach, Donahue, Schiele,
  and Darrell]{hendricks2016generating}
L.~A. Hendricks, Z.~Akata, M.~Rohrbach, J.~Donahue, B.~Schiele, and T.~Darrell.
\newblock Generating visual explanations.
\newblock In \emph{European Conference on Computer Vision}. Springer, 2016.

\bibitem[Henelius et~al.(2014)Henelius, Puolam{\"a}ki, Bostr{\"o}m, Asker, and
  Papapetrou]{henelius2014peek}
A.~Henelius, K.~Puolam{\"a}ki, H.~Bostr{\"o}m, L.~Asker, and P.~Papapetrou.
\newblock A peek into the black box: exploring classifiers by randomization.
\newblock \emph{Data mining and knowledge discovery}, 2014.

\bibitem[Henelius et~al.(2017)Henelius, Puolam{\"a}ki, and
  Ukkonen]{henelius2017interpreting}
A.~Henelius, K.~Puolam{\"a}ki, and A.~Ukkonen.
\newblock Interpreting classifiers through attribute interactions in datasets.
\newblock In \emph{2017 ICML Workshop on Human Interpretability in Machine
  Learning (WHI)}, 2017.

\bibitem[Hepp(2020)]{GoodR}
M.~Hepp.
\newblock Good relations, 2020.
\newblock URL \url{http://www.heppnetz.de/projects/goodrelations/}.

\bibitem[Herman(2017)]{herman2017promise}
B.~Herman.
\newblock The promise and peril of human evaluation for model interpretability.
\newblock \emph{arXiv preprint arXiv:1711.07414}, 2017.

\bibitem[Hilton(1990)]{hilton1990conversational}
D.~J. Hilton.
\newblock Conversational processes and causal explanation.
\newblock \emph{Psychological Bulletin}, 1990.

\bibitem[Hinton and Frosst(2017)]{hinton2017distilling}
G.~Hinton and N.~Frosst.
\newblock Distilling a neural network into a soft decision tree.
\newblock In \emph{Comprehensibility and Explanation in AI and ML (CEX),
  AI*IA}, 2017.

\bibitem[Hoehndorf(2010)]{hoehndorf2010upper}
Robert Hoehndorf.
\newblock What is an upper level ontology?
\newblock \emph{Ontogenesis}, 2010.

\bibitem[Hoffman et~al.(2018)Hoffman, Mueller, Klein, and
  Litman]{hoffman2018metrics}
R.~Hoffman, S.~Mueller, G.~Klein, and J.~Litman.
\newblock Metrics for explainable ai: Challenges and prospects.
\newblock \emph{arXiv preprint arXiv:1812.04608}, 2018.

\bibitem[Holte(1993)]{holte1993very}
R.~C. Holte.
\newblock Very simple classification rules perform well on most commonly used
  datasets.
\newblock \emph{Machine learning}, 1993.

\bibitem[Holzinger et~al.(2017)Holzinger, Plass, Holzinger, Crisan, Pintea, and
  Palade]{holzinger2017glass}
A.~Holzinger, M.~Plass, K.~Holzinger, G.~C. Crisan, C.~M. Pintea, and
  V.~Palade.
\newblock A glass-box interactive machine learning approach for solving np-hard
  problems with the human-in-the-loop.
\newblock \emph{arXiv preprint arXiv:1708.01104}, 2017.

\bibitem[Holzinger et~al.(2019{\natexlab{a}})Holzinger, Kickmeier-Rust, and
  M{\"u}ller]{holzinger2019kandinsky}
A.~Holzinger, M.~Kickmeier-Rust, and H.~M{\"u}ller.
\newblock Kandinsky patterns as iq-test for machine learning.
\newblock In \emph{International Cross-Domain Conference for Machine Learning
  and Knowledge Extraction}, pages 1--14. Springer, 2019{\natexlab{a}}.

\bibitem[Holzinger et~al.(2019{\natexlab{b}})Holzinger, Langs, Denk, Zatloukal,
  and M{\"u}ller]{holzinger2019causability}
A.~Holzinger, G.~Langs, H.~Denk, K.~Zatloukal, and H.~M{\"u}ller.
\newblock Causability and explainabilty of artificial intelligence in medicine.
\newblock \emph{Wiley Interdisciplinary Reviews: Data Mining and Knowledge
  Discovery}, 2019{\natexlab{b}}.

\bibitem[Holzinger et~al.(2019{\natexlab{c}})Holzinger, Plass, Kickmeier-Rust,
  Holzinger, Cri{\c{s}}an, Pintea, and Palade]{holzinger2019interactive}
A.~Holzinger, M.~Plass, M.~Kickmeier-Rust, K.~Holzinger, G.~C. Cri{\c{s}}an,
  C.~M. Pintea, and V.~Palade.
\newblock Interactive machine learning: experimental evidence for the human in
  the algorithmic loop.
\newblock \emph{Applied Intelligence}, 49\penalty0 (7):\penalty0 2401--2414,
  2019{\natexlab{c}}.

\bibitem[Hoyt et~al.(2016)Hoyt, Snider, Thompson, and Mantravadi]{watson}
R.~E. Hoyt, D.~Snider, C.~Thompson, and S.~Mantravadi.
\newblock Ibm watson analytics: Automating visualization, descriptive, and
  predictive statistics.
\newblock \emph{JMIR Public Health Surveill}, 2\penalty0 (2):\penalty0 e157, 10
  2016.
\newblock ISSN 2369-2960.
\newblock \doi{10.2196/publichealth.5810}.
\newblock URL \url{http://publichealth.jmir.org/2016/2/e157/}.

\bibitem[Huysmans et~al.(2006)Huysmans, Baesens, and
  Vanthienen]{huysmans2006iter}
J.~Huysmans, B.~Baesens, and J.~Vanthienen.
\newblock Iter: an algorithm for predictive regression rule extraction.
\newblock In \emph{International Conference on Data Warehousing and Knowledge
  Discovery}. Springer, 2006.

\bibitem[Huysmans et~al.(2011)Huysmans, Dejaeger, Mues, Vanthienen, and
  Baesens]{huysmans2011empirical}
J.~Huysmans, K.~Dejaeger, C.~Mues, J.~Vanthienen, and B.~Baesens.
\newblock An empirical evaluation of the comprehensibility of decision table,
  tree and rule based predictive models.
\newblock \emph{Decision Support Systems}, 2011.

\bibitem[Jain and Wallace(2019)]{jain2019attention}
S.~Jain and B.~C. Wallace.
\newblock Attention is not explanation.
\newblock \emph{arXiv preprint arXiv:1902.10186}, 2019.

\bibitem[Jiang and Owen(2002)]{jiang2002quasi}
T.~Jiang and A.~B. Owen.
\newblock Quasi-regression for visualization and interpretation of black box
  functions, 2002.
\newblock URL
  \url{https://pdfs.semanticscholar.org/92d0/1110d9d365d16f619fb303932bee3274ba8f.pdf}.

\bibitem[Johansson et~al.(2004)Johansson, K{\"o}nig, and
  Niklasson]{johansson2004truth}
U.~Johansson, R.~K{\"o}nig, and L.~Niklasson.
\newblock The truth is in there-rule extraction from opaque models using
  genetic programming.
\newblock In \emph{FLAIRS Conference}. Miami Beach, FL, 2004.

\bibitem[Kabra et~al.(2015)Kabra, Robie, and Branson]{kabra2015understanding}
M.~Kabra, A.~Robie, and K.~Branson.
\newblock Understanding classifier errors by examining influential neighbors.
\newblock In \emph{Proceedings of the IEEE conference on computer vision and
  pattern recognition}, 2015.

\bibitem[Kaggle(2017)]{kaggle2017}
Kaggle.
\newblock The state of data science and machine learning, October 2017.
\newblock URL \url{https://www.kaggle.com/surveys/2017}.

\bibitem[Kamruzzaman(2010)]{kamruzzaman2010rex}
S.~Kamruzzaman.
\newblock Rex: An efficient rule generator.
\newblock \emph{arXiv preprint arXiv:1009.4988}, 2010.

\bibitem[Kass and Finin(1988)]{Kass1988}
R.~Kass and T.~Finin.
\newblock {The Need for User Models in Generating Expert System Explanations}.
\newblock \emph{International Journal of Expert Systems}, 1\penalty0 (4),
  October 1988.

\bibitem[Kim et~al.(2014)Kim, Rudin, and Shah]{kim2014bayesian}
B.~Kim, C.~Rudin, and J.~A. Shah.
\newblock The bayesian case model: A generative approach for case-based
  reasoning and prototype classification.
\newblock In \emph{Advances in Neural Information Processing Systems}, 2014.

\bibitem[Kim et~al.(2015)Kim, Shah, and Doshi-Velez]{kim2015mind}
B.~Kim, J.~A. Shah, and F.~Doshi-Velez.
\newblock Mind the gap: A generative approach to interpretable feature
  selection and extraction.
\newblock In \emph{Advances in Neural Information Processing Systems}, 2015.

\bibitem[Kim et~al.(2016)Kim, Khanna, and Koyejo]{kim2016examples}
B.~Kim, R.~Khanna, and O.~O. Koyejo.
\newblock Examples are not enough, learn to criticize! criticism for
  interpretability.
\newblock In \emph{Advances in Neural Information Processing Systems}, 2016.

\bibitem[Kittler(1986)]{kittler}
J.~Kittler.
\newblock Feature selection and extraction.
\newblock \emph{Handbook of Pattern Recognition and Image Processing}, 1986.

\bibitem[Koh and Liang(2017)]{koh2017understanding}
P.~W. Koh and P.~Liang.
\newblock Understanding black-box predictions via influence functions.
\newblock \emph{arXiv preprint arXiv:1703.04730}, 2017.

\bibitem[Kramer(1991)]{kramer1991nonlinear}
M.~A. Kramer.
\newblock Nonlinear principal component analysis using autoassociative neural
  networks.
\newblock \emph{AIChE journal}, 37\penalty0 (2):\penalty0 233--243, 1991.

\bibitem[Krause et~al.(2016)Krause, Perer, and Ng]{krause2016interacting}
J.~Krause, A.~Perer, and K.~Ng.
\newblock Interacting with predictions: Visual inspection of black-box machine
  learning models.
\newblock In \emph{Proceedings of the 2016 CHI Conference on Human Factors in
  Computing Systems}. ACM, 2016.

\bibitem[Lage et~al.(2019)Lage, Chen, He, Narayanan, Kim, Gershman, and
  Doshi-Velez]{lage2019evaluation}
I.~Lage, E.~Chen, J.~He, M.~Narayanan, B.~Kim, S.~Gershman, and F.~Doshi-Velez.
\newblock An evaluation of the human-interpretability of explanation.
\newblock \emph{arXiv preprint arXiv:1902.00006}, 2019.

\bibitem[Lakkaraju et~al.(2016)Lakkaraju, Bach, and
  Leskovec]{lakkaraju2016interpretable}
H.~Lakkaraju, S.~H. Bach, and J.~Leskovec.
\newblock Interpretable decision sets: A joint framework for description and
  prediction.
\newblock In \emph{Proceedings of the 22nd ACM SIGKDD International Conference
  on Knowledge Discovery and Data Mining}. ACM, 2016.

\bibitem[Lakkaraju et~al.(2017)Lakkaraju, Kamar, Caruana, and
  Leskovec]{lakkaraju2017interpretable}
H.~Lakkaraju, E.~Kamar, R.~Caruana, and J.~Leskovec.
\newblock Interpretable \& explorable approximations of black box models.
\newblock \emph{arXiv preprint arXiv:1707.01154}, 2017.

\bibitem[Lakkaraju et~al.(2019)Lakkaraju, Kamar, Caruana, and
  Leskovec]{lakkaraju2019faithful}
H.~Lakkaraju, E.~Kamar, R.~Caruana, and J.~Leskovec.
\newblock Faithful and customizable explanations of black box models.
\newblock In \emph{Proceedings of the 2019 AAAI/ACM Conference on AI, Ethics,
  and Society}, 2019.

\bibitem[Lash et~al.(2017)Lash, Lin, Street, and Robinson]{lash2017budget}
M.~T. Lash, Q.~Lin, W.~N. Street, and J.~G. Robinson.
\newblock A budget-constrained inverse classification framework for smooth
  classifiers.
\newblock In \emph{2017 IEEE International Conference on Data Mining Workshops
  (ICDMW)}, 2017.

\bibitem[Laugel et~al.(2017)Laugel, Lesot, Marsala, Renard, and
  Detyniecki]{laugel2017inverse}
T.~Laugel, M.~J. Lesot, C.~Marsala, X.~Renard, and M.~Detyniecki.
\newblock Inverse classification for comparison-based interpretability in
  machine learning.
\newblock \emph{arXiv preprint arXiv:1712.08443}, 2017.

\bibitem[Laugel et~al.(2018)Laugel, Renard, Lesot, Marsala, and
  Detyniecki]{laugel2018defining}
T.~Laugel, X.~Renard, M.~Lesot, C.~Marsala, and M.~Detyniecki.
\newblock Defining locality for surrogates in post-hoc interpretablity.
\newblock \emph{arXiv preprint arXiv:1806.07498}, 2018.

\bibitem[L{\'e}cu{\'e} et~al.(2019)L{\'e}cu{\'e}, Abeloos, Anctil, Bergeron,
  Dalla-Rosa, Corbeil-Letourneau, Martet, Pommellet, Salvan, Veilleux, and
  Ziaeefard]{Thales2019}
F.~L{\'e}cu{\'e}, B.~Abeloos, J.~Anctil, M.~Bergeron, D.~Dalla-Rosa,
  S.~Corbeil-Letourneau, F.~Martet, T.~Pommellet, L.~Salvan, S.~Veilleux, and
  M.~Ziaeefard.
\newblock Thales xai platform: Adaptable explanation of machine learning
  systems - a knowledge graphs perspective.
\newblock In \emph{ISWC Satellites}, 2019.

\bibitem[Lei et~al.(2018)Lei, G'Sell, Rinaldo, Tibshirani, and
  Wasserman]{lei2018}
J.~Lei, M.~G'Sell, A.~Rinaldo, R.~J. Tibshirani, and L.~Wasserman.
\newblock Distribution-free predictive inference for regression.
\newblock \emph{Journal of the American Statistical Association}, 2018.

\bibitem[Lei et~al.(2016)Lei, Barzilay, and Jaakkola]{lei2016rationalizing}
T.~Lei, R.~Barzilay, and T.~Jaakkola.
\newblock Rationalizing neural predictions.
\newblock \emph{arXiv preprint arXiv:1606.04155}, 2016.

\bibitem[Lent et~al.(2004)Lent, Fisher, and Mancuso]{van2004explainable}
M.~Van Lent, W.~Fisher, and M.~Mancuso.
\newblock An explainable artificial intelligence system for small-unit tactical
  behavior.
\newblock In \emph{Proceedings of the national conference on artificial
  intelligence}, 2004.

\bibitem[Letham et~al.(2012)Letham, Rudin, McCormick, and
  Madigan]{letham2012building}
B.~Letham, C.~Rudin, T.~H. McCormick, and D.~Madigan.
\newblock Building interpretable classifiers with rules using bayesian
  analysis.
\newblock \emph{Department of Statistics Technical Report tr609, University of
  Washington}, 2012.

\bibitem[Letham et~al.(2015)Letham, Rudin, McCormick, and
  Madigan]{letham2015interpretable}
Benjamin Letham, Cynthia Rudin, Tyler~H. McCormick, and David Madigan.
\newblock Interpretable classifiers using rules and bayesian analysis: Building
  a better stroke prediction model.
\newblock \emph{The Annals of Applied Statistics}, 2015.

\bibitem[Lipton et~al.(2016)Lipton, Kale, and Wetzel]{lipton2016modeling}
Z.~Lipton, D.~Kale, and R.~Wetzel.
\newblock Modeling missing data in clinical time series with rnns.
\newblock \emph{arXiv preprint arXiv:1606.04130}, 2016.

\bibitem[Lipton(2016)]{lipton2016mythos}
Z.~C. Lipton.
\newblock The mythos of model interpretability.
\newblock \emph{arXiv preprint arXiv:1606.03490}, 2016.

\bibitem[Lipton(2017)]{lipton2017doctor}
Z.~C. Lipton.
\newblock The doctor just won't accept that!
\newblock \emph{arXiv preprint arXiv:1711.08037}, 2017.

\bibitem[Looveren and Klaise(2019)]{van2019interpretable}
A.~Van Looveren and J.~Klaise.
\newblock Interpretable counterfactual explanations guided by prototypes.
\newblock \emph{arXiv preprint arXiv:1907.02584}, 2019.

\bibitem[Lou et~al.(2012)Lou, Caruana, and Gehrke]{lou2012intelligible}
Y.~Lou, R.~Caruana, and J.~Gehrke.
\newblock Intelligible models for classification and regression.
\newblock In \emph{Proceedings of the 18th ACM SIGKDD international conference
  on Knowledge discovery and data mining}. ACM, 2012.

\bibitem[Lou et~al.(2013)Lou, Caruana, Gehrke, and Hooker]{lou2013accurate}
Y.~Lou, R.~Caruana, J.~Gehrke, and G.~Hooker.
\newblock Accurate intelligible models with pairwise interactions.
\newblock In \emph{Proceedings of the 19th ACM SIGKDD international conference
  on Knowledge discovery and data mining}. ACM, 2013.

\bibitem[Lu et~al.(1995)Lu, Setiono, and Liu]{setiono1995neurorule}
H.~Lu, R.~Setiono, and H.~Liu.
\newblock Neurorule: A connectionist approach to data mining.
\newblock In \emph{Proceedings of the 21st VLDB Conference Zurich,
  Switzerland}, 1995.

\bibitem[Lundberg and Lee(2017)]{lundberg2017unified}
S.~M. Lundberg and S.~Lee.
\newblock A unified approach to interpreting model predictions.
\newblock In I.~Guyon, U.~V. Luxburg, S.~Bengio, H.~Wallach, R.~Fergus,
  S.~Vishwanathan, and R.~Garnett, editors, \emph{Advances in Neural
  Information Processing Systems 30}, pages 4765--4774. Curran Associates,
  Inc., 2017.

\bibitem[Luong et~al.(2015)Luong, Pham, and Manning]{luong2015effective}
M.~T. Luong, H.~Pham, and C.~D. Manning.
\newblock Effective approaches to attention-based neural machine translation.
\newblock \emph{arXiv preprint arXiv:1508.04025}, 2015.

\bibitem[Maedche and Staab(2001)]{Maedche2001}
A.~Maedche and S.~Staab.
\newblock Ontology learning for the semantic web.
\newblock \emph{IEEE Intelligent Systems}, 16:\penalty0 72--79, 03 2001.
\newblock \doi{10.1109/5254.920602}.

\bibitem[Mahajan et~al.(2019)Mahajan, Tan, and Sharma]{mahajan2019preserving}
Divyat Mahajan, Chenhao Tan, and Amit Sharma.
\newblock Preserving causal constraints in counterfactual explanations for
  machine learning classifiers.
\newblock \emph{arXiv preprint arXiv:1912.03277}, 2019.

\bibitem[Malioutov et~al.(2017)Malioutov, Varshney, Emad, and
  Dash]{malioutov2017learning}
D.~M. Malioutov, K.~R. Varshney, A.~Emad, and S.~Dash.
\newblock Learning interpretable classification rules with boolean compressed
  sensing.
\newblock In \emph{Transparent Data Mining for Big and Small Data}. Springer,
  2017.

\bibitem[Markowska-Kaczmar and Chumieja(2004)]{markowska2004discovering}
U.~Markowska-Kaczmar and M.~Chumieja.
\newblock Discovering the mysteries of neural networks.
\newblock \emph{International Journal of Hybrid Intelligent Systems}, 2004.

\bibitem[Martens and Provost(2014)]{martens2014explaining}
D.~Martens and F.~Provost.
\newblock Explaining data-driven document classifications.
\newblock \emph{Mis Quarterly}, 2014.

\bibitem[Martens et~al.(2007{\natexlab{a}})Martens, Backer, Haesen, Vanthienen,
  Snoeck, and Baesens]{martens2007classification}
D.~Martens, M.~De Backer, R.~Haesen, J.~Vanthienen, M.~Snoeck, and B.~Baesens.
\newblock Classification with ant colony optimization.
\newblock \emph{IEEE Transactions on Evolutionary Computation},
  2007{\natexlab{a}}.

\bibitem[Martens et~al.(2007{\natexlab{b}})Martens, Baesens, Gestel, and
  Vanthienen]{martens2007comprehensible}
D.~Martens, B.~Baesens, T.~Van Gestel, and J.~Vanthienen.
\newblock Comprehensible credit scoring models using rule extraction from
  support vector machines.
\newblock \emph{European journal of operational research}, 2007{\natexlab{b}}.

\bibitem[Martens et~al.(2008)Martens, Huysmans, Setiono, Vanthienen, and
  Baesens]{martens2008rule}
D.~Martens, J.~Huysmans, R.~Setiono, J.~Vanthienen, and B.~Baesens.
\newblock Rule extraction from support vector machines: an overview of issues
  and application in credit scoring.
\newblock \emph{Rule extraction from support vector machines}, 2008.

\bibitem[Martens et~al.(2009)Martens, Baesens, and
  Gestel]{martens2009decompositional}
D.~Martens, B.~Baesens, and T.~Van Gestel.
\newblock Decompositional rule extraction from support vector machines by
  active learning.
\newblock \emph{IEEE Transactions on Knowledge and Data Engineering}, 2009.

\bibitem[Martens et~al.(2011)Martens, Vanthienen, Verbeke, and
  Baesens]{martens2011performance}
D.~Martens, J.~Vanthienen, W.~Verbeke, and B.~Baesens.
\newblock Performance of classification models from a user perspective.
\newblock \emph{Decision Support Systems}, 2011.

\bibitem[Mashayekhi and Gras(2017)]{mashayekhi2017rule}
M.~Mashayekhi and R.~Gras.
\newblock Rule extraction from decision trees ensembles: New algorithms based
  on heuristic search and sparse group lasso methods.
\newblock \emph{International Journal of Information Technology \& Decision
  Making}, 2017.

\bibitem[{\relax Massachusetts Institute of Technology}(2017)]{mit}
{\relax Massachusetts Institute of Technology}.
\newblock The moral machine, November 2017.
\newblock URL \url{http://moralmachine.mit.edu}.

\bibitem[McGuinness et~al.(2007)McGuinness, Ding, da~Silva, and
  Chang]{McGuinness2007}
D.~L. McGuinness, L.~Ding, P.Pinheiro da~Silva, and C.~Chang.
\newblock Pml 2: A modular explanation interlingua.
\newblock In \emph{ExaCt}, 2007.

\bibitem[Mead(1992)]{mead1992review}
A.~Mead.
\newblock Review of the development of multidimensional scaling methods.
\newblock \emph{Journal of the Royal Statistical Society: Series D (The
  Statistician)}, 41\penalty0 (1):\penalty0 27--39, 1992.

\bibitem[Meinshausen(2010)]{meinshausen2010node}
N.~Meinshausen.
\newblock Node harvest.
\newblock \emph{The Annals of Applied Statistics}, 2010.

\bibitem[Melis and Jaakkola(2018)]{melis2018towards}
D.~A. Melis and T.~Jaakkola.
\newblock Towards robust interpretability with self-explaining neural networks.
\newblock In \emph{Advances in Neural Information Processing Systems}, 2018.

\bibitem[Miller(2017)]{miller2017explanation}
T.~Miller.
\newblock Explanation in artificial intelligence: Insights from the social
  sciences.
\newblock \emph{arXiv preprint arXiv:1706.07269}, 2017.

\bibitem[Mohammed et~al.(2012)Mohammed, Benlamri, and Fong]{Osama2012}
O.~Mohammed, R.~Benlamri, and S.~Fong.
\newblock Building a diseases symptoms ontology for medical diagnosis: An
  integrative approach.
\newblock In \emph{The First International Conference on Future Generation
  Communication Technologies}, 12 2012.
\newblock \doi{10.1109/FGCT.2012.6476567}.

\bibitem[Molnar(2018)]{molnar2018guide}
C.~Molnar.
\newblock \emph{A guide for making black box models explainable}.
\newblock 2018.

\bibitem[Montavon et~al.(2017)Montavon, Lapuschkin, Binder, Samek, and
  M{\"u}ller]{montavon2017explaining}
G.~Montavon, S.~Lapuschkin, A.~Binder, W.~Samek, and K.~R. M{\"u}ller.
\newblock Explaining nonlinear classification decisions with deep taylor
  decomposition.
\newblock \emph{Pattern Recognition}, 2017.

\bibitem[Montavon et~al.(2018)Montavon, Samek, and
  M{\"u}ller]{montavon2018methods}
G.~Montavon, W.~Samek, and K.~R. M{\"u}ller.
\newblock Methods for interpreting and understanding deep neural networks.
\newblock \emph{Digital Signal Processing}, 2018.

\bibitem[Murdoch et~al.(2019)Murdoch, Singh, Kumbier, Abbasi-Asl, and
  Yu]{murdoch2019interpretable}
J.~Murdoch, C.~Singh, K.~Kumbier, R.~Abbasi-Asl, and B.~Yu.
\newblock Interpretable machine learning: definitions, methods, and
  applications.
\newblock \emph{arXiv preprint arXiv:1901.04592}, 2019.

\bibitem[Navigli and Velardi(2004)]{Navigli2004}
R.~Navigli and P.~Velardi.
\newblock Learning domain ontologies from document warehouses and dedicated web
  sites.
\newblock \emph{Computational Linguistics}, 30\penalty0 (2):\penalty0 151--179,
  2004.
\newblock \doi{10.1162/089120104323093276}.

\bibitem[Ninama(2013)]{ninama2013ensemble}
H.~Ninama.
\newblock Ensemble approach for rule extraction in data mining.
\newblock \emph{Golden Reaserch Thoughts}, 2013.

\bibitem[Odajima et~al.(2008)Odajima, Hayashi, Tianxia, and
  Setiono]{odajima2008greedy}
K.~Odajima, Y.~Hayashi, G.~Tianxia, and R.~Setiono.
\newblock Greedy rule generation from discrete data and its use in neural
  network rule extraction.
\newblock \emph{Neural Networks}, 2008.

\bibitem[Otero and Freitas(2016)]{otero2016improving}
F.~E.~B. Otero and A.~Freitas.
\newblock Improving the interpretability of classification rules discovered by
  an ant colony algorithm: Extended results.
\newblock \emph{Evolutionary Computation}, 2016.

\bibitem[Panigutti et~al.(2020)Panigutti, Perotti, and
  Pedreschi]{Panigutti2020}
C.~Panigutti, A.~Perotti, and D.~Pedreschi.
\newblock Doctor xai: An ontology-based approach to black-box sequential data
  classification explanations.
\newblock In \emph{Proceedings of the 2020 Conference on Fairness,
  Accountability, and Transparency}, FAT* '20, pages 629--639, New York, NY,
  USA, 2020. Association for Computing Machinery.
\newblock ISBN 9781450369367.
\newblock \doi{10.1145/3351095.3372855}.
\newblock URL \url{https://doi.org/10.1145/3351095.3372855}.

\bibitem[Park et~al.(2016)Park, Hendricks, Akata, Schiele, Darrell, and
  Rohrbach]{park2016attentive}
D.~H. Park, L.~A. Hendricks, Z.~Akata, B.~Schiele, T.~Darrell, and M.~Rohrbach.
\newblock Attentive explanations: Justifying decisions and pointing to the
  evidence.
\newblock \emph{arXiv preprint arXiv:1612.04757}, 2016.

\bibitem[Phillips et~al.(2017)Phillips, Chang, and
  Friedler]{phillips2017interpretable}
R.~L. Phillips, K.~H. Chang, and S.~A. Friedler.
\newblock Interpretable active learning.
\newblock \emph{arXiv preprint arXiv:1708.00049}, 2017.

\bibitem[Plumb et~al.(2018)Plumb, Molitor, and Talwalkar]{maple}
G.~Plumb, D.~Molitor, and A.~S. Talwalkar.
\newblock Model agnostic supervised local explanations.
\newblock In \emph{Advances in Neural Information Processing Systems}, 2018.

\bibitem[Poursabzi-Sangdeh et~al.(2018)Poursabzi-Sangdeh, Goldstein, Hofman,
  Vaughan, and Wallach]{poursabzi2018manipulating}
F.~Poursabzi-Sangdeh, D.~G. Goldstein, J.~M. Hofman, J.~Wortman Vaughan, and
  H.~Wallach.
\newblock Manipulating and measuring model interpretability.
\newblock \emph{arXiv preprint arXiv:1802.07810}, 2018.

\bibitem[Poyiadzi et~al.(2020)Poyiadzi, Sokol, Santos-Rodriguez, De~Bie, and
  Flach]{poyiadzi2020face}
Rafael Poyiadzi, Kacper Sokol, Raul Santos-Rodriguez, Tijl De~Bie, and Peter
  Flach.
\newblock Face: feasible and actionable counterfactual explanations.
\newblock In \emph{Proceedings of the AAAI/ACM Conference on AI, Ethics, and
  Society}, pages 344--350, 2020.

\bibitem[Publio et~al.(2018)Publio, Esteves, Lawrynowicz, ce~Panov, Soldatova,
  Soru, Vanschoren, and Zafar]{Publio2018}
Gustavo~Correa Publio, Diego Esteves, Agnieszka Lawrynowicz, Pan\v ce~Panov,
  Larisa Soldatova, Tommaso Soru, Joaquin Vanschoren, and Hamid Zafar.
\newblock Ml-schema: Exposing the semantics of machine learning with schemas
  and ontologies, 2018.

\bibitem[Quinlan(1986)]{quinlan1986induction}
J.~R. Quinlan.
\newblock Induction of decision trees.
\newblock \emph{Machine learning}, 1986.

\bibitem[Quinlan(1996)]{quinlan1996bagging}
J.~R. Quinlan.
\newblock Bagging, boosting, and c4.5.
\newblock In \emph{AAAI/IAAI, Vol. 1}, 1996.

\bibitem[Quinlan(2014)]{quinlan2014c4}
J.~R. Quinlan.
\newblock \emph{C4.5: programs for machine learning}.
\newblock Elsevier, 2014.

\bibitem[Raimond et~al.(2007)Raimond, Abdallah, Sandler, and
  Giasson]{Raimond2007}
Y.~Raimond, S.~Abdallah, M.~Sandler, and F.~Giasson.
\newblock The music ontology.
\newblock In \emph{Proceedings of the 8th International Conference on Music
  Information Retrieval (ISMIR)}, 09 2007.

\bibitem[Raimond et~al.(2020)Raimond, Abdallah, Sandler, and Giasson]{Music}
Y.~Raimond, S.~Abdallah, M.~Sandler, and F.~Giasson.
\newblock The music ontology, 2020.
\newblock URL \url{http://musicontology.com/}.

\bibitem[Rezaul et~al.(2020)Rezaul, Döhmen, Rebholz-Schuhmann, Decker, Cochez,
  and Beyan]{rezaul2020deepcovidexplainer}
Karim Rezaul, Till Döhmen, Dietrich Rebholz-Schuhmann, Stefan Decker, Michael
  Cochez, and Oya Beyan.
\newblock Deepcovidexplainer: Explainable covid-19 predictions based on chest
  x-ray images.
\newblock \emph{arXiv}, pages arXiv--2004, 2020.

\bibitem[Ribeiro et~al.(2016{\natexlab{a}})Ribeiro, Singh, and
  Guestrin]{Ribeiro2016}
M.~T. Ribeiro, S.~Singh, and C.~Guestrin.
\newblock Why should i trust you?: Explaining the predictions of any
  classifier.
\newblock In \emph{Proceedings of the 22nd ACM SIGKDD International Conference
  on Knowledge Discovery and Data Mining}. ACM, 2016{\natexlab{a}}.

\bibitem[Ribeiro et~al.(2016{\natexlab{b}})Ribeiro, Singh, and
  Guestrin]{ribeiro2016model}
M.~T. Ribeiro, S.~Singh, and C.~Guestrin.
\newblock Model-agnostic interpretability of machine learning.
\newblock \emph{arXiv preprint arXiv:1606.05386}, 2016{\natexlab{b}}.

\bibitem[Ribeiro et~al.(2016{\natexlab{c}})Ribeiro, Singh, and
  Guestrin]{ribeiro2016should}
M.~T. Ribeiro, S.~Singh, and C.~Guestrin.
\newblock Why should i trust you?: Explaining the predictions of any
  classifier.
\newblock In \emph{Proceedings of the 22nd ACM SIGKDD international conference
  on knowledge discovery and data mining}. ACM, 2016{\natexlab{c}}.

\bibitem[Ribeiro et~al.(2018)Ribeiro, Singh, and Guestrin]{ribeiro2018anchors}
M.~T. Ribeiro, S.~Singh, and C.~Guestrin.
\newblock Anchors: High-precision model-agnostic explanations.
\newblock In \emph{Proceedings of the Thirty-Second AAAI Conference on
  Artificial Intelligence (AAAI)}, 2018.

\bibitem[Robnik and Kononenko(2008)]{robnik2007explaining}
M.~Robnik and I.~Kononenko.
\newblock Explaining classifications for individual instances.
\newblock \emph{IEEE Transactions on Knowledge and Data Engineering}, 2008.

\bibitem[Robnik-{\v{S}}ikonja and Kononenko(2008)]{robnik2008explaining}
M.~Robnik-{\v{S}}ikonja and I.~Kononenko.
\newblock Explaining classifications for individual instances.
\newblock \emph{IEEE Transactions on Knowledge and Data Engineering}, 2008.

\bibitem[Rudin(2018)]{stoprudin}
C.~Rudin.
\newblock Please stop explaining black box models for high stakes decisions.
\newblock \emph{CoRR}, 2018.

\bibitem[R{\"u}ping(2005)]{ruping2005learning}
S.~R{\"u}ping.
\newblock Learning with local models.
\newblock In \emph{Local Pattern Detection}, pages 153--170, 2005.

\bibitem[R{\"u}ping(2006)]{ruping2006learning}
S.~R{\"u}ping.
\newblock Learning interpretable models.
\newblock \emph{Doctoral Dissertation, University of Dortmund}, 2006.

\bibitem[Rush et~al.(2015)Rush, Chopra, and Weston]{rush2015neural}
A.~M. Rush, S.~Chopra, and J.~Weston.
\newblock A neural attention model for abstractive sentence summarization.
\newblock \emph{arXiv preprint arXiv:1509.00685}, 2015.

\bibitem[Russell(2019)]{russell2019efficient}
Chris Russell.
\newblock Efficient search for diverse coherent explanations.
\newblock In \emph{Proceedings of the Conference on Fairness, Accountability,
  and Transparency}, pages 20--28, 2019.

\bibitem[Saabas(2015)]{treeinterpreter}
A.~Saabas.
\newblock Treeinterpreter.
\newblock \url{https://github.com/andosa/treeinterpreter}, 2015.

\bibitem[Samek et~al.(2017)Samek, Wiegand, and
  M{\"u}ller]{samek2017explainable}
W.~Samek, T.~Wiegand, and K.~R. M{\"u}ller.
\newblock Explainable artificial intelligence: Understanding, visualizing and
  interpreting deep learning models.
\newblock \emph{arXiv preprint arXiv:1708.08296}, 2017.

\bibitem[Sarker et~al.(2017)Sarker, Xie, Doran, Raymer, and
  Hitzler]{Sarker2017}
M.~K. Sarker, N.~Xie, D.~Doran, M.~Raymer, and P.~Hitzler.
\newblock Explaining trained neural networks with semantic web technologies:
  First steps, 2017.

\bibitem[Schaaf and Huber(2019)]{schaaf2019enhancing}
N.~Schaaf and M.~F. Huber.
\newblock Enhancing decision tree based interpretation of deep neural networks
  through l1-orthogonal regularization.
\newblock \emph{arXiv preprint arXiv:1904.05394}, 2019.

\bibitem[Schetinin et~al.(2007)Schetinin, Fieldsend, Partridge, Coats,
  Krzanowski, Everson, Bailey, and Hernandez]{schetinin2007confident}
V.~Schetinin, J.~E. Fieldsend, D.~Partridge, T.~J. Coats, W.~J. Krzanowski,
  R.~M. Everson, T.~C. Bailey, and A.~Hernandez.
\newblock Confident interpretation of bayesian decision tree ensembles for
  clinical applications.
\newblock \emph{IEEE Transactions on Information Technology in Biomedicine},
  2007.

\bibitem[Schmidt and Biessmann(2019)]{schmidt2019quantifying}
P.~Schmidt and F.~Biessmann.
\newblock Quantifying interpretability and trust in machine learning systems.
\newblock \emph{arXiv preprint arXiv:1901.08558}, 2019.

\bibitem[Schmitz et~al.(1999)Schmitz, Aldrich, and Gouws]{schmitz1999ann}
G.~Schmitz, C.~Aldrich, and F.~S. Gouws.
\newblock Ann-dt: an algorithm for extraction of decision trees from artificial
  neural networks.
\newblock \emph{IEEE Transactions on Neural Networks}, 1999.

\bibitem[Selvaraju et~al.(2016)Selvaraju, Das, Vedantam, Cogswell, Parikh, and
  Batra]{selvaraju2017visual}
R.~R. Selvaraju, A.~Das, R.~Vedantam, M.~Cogswell, D.~Parikh, and D.~Batra.
\newblock Grad-cam: Why did you say that? visual explanations from deep
  networks via gradient-based localization.
\newblock \emph{arXiv preprint arXiv:1610.02391}, 2016.

\bibitem[Sestito and Dillon(1992)]{sestito1992automated}
S.~Sestito and T.S. Dillon.
\newblock Automated knowledge acquisition of rules with continuously valued
  attributes.
\newblock In \emph{Proceedings of the 12th international conference on expert
  systems and their applications, 1992}, 1992.

\bibitem[Sethi et~al.(2012)Sethi, Mishra, and Mishra]{sethi2012extended}
K.~K. Sethi, D.~K. Mishra, and B.~Mishra.
\newblock Extended taxonomy of rule extraction techniques and assessment of
  kdruleex.
\newblock \emph{International Journal of Computer Applications}, 2012.

\bibitem[Setiono and Liu(1997)]{setiono1997neurolinear}
R.~Setiono and H.~Liu.
\newblock Neurolinear: From neural networks to oblique decision rules.
\newblock \emph{Neurocomputing}, 1997.

\bibitem[Setiono et~al.(2008)Setiono, Baesens, and Mues]{setiono2008recursive}
R.~Setiono, B.~Baesens, and C.~Mues.
\newblock Recursive neural network rule extraction for data with mixed
  attributes.
\newblock \emph{IEEE Transactions on Neural Networks}, 2008.

\bibitem[Setiono et~al.(2014)Setiono, Azcarraga, and Hayashi]{setiono2014mofn}
R.~Setiono, A.~Azcarraga, and Y.~Hayashi.
\newblock Mofn rule extraction from neural networks trained with augmented
  discretized input.
\newblock In \emph{Neural Networks (IJCNN), 2014 International Joint Conference
  on}. IEEE, 2014.

\bibitem[Shapley(1951)]{shapley1951notes}
L.~S. Shapley.
\newblock Notes on the n-person game--ii: The value of an n-person game.
\newblock Technical report, U.S. Air Force, Project Rand, 1951.

\bibitem[Shrikumar et~al.(2016)Shrikumar, Greenside, Shcherbina, and
  Kundaje]{shrikumar}
A.~Shrikumar, P.~Greenside, A.~Shcherbina, and A.~Kundaje.
\newblock Not just a black box: Learning important features through propagating
  activation differences.
\newblock In \emph{33rd International Conference on Machine Learning}, 2016.

\bibitem[Si and Zhu(2013)]{si2013learning}
Z.~Si and S.~C. Zhu.
\newblock Learning and-or templates for object recognition and detection.
\newblock \emph{IEEE transactions on pattern analysis and machine
  intelligence}, 2013.

\bibitem[Smilkov et~al.(2017)Smilkov, Thorat, Kim, Vi{\'e}gas, and
  Wattenberg]{smilkov2017smoothgrad}
D.~Smilkov, N.~Thorat, B.~Kim, F.~Vi{\'e}gas, and M.~Wattenberg.
\newblock Smoothgrad: removing noise by adding noise.
\newblock \emph{arXiv preprint arXiv:1706.03825}, 2017.

\bibitem[{S}trumbelj and Kononenko(2014)]{vstrumbelj2014explaining}
E.~{S}trumbelj and I.~Kononenko.
\newblock Explaining prediction models and individual predictions with feature
  contributions.
\newblock \emph{Knowledge and information systems}, 2014.

\bibitem[{S}trumbelj et~al.(2010){S}trumbelj, Bosni{\'c}, Kononenko, Zakotnik,
  and Kuhar]{strumbelj2010explanation}
E.~{S}trumbelj, Z.~Bosni{\'c}, I.~Kononenko, B.~Zakotnik, and C.~Kuhar.
\newblock Explanation and reliability of prediction models: the case of breast
  cancer recurrence.
\newblock \emph{Knowledge and information systems}, 2010.

\bibitem[Su et~al.(2015)Su, Wei, Varshney, and Malioutov]{su2015interpretable}
G.~Su, D.~Wei, K.~R. Varshney, and D.~M. Malioutov.
\newblock Interpretable two-level boolean rule learning for classification.
\newblock \emph{arXiv preprint arXiv:1511.07361}, 2015.

\bibitem[Su et~al.(2016)Su, Wei, Varshney, and Malioutov]{su2016learning}
G.~Su, D.~Wei, K.~R. Varshney, and D.~M. Malioutov.
\newblock Learning sparse two-level boolean rules.
\newblock In \emph{2016 IEEE 26th International Workshop on Machine Learning
  for Signal Processing (MLSP)}. IEEE, 2016.

\bibitem[Subianto and Siebes(2007)]{subianto2007understanding}
M.~Subianto and A.~Siebes.
\newblock Understanding discrete classifiers with a case study in gene
  prediction.
\newblock In \emph{Seventh IEEE International Conference on Data Mining 2007},
  pages 661--666. IEEE, 2007.

\bibitem[Sundararajan et~al.(2016)Sundararajan, Taly, and
  Yan]{sundararajan2016gradients}
M.~Sundararajan, A.~Taly, and Q.~Yan.
\newblock Gradients of counterfactuals.
\newblock \emph{arXiv preprint arXiv:1611.02639}, 2016.

\bibitem[{Swartout} et~al.(1991){Swartout}, {Paris}, and {Moore}]{Swartout1991}
W.~{Swartout}, C.~{Paris}, and J.~{Moore}.
\newblock Explanations in knowledge systems: design for explainable expert
  systems.
\newblock \emph{IEEE Expert}, 6\penalty0 (3):\penalty0 58--64, 1991.

\bibitem[Taha and Ghosh(1996)]{taha1996three}
I.~Taha and J.~Ghosh.
\newblock Three techniques for extracting rules from feedforward networks.
\newblock \emph{Intelligent Engineering Systems Through Artificial Neural
  Networks}, 1996.

\bibitem[Tibshirani(1996)]{tibshirani1996regression}
R.~Tibshirani.
\newblock Regression shrinkage and selection via the lasso.
\newblock \emph{Journal of the Royal Statistical Society: Series B
  (Methodological)}, 1996.

\bibitem[Tjoa and Guan(2019)]{tjoa2019survey}
E.~Tjoa and C.~Guan.
\newblock A survey on explainable artificial intelligence (xai): Towards
  medical xai.
\newblock \emph{arXiv preprint arXiv:1907.07374}, 2019.

\bibitem[Tolomei et~al.(2017)Tolomei, Silvestri, Haines, and
  Lalmas]{tolomei2017interpretable}
G.~Tolomei, F.~Silvestri, A.~Haines, and M.~Lalmas.
\newblock Interpretable predictions of tree-based ensembles via actionable
  feature tweaking.
\newblock In \emph{Proceedings of the 23rd ACM SIGKDD International Conference
  on Knowledge Discovery and Data Mining}. ACM, 2017.

\bibitem[Tsymbal et~al.(2007)Tsymbal, Zillner, and Huber]{Tysmbal2007}
A.~Tsymbal, S.~Zillner, and M.~Huber.
\newblock {Ontology - Supported Machine Learning and Decision Support in
  Biomedicine}.
\newblock In \emph{International Conference on Data Integration in the Life
  Sciences}, volume 4544, pages 156--171, 06 2007.
\newblock \doi{10.1007/978-3-540-73255-6_14}.

\bibitem[Turner(2016)]{turner}
R.~Turner.
\newblock A model explanation system.
\newblock In \emph{2016 IEEE 26th International Workshop on Machine Learning
  for Signal Processing (MLSP)}, 2016.

\bibitem[Tversky and Kahneman(1974)]{tversky1974judgment}
A.~Tversky and D.~Kahneman.
\newblock Judgment under uncertainty: Heuristics and biases.
\newblock \emph{Science}, 1974.

\bibitem[Tversky and Kahneman(1981)]{tversky1981framing}
A.~Tversky and D.~Kahneman.
\newblock The framing of decisions and the psychology of choice.
\newblock \emph{Science}, 1981.

\bibitem[U.~Bojars(2020)]{SIOC}
J.~G.~Breslin U.~Bojars.
\newblock Semantically-interlinked online communities, 2020.
\newblock URL \url{http://sioc-project.org/}.

\bibitem[Ustun and Rudin(2014)]{ustun2014methods}
B.~Ustun and C.~Rudin.
\newblock Methods and models for interpretable linear classification.
\newblock \emph{arXiv preprint arXiv:1405.4047}, 2014.

\bibitem[Ustun and Rudin(2016)]{ustun2016supersparse}
B.~Ustun and C.~Rudin.
\newblock Supersparse linear integer models for optimized medical scoring
  systems.
\newblock \emph{Machine Learning}, 2016.

\bibitem[Ustun and Rudin(2017)]{ustun2017optimized}
B.~Ustun and C.~Rudin.
\newblock Optimized risk scores.
\newblock In \emph{Proceedings of the 23rd ACM SIGKDD International Conference
  on Knowledge Discovery and Data Mining}. ACM, 2017.

\bibitem[van der and Hinton(2008)]{maaten2008visualizing}
L.~Maaten van der and G.~Hinton.
\newblock Visualizing data using t-sne.
\newblock \emph{Journal of machine learning research}, 9\penalty0
  (Nov):\penalty0 2579--2605, 2008.

\bibitem[Vedantam et~al.(2017)Vedantam, Bengio, Murphy, Parikh, and
  Chechik]{vedantam2017context}
R.~Vedantam, S.~Bengio, K.~Murphy, D.~Parikh, and G.~Chechik.
\newblock Context-aware captions from context-agnostic supervision.
\newblock In \emph{Proceedings of the IEEE Conference on Computer Vision and
  Pattern Recognition}, 2017.

\bibitem[Verbeke et~al.(2011)Verbeke, Martens, Mues, and
  Baesens]{verbeke2011building}
W.~Verbeke, D.~Martens, C.~Mues, and B.~Baesens.
\newblock Building comprehensible customer churn prediction models with
  advanced rule induction techniques.
\newblock \emph{Expert Systems with Applications}, 2011.

\bibitem[Voosen(2017)]{voosen2017}
P.~Voosen.
\newblock How \textsc{AI} detectives are cracking open the black box of deep
  learning.
\newblock \emph{Science Magazine}, 2017.

\bibitem[Wachter et~al.(2018)Wachter, Mittelstadt, and
  Russell]{wachter2017counterfactual}
S.~Wachter, B.~Mittelstadt, and C.~Russell.
\newblock Counterfactual explanations without opening the black box: Automated
  decisions and the gdpr.
\newblock \emph{Harvard Journal of Law \& Technology}, 31\penalty0 (2), 2018.

\bibitem[Wang and Rudin(2014)]{wang2015falling}
F.~Wang and C.~Rudin.
\newblock Falling rule lists.
\newblock \emph{arXiv preprint arXiv:1411.5899}, 2014.

\bibitem[Wang and Rudin(2015)]{wang}
F.~Wang and C.~Rudin.
\newblock Falling rule lists.
\newblock In \emph{18th International Conference on Artificial Intelligence and
  Statistics (AISTATS)}, 2015.

\bibitem[Wang et~al.(2015{\natexlab{a}})Wang, Fujimaki, and
  Motohashi]{wang2015trading}
J.~Wang, R.~Fujimaki, and Y.~Motohashi.
\newblock Trading interpretability for accuracy: Oblique treed sparse additive
  models.
\newblock In \emph{Proceedings of the 21th ACM SIGKDD International Conference
  on Knowledge Discovery and Data Mining}. ACM, 2015{\natexlab{a}}.

\bibitem[Wang and Strong(1996)]{wang1996beyond}
R.~Y. Wang and D.~M. Strong.
\newblock Beyond accuracy: What data quality means to data consumers.
\newblock \emph{Journal of management information systems}, 12\penalty0
  (4):\penalty0 5--33, 1996.

\bibitem[Wang et~al.(2015{\natexlab{b}})Wang, Rudin, Doshi-Velez, Liu, Klampfl,
  and MacNeille]{wang2015bayesian}
T.~Wang, C.~Rudin, F.~Doshi-Velez, Y.~Liu, E.~Klampfl, and P.~MacNeille.
\newblock Or's of and's for interpretable classification, with application to
  context-aware recommender systems.
\newblock \emph{arXiv preprint arXiv:1504.07614}, 2015{\natexlab{b}}.

\bibitem[Wang et~al.(2016)Wang, Rudin, Velez-Doshi, Liu, Klampfl, and
  MacNeille]{wang2016bayesian}
T.~Wang, C.~Rudin, F.~Velez-Doshi, Y.~Liu, E.~Klampfl, and P.~MacNeille.
\newblock Bayesian rule sets for interpretable classification.
\newblock In \emph{Data Mining (ICDM), 2016 IEEE 16th International Conference
  on}. IEEE, 2016.

\bibitem[Weiner(1980)]{weiner1980blah}
JL~Weiner.
\newblock Blah, a system which explains its reasoning.
\newblock \emph{Artificial intelligence}, 15\penalty0 (1-2):\penalty0 19--48,
  1980.

\bibitem[Weller(2017)]{weller2017challenges}
A.~Weller.
\newblock Challenges for transparency.
\newblock \emph{arXiv preprint arXiv:1708.01870}, 2017.

\bibitem[West et~al.(2007)West, Ventura, and Warnick]{west2007}
J.~West, D.~Ventura, and S.~Warnick.
\newblock Spring research presentation: A theoretical foundation for inductive
  transfer, 2007.
\newblock Retrieved 2007-08-05.

\bibitem[Wiegreffe and Pinter(2019)]{wiegreffe2019attention}
S.~Wiegreffe and Y.~Pinter.
\newblock Attention is not not explanation.
\newblock \emph{arXiv preprint arXiv:1908.04626}, 2019.

\bibitem[Wold et~al.(1987)Wold, Esbense, and Geladi]{wold1987principal}
S.~Wold, K.~Esbense, and P.~Geladi.
\newblock Principal component analysis.
\newblock \emph{Chemometrics and intelligent laboratory systems}, 2\penalty0
  (1-3):\penalty0 37--52, 1987.

\bibitem[Wong et~al.(2011)Wong, Liu, and Bennamoun]{Wong2011}
W.~Wong, W.~Liu, and M.~Bennamoun.
\newblock Ontology learning from text: A look back and into the future.
\newblock \emph{ACM Computing Surveys - CSUR}, 44:\penalty0 1--36, 01 2011.
\newblock \doi{10.1145/2333112.2333115}.

\bibitem[Wu et~al.(2018)Wu, Hughes, Parbhoo, Zazzi, Roth, and
  Doshi-Velez]{wu2018beyond}
M.~Wu, M.~C. Hughes, S.~Parbhoo, M.~Zazzi, V.~Roth, and F.~Doshi-Velez.
\newblock Beyond sparsity: Tree regularization of deep models for
  interpretability.
\newblock In \emph{Thirty-Second AAAI Conference on Artificial Intelligence},
  2018.

\bibitem[Xu et~al.(2015{\natexlab{a}})Xu, Ba, Kiros, Cho, Courville,
  Salakhudinov, Zemel, and Bengio]{xu2015show}
K.~Xu, J.~Ba, R.~Kiros, K.~Cho, A.~Courville, R.~Salakhudinov, R.~Zemel, and
  Y.~Bengio.
\newblock Show, attend and tell: Neural image caption generation with visual
  attention.
\newblock In \emph{International conference on machine learning},
  2015{\natexlab{a}}.

\bibitem[Xu et~al.(2015{\natexlab{b}})Xu, Jiangping, Qi, Huang, and
  Lin]{Ning2015}
N.~Xu, W.~Jiangping, G.~Qi, T.~Huang, and W.~Lin.
\newblock Ontological random forests for image classification.
\newblock \emph{International Journal of Information Retrieval Research},
  5:\penalty0 61--74, 07 2015{\natexlab{b}}.
\newblock \doi{10.4018/IJIRR.2015070104}.

\bibitem[Yang et~al.(2018{\natexlab{a}})Yang, Rangarajan, and
  Ranka]{yang2018global}
C.~Yang, A.~Rangarajan, and S.~Ranka.
\newblock Global model interpretation via recursive partitioning.
\newblock \emph{arXiv preprint arXiv:1802.04253}, 2018{\natexlab{a}}.

\bibitem[Yang et~al.(2016)Yang, Rudin, and Seltzer]{hongyu}
H.~Yang, C.~Rudin, and M.~Seltzer.
\newblock Scalable bayesian rule lists.
\newblock unpublished, 2016.

\bibitem[Yang et~al.(2018{\natexlab{b}})Yang, Morillo, and
  Hospedales]{yang2018deep}
Y.~Yang, I.~G. Morillo, and T.~M. Hospedales.
\newblock Deep neural decision trees.
\newblock \emph{arXiv preprint arXiv:1806.06988}, 2018{\natexlab{b}}.

\bibitem[Yin et~al.(2019)Yin, Vaughan, and Wallach]{yin2019understanding}
M.~Yin, J.~Wortman Vaughan, and H.~Wallach.
\newblock Understanding the effect of accuracy on trust in machine learning
  models.
\newblock In \emph{Proceedings of the 2019 CHI Conference on Human Factors in
  Computing Systems}. ACM, 2019.

\bibitem[Yin and Han(2003)]{yin2003cpar}
X.~Yin and J.~Han.
\newblock Cpar: Classification based on predictive association rules.
\newblock In \emph{Proceedings of the 2003 SIAM International Conference on
  Data Mining}. SIAM, 2003.

\bibitem[Zhang and Chen(2018)]{zhang2018explainable}
Y.~Zhang and X.~Chen.
\newblock Explainable recommendation: A survey and new perspectives.
\newblock \emph{arXiv preprint arXiv:1804.11192}, 2018.

\bibitem[Zhao et~al.(2019)Zhao, Wu, Lee, and Cui]{zhao2019iforest}
X.~Zhao, Y.~Wu, D.~L. Lee, and W.~Cui.
\newblock iforest: Interpreting random forests via visual analytics.
\newblock \emph{IEEE transactions on visualization and computer graphics},
  2019.

\bibitem[Zhou et~al.(2000)Zhou, Chen, and Chen]{zhou2000statistics}
Z.~H. Zhou, S.~F. Chen, and Z.~Q. Chen.
\newblock A statistics based approach for extracting priority rules from
  trained neural networks.
\newblock In \emph{ijcnn}. IEEE, 2000.

\bibitem[Zhou et~al.(2003)Zhou, Jiang, and Chen]{zhou2003extracting}
Z.~H. Zhou, Y.~Jiang, and S.~F. Chen.
\newblock Extracting symbolic rules from trained neural network ensembles.
\newblock \emph{Ai Communications}, 2003.

\bibitem[Zilke et~al.(2016)Zilke, Menc{\'\i}a, and Janssen]{zilke2016deepred}
E.~Zilke, L.~Menc{\'\i}a, and F.~Janssen.
\newblock Deepred--rule extraction from deep neural networks.
\newblock In \emph{International Conference on Discovery Science}. Springer,
  2016.

\end{thebibliography}
\end{document}